\pdfoutput=1

\documentclass[11pt]{article}

\usepackage{EMNLP2022}

\usepackage{times}
\usepackage{latexsym}
\usepackage{xcolor}
\usepackage{soul}
\usepackage{hyperref}
\usepackage{amsmath}
\usepackage{graphicx}
\usepackage{booktabs}
\usepackage{multirow}
\usepackage{algorithm2e}
\usepackage[export]{adjustbox}
\usepackage{amssymb}
\usepackage{pifont}
\usepackage{color,soul}
\usepackage{pifont}
\usepackage{listings}
\definecolor{mygreen}{HTML}{006400}
\definecolor{mypurple}{HTML}{9D3972}

\definecolor{codegreen}{rgb}{0,0.6,0}
\definecolor{codegray}{rgb}{0.5,0.5,0.5}
\definecolor{codepurple}{rgb}{0.58,0,0.82}
\definecolor{backcolour}{rgb}{0.95,0.95,0.92}

\lstdefinestyle{mystyle}{
  commentstyle=\color{codegreen},
  keywordstyle=\color{magenta},
  numberstyle=\tiny\color{codegray},
  stringstyle=\color{codepurple},
  basicstyle=\ttfamily,
  breakatwhitespace=false,         
  breaklines=true,                 
  captionpos=b,                    
  keepspaces=true,                 
  numbers=left,                    
  numbersep=-7pt,                  
  showspaces=false,                
  showstringspaces=false,
  showtabs=false,                  
  tabsize=4
}
\usepackage[T1]{fontenc}

\usepackage[utf8]{inputenc}

\usepackage{microtype}

\newcommand{\titlestr}{\textsc{RankGen}: Improving Text Generation with Large Ranking Models}

\title{\titlestr}

\author{Kalpesh Krishna$^{\spadesuit\,*}$ \quad Yapei Chang$^{\spadesuit}$ \quad John Wieting$^\diamondsuit$ \quad Mohit Iyyer$^\spadesuit$ \\\\
$^\spadesuit$University of Massachusetts Amherst, $^\diamondsuit$Google Research \\ \texttt{\{kalpesh,miyyer\}@cs.umass.edu}\\ \texttt{jwieting@google.com} }

\newcommand{\model}{\textsc{RankGen}}
\newcommand{\inbookobjective}{\textsc{InBook}}
\newcommand{\genobjective}{\textsc{Generative}}
\newcommand{\modelallboth}{all-XL-both}

\newcommand{\namedref}[2]{\hyperref[#2]{#1~\ref*{#2}}}

\newcommand{\sectionref}[1]{\namedref{Section}{#1}}
\newcommand{\tableref}[1]{\namedref{Table}{#1}}
\newcommand{\figureref}[1]{\namedref{Figure}{#1}}
\newcommand{\appendixref}[1]{\namedref{Appendix}{#1}}

\newcommand{\prefixcolor}[1]{\textcolor{mygreen}{#1}}
\newcommand{\suffixcolor}[1]{\textcolor{mypurple}{#1}}
\newcommand{\prefix}[1]{#1}
\newcommand{\suffix}[1]{#1}

\definecolor{colorGrammarUsage}{HTML}{F5F5F5}
\definecolor{colorRedundant}{HTML}{FFC2BA}
\definecolor{colorOffPrompt}{HTML}{FFFFB3}
\definecolor{colorSelfContradiction}{HTML}{B9DEFF}
\definecolor{colorIncoherent}{HTML}{FCCDE5}
\definecolor{colorJargon}{HTML}{FEC88B}
\definecolor{colorBadMath}{HTML}{D9D9D9}
\definecolor{colorCommonsense}{HTML}{D7F3E7}
\definecolor{colorEncyclopedic}{HTML}{C8E792}
\definecolor{colorNeedsGoogle}{HTML}{E4B7FF}
\definecolor{someSpanText}{HTML}{000000} 
\definecolor{mycolor1}{HTML}{0000AA}
\definecolor{mycolor2}{HTML}{BB0000}
\definecolor{mycolor3}{HTML}{002200}

\newcommand{\errGrammarUsage[1]}{\sethlcolor{colorGrammarUsage}{\textbf{\hl{~#1~}}}}
\newcommand{\errRedundant[1]}{{\sethlcolor{colorRedundant}\textcolor{someSpanText}{\textbf{\hl{~#1~}}}}}
\newcommand{\errRedundantAntecedent[1]}{\setulcolor{colorRedundant}{\textcolor{someSpanText}{\textbf{\ul{~#1~}}}}}
\newcommand{\errOffPrompt[1]}{\sethlcolor{colorOffPrompt}{\textbf{\hl{~#1~}}}}
\newcommand{\errSelfContradiction[1]}{\sethlcolor{colorSelfContradiction}{\textcolor{someSpanText}{\textbf{\hl{~#1~}}}}}
\newcommand{\errSelfContradictionAntecedent[1]}{\setulcolor{colorSelfContradiction}{{\textbf{\ul{~#1~}}}}}
\newcommand{\errIncoherent[1]}{\sethlcolor{colorIncoherent}{\textcolor{someSpanText}{\textbf{\hl{~#1~}}}}}
\newcommand{\errJargon[1]}{\sethlcolor{colorJargon}{\textcolor{someSpanText}{\textbf{\hl{~#1~}}}}}
\newcommand{\errBadMath[1]}{\sethlcolor{colorBadMath}{\textcolor{someSpanText}{\textbf{\hl{~#1~}}}}}
\newcommand{\errCommonsense[1]}{\sethlcolor{colorCommonsense}{\textbf{\hl{~#1~}}}}
\newcommand{\errEncyclopedic[1]}{\sethlcolor{colorEncyclopedic}{\textcolor{someSpanText}{\textbf{\hl{~#1~}}}}}
\newcommand{\errNeedsGoogle[1]}{\sethlcolor{colorNeedsGoogle}{\textcolor{someSpanText}{\textbf{\hl{~#1~}}}}}

\begin{document}
\maketitle

\begin{abstract}
Given an input sequence (or \emph{prefix}), modern language models often assign high probabilities to output sequences that are repetitive, incoherent, or irrelevant to the prefix; as such, model-generated text also contains such artifacts. To address these issues we present \model,~a 1.2B parameter encoder model for English that scores model generations given a prefix. \model~can be flexibly incorporated as a scoring function in beam search and used to decode from any pretrained language model. We train \model\ using large-scale contrastive learning to map a prefix close to the ground-truth sequence that follows it and far away from two types of negatives: (1) random sequences from the same document as the prefix, and (2) sequences generated from a large language model conditioned on the prefix.
Experiments across four different language models (345M-11B parameters) and two domains show that \model~significantly outperforms decoding algorithms like nucleus, top-$k$, typical sampling, as well as contrastive decoding and search, on both automatic metrics (85.0 vs 77.3 MAUVE over nucleus) as well as human evaluations with English writers (74.5\% human preference over nucleus sampling).  Analysis reveals that \model\ outputs are more relevant to the prefix and improve continuity and coherence compared to baselines. We release our model checkpoints, code, and human preference data with explanations to facilitate future research.\footnote{All resources are available at \url{https://github.com/martiansideofthemoon/rankgen}. \\ *Work done as a student researcher at Google Research.} 
\end{abstract}
\section{Introduction}
Despite exciting recent progress in large-scale language modeling~\citep{radford2019language,brown2020language},  text generated from these language models (LMs) continues to be riddled with artifacts. Modern LMs suffer from the ``likelihood trap''~\citep{see-etal-2019-massively,zhang-etal-2021-trading}, in which high likelihood (low perplexity) sequences produced by greedy decoding or beam search tend to be dull and repetitive. While truncated sampling methods such as top-$k$~\citep{fan-etal-2018-hierarchical}, nucleus~\citep{holtzman2020curious}, and typical sampling~\citep{meister2022typical} alleviate these issues, they can also produce text with inconsistencies, hallucinations, factual errors, or commonsense issues~\citep{massarelli-etal-2020-decoding,dou-etal-2022-gpt,krishna-etal-2021-hurdles}.


\begin{figure*}[t!]
    \centering
    \includegraphics[width=\textwidth]{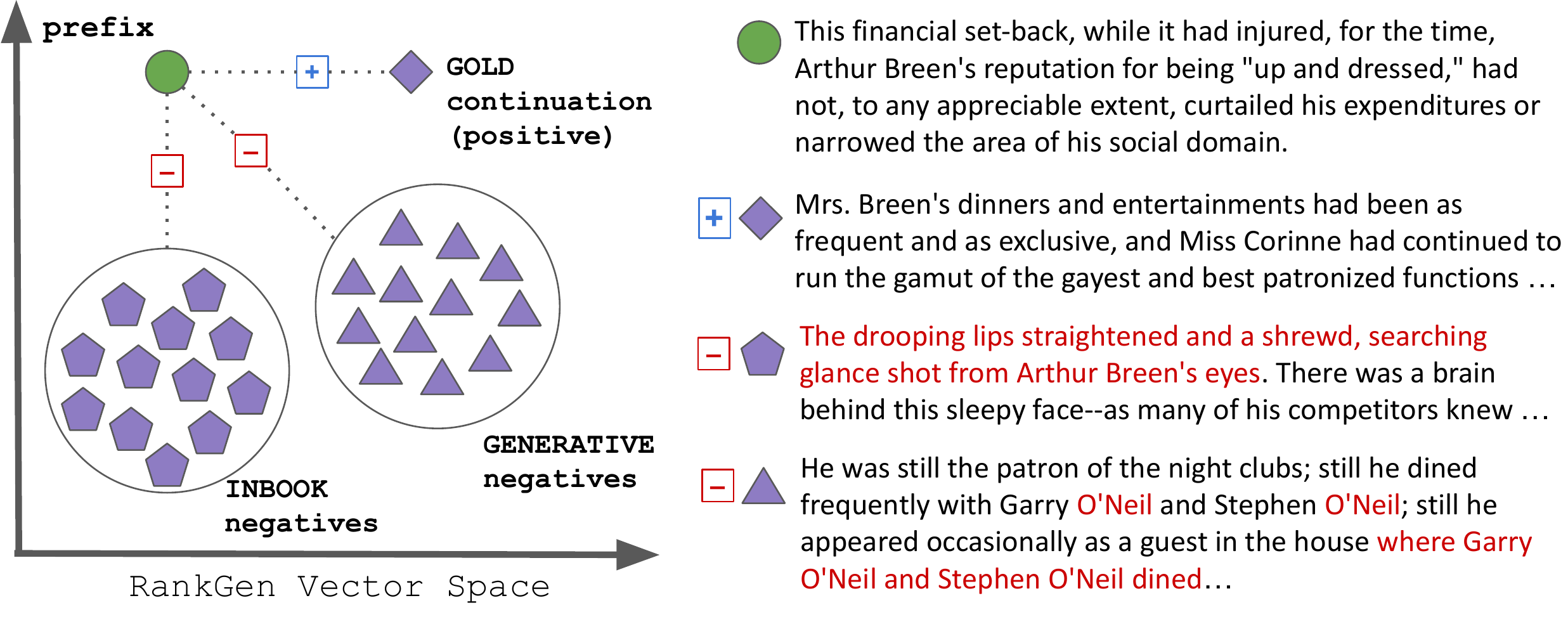}
    \caption{A datapoint from the novel ``\emph{Peter}''~\citep{smith1911peter} used to train \model~with contrastive learning. The \prefixcolor{prefix vector} is pushed towards the \suffixcolor{gold continuation} and away from the vectors of \suffixcolor{several incorrect continuation} with errors (shown in red). These negative samples are either human-written \textbf{\inbookobjective} sequences taken from random locations in the same document (fluent and sometimes topically-similar, but irrelevant and incoherent), or \textbf{\genobjective} samples from a pretrained LM (relevant, but potentially containing hallucination or repetition).}
    \vspace{-0.1in}
    \label{fig:rankgen-train}
\end{figure*}


Part of the problem is that LMs are trained using ``teacher forcing'', where they are always given the ground-truth prefix\footnote{A \emph{prefix} is a sequence of tokens fed as input to an LM, which then generates continuations conditioned on the prefix. A prefix is also called a \emph{prompt} in prior work~\citep{fan-etal-2018-hierarchical}.} and asked to predict the next token. At test-time, however, the prefix can contain model-generated text, allowing errors to propagate during decoding~\citep{bengio2015scheduled}. This issue, combined with the observation that LMs overly rely on \emph{local} context~\citep{khandelwal-etal-2018-sharp, sun-etal-2021-long}, contributes to the generation of sequences that break coherence or consistency within a larger discourse-level context~\citep{wang2022language}. 

To address this issue we present \model, a 1.2 billion parameter English encoder model that maps both human-written \prefix{prefixes} and model-generated continuations of those prefixes (\suffix{generations}) to a shared vector space. \model~efficiently measures the compatibility between a given \prefix{prefix} and \suffix{generations} from any external LM  by ranking the generations via their dot product with the prefix (\figureref{fig:rankgen-infer}). We train \model\ using large-scale contrastive learning, encouraging \prefix{prefixes} to be closer to their \suffix{gold continuation} and far away from incorrect \suffix{negatives}. Since our objective considers two \emph{sequences} rather than just single token prediction, it encourages \model~to consider longer-distance relationships between the prefix and continuation rather than just local context.

We devise two different strategies (shown in \figureref{fig:rankgen-train}) for selecting challenging negative samples, and empirically show that current large LMs cannot distinguish gold continuations from the negatives via perplexity (\sectionref{sec:suffix-identification-ppl}). In the first strategy, \textbf{\inbookobjective}, we select random sequences that occur within the same document as the prefix. While these human-written negatives are fluent and might contain topic or entity overlap, they are irrelevant as continuations to the prefix. In the second strategy, \textbf{\genobjective}, we generate continuations by conditioning a large pretrained LM on a given prefix. Compared to \inbookobjective\ negatives, these negatives are much more relevant to the prefix, but they suffer from issues like hallucination and repetition.



While \model\ can be easily used to rerank full-length samples from any external LM, we demonstrate further improvements in generation quality when it is integrated as a scoring function into beam search. On automatic and human evaluations across four large pretrained models (345M to 11B parameters) and two datasets, we observe that \model~significantly and consistently outperforms sampling-based methods (nucleus, typical, top-$k$) as well as perplexity-based reranking (85.0 vs 77.3 MAUVE, 74.5\% human preference over nucleus sampling\footnote{See \tableref{tab:mauve-scores-full-rerank}, \ref{tab:human-eval-main} for all results. MAUVE~\citep{pillutla2021mauve} is a recently introduced automatic metric for open-ended generation which has high correlation with human judgements.}). Additionally, \model\ outperforms newer decoding algorithms like contrastive decoding and search (89.4 vs 84.9 MAUVE on Wikipedia) which were proposed \emph{after} the initial \model\ release in May 2022. Qualitative analysis from our human annotators (English writers) suggests that most of the improvements stem from increased relevance and continuity between the generated text and the prefix. Finally, we explore applications of \model\ outside of text generation and report state-of-the-art results on two complex literary retrieval benchmarks: RELiC~\citep{relic22} and ChapterBreak~\citep{sun2022chapter}. We open source code, data and model checkpoints.\footnotemark[1]

\section{\model: a generation ranker}

\model~is a deep encoder network that projects prefixes and generations to a shared vector space. Given a prefix vector and a generation vector, we compute a \emph{score} for the generation via the dot product between the two vectors. To ensure that these scores are meaningful, we train \model~using large-scale contrastive learning~\citep{pmlr-v139-radford21a}, pushing the prefix vector close to the gold completion and away from the vectors of negative samples (\figureref{fig:rankgen-train}). We use two types of negative samples for learning the metric space: (1) sequences at random locations in the same document (\inbookobjective), and (2) model generations (\genobjective). This section empirically justifies our negative sample choice (\sectionref{sec:suffix-identification-ppl}) before presenting a precise model formulation (\sectionref{sec:precise-rankgen-formulation}).

\subsection{LMs do not choose gold over negatives}
\label{sec:suffix-identification-ppl}
We explicitly choose our negatives to focus on a weakness of modern LMs which we empirically verify below: LMs often assign high probability to implausible or irrelevant continuations of a prefix.

\paragraph{\inbookobjective~negatives:} Our first type of negative samples are sequences from random locations in the same document as the prefix, whose lengths match those of the ground-truth continuations. As these negatives are written by humans, they are always fluent and coherent, and often topically similar to the prefix (with overlapping entities). However, they are irrelevant as continuations to the prefix, breaking discourse-level continuity and coherence~\citep{hobbs1979coherence, grosz1995centering}.

\begin{table}[t!]
\small
\begin{center}
\begin{tabular}{ lrrrr } 
 \toprule
 \inbookobjective~neg type $\rightarrow$  & \multicolumn{2}{c}{Random} & \multicolumn{2}{c}{Hard} \\
  \cmidrule(lr){2-3} \cmidrule(lr){4-5}\vspace{-0.3cm}\\
  & PG19 & Wiki & PG19 & Wiki\\
 \midrule
 Random & 50.0 & 50.0 & 50.0\phantom{$^\dagger$} & 50.0\phantom{$^\dagger$} \\
 Unigram Overlap & 79.4 & 69.1 & 55.9\phantom{$^\dagger$} & 51.6\phantom{$^\dagger$} \\
 GPT2-medium & 70.4 & 61.9 & 53.1\phantom{$^\dagger$} & 50.1\phantom{$^\dagger$} \\
 GPT2-XL~\shortcite{radford2019language} & 72.9 & 63.3 & 54.6\phantom{$^\dagger$} & 50.6\phantom{$^\dagger$} \\
 T5-base (f.t. PG19) & 73.0 & 64.0 & 54.0\phantom{$^\dagger$} & 50.5\phantom{$^\dagger$} \\
 T5-XXL (f.t. PG19) & 79.6 & 68.6 & 58.5\phantom{$^\dagger$} & 53.1\phantom{$^\dagger$} \\
 T5-XXL-C4~\shortcite{lester-etal-2021-power}  & 76.4 & 66.2 &  57.4\phantom{$^\dagger$} & 52.2\phantom{$^\dagger$} \\
 GPT3 170B*~\shortcite{brown2020language} & 77.3 & 67.0 & 63.2\phantom{$^\dagger$} & 63.2\phantom{$^\dagger$} \\
 \midrule
 \model~(ours) \\
 PG-XL-\inbookobjective & \textbf{99.1} & 92.7  & 77.4\phantom{$^\dagger$} & 72.0\phantom{$^\dagger$} \\
 PG-XL-\genobjective & 80.2 & 68.3 & 52.5\phantom{$^\dagger$} & 53.5\phantom{$^\dagger$} \\
 PG-XL-both & \textbf{99.1} & 92.3 & 78.0\phantom{$^\dagger$} & 71.4\phantom{$^\dagger$} \\
 \modelallboth & 98.7 & \textbf{97.3} & 61.3$^\dagger$ & 77.2$^\dagger$ \\
 \midrule
 Humans & 94.5  & 91.0 & \textbf{82.0}\phantom{$^\dagger$} & \textbf{90.5}\phantom{$^\dagger$}\\
\bottomrule
\end{tabular}
\end{center}
\vspace{-0.1in}
\caption{How often do models prefer the gold continuation to a prefix over an \inbookobjective~negative (text from a different location in same document)? Overall, large LMs (via perplexity) perform poorly compared to both \model\ and humans.  *GPT3 scores use 1000 datapoints; $^\dagger$hard sets adversarially built with this model.}
\vspace{-0.1in}
\label{tab:gold-beats-neg-main}
\end{table}

\paragraph{LMs struggle to distinguish gold continuations from \inbookobjective\ negatives:} Given a \prefix{prefix} of 256 tokens from Wikipedia or a PG19 book~\citep{rae2019compressive}, we measure how often LMs assign higher probability (lower perplexity) to the \suffix{gold 128-token continuation} over a single \inbookobjective~negative.\footnote{We experiment with multiple \inbookobjective~negatives in appendix \S \ref{appendix:gold-beats-neg-extra}. This task is similar to suffix identification tasks like ROCStories~\shortcite{mostafazadeh-etal-2016-corpus}; see \S \ref{sec:suffix-id-results} for experiments on them.} We break all prefixes and continuations at sentence boundaries to make the task less reliant on local syntactic patterns. \tableref{tab:gold-beats-neg-main} shows that even large LMs perform far below human estimates on this task (63.3\% for GPT2-XL vs 91.0\% human on Wiki),\footnote{Human study done on Upwork; details in \appendixref{appendix:human-eval-details}.} and repeating this experiment with ``hard'' negatives selected from a trained \model\ model drops LM performance even further (50.6\% for GPT2-XL vs. 90.5\% human on Wiki).\footnote{See \appendixref{sec:gold-beats-neg-hard} for more details on ``hard negatives''.} We hypothesize that LMs perform poorly because (1) they overly focus on local context instead of long-range dependencies from the prefix~\citep{khandelwal-etal-2018-sharp,sun-etal-2021-long}; and (2) LMs assign high likelihood to words with high frequency in their training data~\citep{holtzman-etal-2021-surface} which may occur in \inbookobjective~but not in the gold continuation. We analyze the latter further in \appendixref{sec:choice-metric-gold-beats-neg} using alternative scoring functions like PMI.

\vspace{-0.1in}

\begin{table}[t!]
\small
\begin{center}
\begin{tabular}{ lrrr } 
 \toprule
 Discriminator & PG19 & Wikipedia & Average \\
 \midrule
 Random & 50.0 & 50.0 & 50.0 \\
 Unigram Overlap & 40.2 & 44.4 & 42.3\\
 GPT2-medium~\shortcite{radford2019language} & 14.7 & 23.3 & 19.0 \\
 GPT2-XL~\shortcite{radford2019language} & 21.5 & 31.5 & 26.5\\ 
 T5-XXL (f.t. PG19) & 32.4 & 33.7 & 33.1\\
 T5-XXL-C4~\shortcite{lester-etal-2021-power} & 19.0 & 39.1 & 29.1 \\
 \midrule
 \model~(ours)   \\
 PG-XL-\genobjective & \textbf{94.7} & \textbf{89.2} & \textbf{91.9} \\
 PG-XL-\inbookobjective & 69.8 & 59.7 & 64.8 \\
 PG-XL-both & 92.0 & 74.9 & 83.5 \\
 \modelallboth & 86.2 & 81.3 & 83.7 \\
\bottomrule
\end{tabular}
\end{center}
\vspace{-0.1in}
\caption{How often do different models prefer the gold continuation to a prefix over a \genobjective~negative (model-generated continuation)? LM perplexity strongly prefers \genobjective~over gold continuations, while \model\ accurately prefers the gold. Negatives were generated from all four LM models in table using nucleus sampling~\shortcite{holtzman2020curious} with $p=0.9$ and then pooled (\appendixref{appendix:gold-beats-gen-extra} breaks down scores by LM).}
\vspace{-0.1in}
\label{tab:gold-beats-generation-main}
\end{table}

\begin{figure*}[t!]
    \centering
    \includegraphics[width=0.96\textwidth]{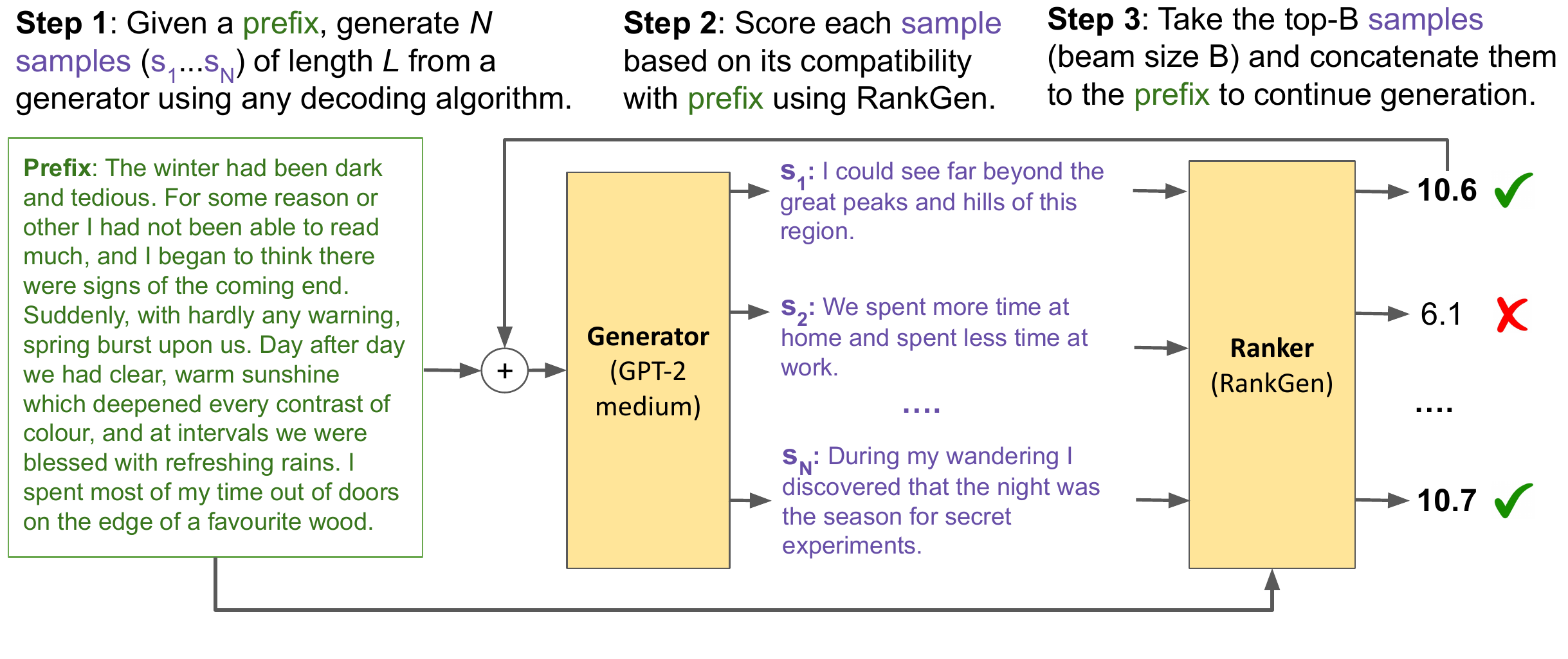}
    \vspace{-0.2in}
    \caption{The \model~setup during inference. \model~can be flexibly plugged into any generative model (like GPT2) using any decoding algorithm (like nucleus sampling) during inference in a beam-search like setup. The examples shown here are actual generations from GPT2-md (with nucleus $p$=0.9) and scores from \model.}
    \vspace{-0.1in}
    \label{fig:rankgen-infer}
\end{figure*}

\vspace{0.05in}

\paragraph{LMs also struggle to distinguish gold continuations from \genobjective~negatives:} Our second type of negative samples are continuations to a prefix that are generated by a pretrained LM. Machine-generated text is known to differ significantly from human text, containing repetitions, hallucinations, and artifacts~\citep{zellers2019defending,maynez-etal-2020-faithfulness,holtzman2020curious}. We use these negatives to encourage \model~to prefer generations closer to the human distribution, similar in spirit to GAN discriminators~\citep{goodfellow2014generative}. \genobjective~negatives have also been used in previous energy-based LMs~\citep{deng2020residual}, although not at this scale; see \sectionref{sec:related-work} for more related work.
In \tableref{tab:gold-beats-generation-main}, we show that LM perplexity is poor at identifying human text over \genobjective~negatives (GPT2-XL gets just 26.5\% accuracy, well below 50\% random chance). This relates to prior work showing LMs have high confidence in machine-generated text~\citep{gehrmann-etal-2019-gltr}, especially their own (\appendixref{appendix:gold-beats-gen-extra}).

\subsection{Training \model}
\label{sec:precise-rankgen-formulation}

Having motivated our negative sampling strategies, we now  describe \model's training process. We train \model~using large-scale contrastive learning with in-batch negative sampling, which is a popular metric learning technique~\citep{sohn2016improved} previously used for dense retrieval (DPR,~\citealp{karpukhin-etal-2020-dense}), image classification (SimCLR,~\citealp{chen2020simple}), and multimodal representation learning (CLIP,~\citealp{pmlr-v139-radford21a}).

A single \model\ training instance consists of a triple $(p_i, c_i, g_i)$, where $p_i$ is a prefix, $c_i$ is the ground-truth continuation of that prefix, and $g_i$ is a continuation generated by an LM. We prepend a special token (\texttt{pre}) to each prefix, and \texttt{suf} (\emph{suffix}) to each continuation and generation. We then pass each element of the triple through a shared Transformer encoder~\citep{vaswani2017attention}, projecting them to fixed-size vectors ($\mathbf{p}_i$, $\mathbf{c}_i$, $\mathbf{g}_i$) using the representation of the special token. To train this model, we use a contrastive objective that pushes the prefix vector \prefix{$\mathbf{p}_i$} close to the gold continuation vector \suffix{$\mathbf{c}_i$}, but away from both the generation vector \suffix{$\mathbf{g}_i$} as well as all other continuation vectors \suffix{$\mathbf{c}_j$} in the same minibatch (``in-batch negative sampling''),
\begin{align*}
    Z(\prefix{\mathbf{p}_i}) &= \sum_{\suffix{c_j} \in B} \exp \prefix{\mathbf{p}_i} \cdot \suffix{\mathbf{c}_j} + \sum_{\suffix{g_j} \in B} \exp \prefix{\mathbf{p}_i} \cdot \suffix{\mathbf{g}_j} \\
    P(\suffix{c_i} | \prefix{p_i}) &= \exp ( \prefix{\mathbf{p}_i} \cdot \suffix{\mathbf{c}_i}) ~/~ Z(\prefix{\mathbf{p}_i})\\
    \text{loss} &= - \sum_{(\prefix{p_i}, \suffix{c_i}) \in \mathcal{B}} \log P(\suffix{c_i} | \prefix{p_i})
\end{align*}

\noindent where $\mathcal{B}$ is a minibatch.  All minibatch elements are sampled from the \emph{same document}, which provides the \inbookobjective\ negatives. Note that the minibatch size $|\mathcal{B}|$ is an important hyperparameter since it determines the number of negative samples; we set $|\mathcal{B}|=1536$ for our XL variant.\footnote{ See \S\ref{sec:rankgen-training-details} for training details and sizes of model variants.}

\vspace{0.1in}

\noindent \textbf{Dataset construction}: We consider all possible 256-word prefixes $p_i$ in our document, ensuring that prefixes begin and end at sentence boundaries. We then select continuations $c_i$ of variable length (10-128 words long) for each prefix $p_i$ so that \model\ can re-rank candidates of different lengths at test-time. To produce \genobjective~negatives, we first use 50\% of our ($p_i$, $c_i$) training data pairs to fine-tune T5-XXL~\citep{raffel2020exploring} for causal language modeling (one per domain). For the remaining half of the dataset, we use this LM to generate a single continuation $g_i$ to the prefix $p_i$ of variable length (10-128 words) using nucleus sampling~\citep{holtzman2020curious} with $p=0.9$.

\subsection{Using \model~at inference}
\label{sec:rankgen-inference}
After model training, the dot product between the prefix and continuation vectors denotes their compatibility score. We experiment with two strategies for using these scores during generation: (1) over-generation and reranking, in which we use any pretrained LM and decoding algorithm to generate multiple samples (20 in our experiments) and then re-rank them; and (2) beam search (\figureref{fig:rankgen-infer}), in which we generate $N$ samples of length $L$ via nucleus or ancestral sampling, compute the top $B$ highest-scoring samples via \model, and concatenate them to the prefix to continue generation. 
There are three hyperparameters for our beam search: (i) the rerank length $L$, or the number of tokens generated before each re-ranking; (ii) the beam size $B$; and (iii) the number of samples generated per beam $N$. Setting $N$=20, $B$=1, $L$=128 (max generation length) is equivalent to the first strategy of over-generation and re-ranking. Details of our implementation and hyperparameter search are in \appendixref{sec:implementations},~\ref{appendix:grid-search}. Overall all tested hyperparameters improve over baselines, but $N$=10, $B$=2, $L$=20 performs best but all tested hyperparameter choices improve over baselines (\figureref{fig:scatter-time-mauve}).


\section{Experiments}
\label{sec:experiments}


\subsection{Model configurations}
\label{sec:models-evaluated}

\paragraph{\model~variants:} We study four configurations of \model, each with 1.2B parameters (XL size) and trained with minibatch size 1536. Three variants are trained on the PG19 dataset~\citep{rae2019compressive}, which consists of long-form books, using (1) only \inbookobjective~negatives, (2) only \genobjective~negatives, and (3) both types of negatives. Since PG-19 contains mainly historical literature, we also experiment with different data sources by training \model\  on the union of four domains (``all'') --- PG19, Wikipedia, C4-NewsLike and C4-WebTextLike~\citep{raffel2020exploring}. This last model is trained using both types of negatives. More ablations varying the model size and minibatch size (number of negatives) are provided in \appendixref{appendix:ablations}.

\vspace{-0.05in}

\paragraph{Pretrained language models:} Does \model\ improve generation quality regardless of the size and pretraining dataset of the LM? To check this we evaluate four different pretrained LMs whose sizes vary considerably from that of \model~(1.2B parameters). We experiment with two variants of GPT-2~\citep{radford2019language}: GPT2-medium (345M) and GPT2-XL (1.5B parameters). We also evaluate a pretrained T5-XXL-v1.1~\citep{raffel2020exploring} model (11B parameters) that we fine-tune to perform language modeling on the training set of PG19~\citep{rae2019compressive}. Finally, to experiment with a large LM trained on out-of-domain data for \model-PG19, we evaluate the T5-XXL model from~\citet{lester-etal-2021-power} (11B parameters) that was fine-tuned for language modeling on the C4 corpus.

\begin{table*}[t!]
\small
\begin{center}
\begin{tabular}{ lrr|rr|rr|rr|r } 
 \toprule
 & \multicolumn{8}{c}{\textbf{Generator Language Model / Prefix Dataset}} \\
 \cmidrule{2-9} \vspace{-0.3cm} \\
 & \multicolumn{2}{c|}{T5-XXL-C4} & \multicolumn{2}{c|}{GPT2-md} & \multicolumn{2}{c|}{GPT2-XL} & \multicolumn{2}{c|}{T5-XXL-PG19}  & Average \\
\textbf{Decoding method} & PG19 & wiki & PG19 & wiki & PG19 & wiki & PG19 & wiki \\
 \midrule
 Greedy decoding & 6.6 & 15.2 & 3.8 & 11.2 & 6.4 & 18.3 & 23.4 & 38.5 & 15.4\\
 Ancestral sampling & 67.7 & 71.6 & 75.5 & 73.2 & 77.4 & 75.0 & 90.2 & 67.7 & 74.8 \\
 Nucleus, $p=0.9$~\citep{holtzman2020curious} & 69.7 & 77.9 & 73.0 & 74.6 & 74.4 & 75.0 & 92.6 & 81.8 & 77.3 \\
 Top-k, $k=40$~\citep{fan-etal-2018-hierarchical} & 68.3 & 77.3 & 74.8 & 73.4 & 76.0 & 75.2 & 92.2 & 81.8 & 77.4 \\
 Typical, $p=0.9$~\citep{meister2022typical} & 69.5 & 77.4  & 73.2 & 73.5 & 73.6 & 76.4 & 92.7 & 81.1 & 77.1 \\
 \midrule
   \multicolumn{3}{l}{\emph{Re-ranking 20 full-length ancestral samples}}\vspace{0.15cm} \\
   \model~PG19-XL-both & 79.9 & 83.3 & 78.8 & 78.5 & 78.2 & 79.6 & 92.2 & 79.2 & 81.2 \\
   \model~all-XL-both & 71.0 & 85.8 & \textbf{79.0} & 84.9 & \textbf{79.0} & 86.4 & 92.1 & 82.9 & 82.6 \\
 \midrule 
 \multicolumn{3}{l}{\emph{Re-ranking 20 full-length nucleus samples}}\vspace{0.15cm} \\
 Unigram overlap & 65.6 & 80.7 & 74.8 & 78.7 & 73.9 & 79.4 & 93.6 & 90.6  & 79.7 \\
 LM perplexity & 62.6 & 55.1 & 55.5 & 63.1 & 58.3 & 61.6 & 88.4 & 77.1 & 65.2 \\
  \model~PG19-XL-\genobjective & 78.3 & 82.4 & 76.2 & 73.8 & 76.2 & 73.0 & \textbf{95.0} & 87.1 & 80.2 \\
  \model~PG19-XL-\inbookobjective & 70.7 & 83.4 & 76.7 & 81.7 & 76.0 & 83.6 & 93.3 & 85.9 & 81.4 \\
  \model~PG19-XL-both & \textbf{80.7} & 86.4 & 76.3 & 79.4 & 75.2 & 81.3 & \textbf{94.3} & 87.3 & 82.6 \\
  \model~all-XL-both &  73.0 & 88.1 & 74.8 & 83.9 & 75.9 & 85.7 & 93.6 & 91.8 & 83.4 \\
  ~~+ beam search ($B$=2, $L$=20, $N$=10) & 74.0 & \textbf{89.4} & 76.2 & \textbf{88.9} & 77.0 & \textbf{89.4} & 92.2 & \textbf{93.0} & \textbf{85.0} \\
\bottomrule
\end{tabular}
\end{center}
\vspace{-0.1in}
\caption{A comparison between \model~variants and baseline decoding algorithms using MAUVE~\citep{pillutla2021mauve}, an automatic text generation metric with high human correlation. \model~significantly outperforms baselines like nucleus \& typical sampling, as well as other re-ranking strategies using LM perplexity and unigram overlap. Incorporating \model\ into beam search (last row) results in the best average MAUVE score. All \model\ rows follow the format, "<training\_data>-<size>-<negatives>", for example "PG19-XL-\inbookobjective".}
\vspace{-0.1in}
\label{tab:mauve-scores-full-rerank}
\end{table*}

\subsection{Open-ended text generation}
\label{sec:eval-setup-metrics}

Following prior work on text generation~\citep{welleck2019non,holtzman2020curious,su2022contrastive}, we primarily focus on open-ended text generation, which has wide applications for tasks such as generating stories~\citep{fan-etal-2018-hierarchical}, poetry~\citep{zhang-lapata-2014-chinese}, and dialog~\citep{miller-etal-2017-parlai} and few-shot NLP~\citep{brown2020language}. We consider \textbf{two domains} in our study: (1) prefixes from Wikipedia, and (2) literary text from PG19~\citep{rae2019compressive}. Since it is difficult to conduct human evaluations of long sequences of machine-generated text~\citep{karpinska-etal-2021-perils}, our main experiments consider a \prefix{256-token prefix} and \suffix{128-token generations}. We analyze generation quality given varying prefix lengths in \sectionref{sec:prefix-suffix-length-vary}.

\vspace{-0.05in}

\paragraph{Decoding algorithms:} For each LM considered we decode outputs using greedy decoding, ancestral sampling, nucleus sampling~\citep{holtzman2020curious}, top-k sampling~\citep{fan-etal-2018-hierarchical}, and typical sampling~\citep{meister2022typical}. Since \model\ is fundamentally a re-ranker of multiple samples, we also compare to two other re-rankers using LM perplexity and unigram overlap, respectively. In all re-ranking settings, we generate 20 samples and then re-rank them with each method. For \model, we also use beam search (\S \ref{sec:rankgen-inference}) that re-ranks partially generated hypotheses.

In addition to these baselines, in \tableref{tab:mauve-newer-methods} we also compare \model\ to newer decoding algorithms proposed after the \model\ release (May 2022).

\vspace{-0.05in}

\paragraph{Automatic \& human evaluation metrics:} We use MAUVE~\citep{pillutla2021mauve} as our primary metric for automatic evaluation. MAUVE computes the similarity of the distribution of human-written text and machine-generated text, and has high correlation with human judgments.\footnote{Details about our MAUVE setup in \appendixref{appendix:mauve-details}. More evaluations with metrics like \textsc{rep}~\shortcite{welleck2019neural} in \appendixref{sec:token-overlap}.}  Since automatic metrics are insufficient for text generation evaluation~\citep{celikyilmaz2020evaluation}, we also conduct a human evaluation by hiring English teachers and writers from Upwork;\footnote{\url{https://www.upwork.com}} see \appendixref{appendix:human-eval-details} for more details. For each of GPT2-medium and T5-XXL-C4 we choose 50 Wikipedia and 50 PG19 prefixes, and show \emph{three} annotators a pair of continuations from different decoding strategies in a random order (blind A/B testing). Annotators are asked to choose the better continuation and provide a 1-3 sentence explanation for their choice. This gives us 600 annotations, analyzed in \S \ref{sec:human-eval-main}, \ref{sec:scarecrow}.

\subsection{Results from automatic evaluations}
\tableref{tab:mauve-scores-full-rerank} contains MAUVE scores for all decoding configurations and datasets. Overall, we see that:
\vspace{-0.05in}

\paragraph{\model~re-ranking and beam search significantly improves MAUVE:} Re-ranking full-length samples with \model~yields an average MAUVE score of 83.4 across all configurations, significantly outperforming other decoding strategies like greedy decoding (15.4), ancestral sampling (74.8), and nucleus / top-k / typical sampling (77.1-77.4). Adding beam search further boosts performance to 85.0.\footnote{Hyperparameter grid search details in \appendixref{appendix:grid-search}.} Surprisingly, re-ranking 20 full-length ancestral samples with \model~performs better than standard nucleus sampling (77.3 vs 82.6). However, re-ranking 20 ancestral samples is slightly worse than re-ranking 20 nucleus samples (82.6 vs 83.4) due to worse inherent quality of ancestral vs nucleus (74.8 vs 77.3).  Re-ranking generations by unigram overlap to the prefix is a surprisingly good baseline (79.7), while re-ranking by LM perplexity reduces MAUVE to 65.2, since it emulates likelihood-based methods like greedy decoding. Finally, \model~performs best on in-domain data, with the PG19-XL-both variant obtaining better scores than the model trained on four domains (80.7 vs 73.0 on T5-XXL-C4, PG19).

\vspace{-0.05in}

\paragraph{\inbookobjective~negatives help more than \genobjective, but using both maximizes MAUVE:} In \tableref{tab:mauve-scores-full-rerank} (bottom), we perform ablations by removing the \inbookobjective~and \genobjective~for \model~PG19 variants. All three variants outperform nucleus sampling (77.3), but keeping both objectives performs best (82.6). A model trained with only \inbookobjective\ is more effective (81.4) than one trained with only \genobjective\ (80.2).

\vspace{-0.05in}

\begin{table}[t]
\small
\begin{center}
\begin{tabular}{ lrrrr } 
 \toprule
  & \multicolumn{2}{c}{GPT2-md} & \multicolumn{2}{c}{GPT2-XL}\\
  \textbf{Decoding Method} & PG19 & wiki & PG19 & wiki \\
 \midrule
Nucleus ($p = 0.9$) & 73.0 & 74.6 & 74.4 & 75.0 \\
Eta~\citep{hewitt2022truncation} & 76.4 & 72.8 & 77.7 & 76.2 \\
\midrule
\multicolumn{3}{l}{\emph{Contrastive methods}}\vspace{0.1cm} \\
~search~\citep{su2022contrastive} & 5.3 & 21.2 & 54.0 & 43.2 \\
~decode~\citep{li2022contrastive} & 65.2 & 83.2 & 73.2 & 84.9 \\
\midrule
\multicolumn{3}{l}{\model-all-XL (ours)}\vspace{0.1cm}  \\
~~~rerank full ancestral & \textbf{79.0} & 84.9 & \textbf{79.0} & 86.4 \\
~~~beam search nucleus & 76.2 & \textbf{88.9} & 77.0 & \textbf{89.4} \\
 \bottomrule
\end{tabular}
\end{center}
\vspace{-0.1in}
\caption{A comparison of \model\ with newer decoding methods proposed after the initial \model\ release (May 2022). \model\ outperforms all methods in terms of MAUVE scores~\citep{pillutla2021mauve}.}
\vspace{-0.1in}
\label{tab:mauve-newer-methods}
\end{table}

\paragraph{\model\ outperforms newer decoding algorithms proposed after \model\ release:} Since the release of \model\ in May 2022, several new decoding algorithms have been proposed including contrastive search~\cite{su2022contrastive,su2022contrastive2}, contrastive decoding~\citep{li2022contrastive}, and eta sampling~\citep{hewitt2022truncation}. In \tableref{tab:mauve-newer-methods}, we compare \model\ to these newer methods\footnote{We use the official implementations for all these methods. Links - \href{https://huggingface.co/docs/transformers/v4.24.0/en/main_classes/text_generation\#transformers.generation_utils.GenerationMixin.contrastive_search}{contrastive search}, \href{https://github.com/XiangLi1999/ContrastiveDecoding}{contrastive decoding}, \href{https://github.com/john-hewitt/truncation-sampling}{eta sampling}.} on GPT2-md and GPT2-XL. Overall, we find that \model\ significantly outperforms all newly proposed decoding algorithms (89.4 vs 84.9 on GPT2-XL wikipedia against the best baseline contrastive decoding).


\subsection{Human evaluation with A/B tests}
\label{sec:human-eval-main}
Despite the high human correlation of MAUVE, human evaluation remains critical for open-ended generation~\citep{celikyilmaz2020evaluation,gehrmann2022repairing}. Since human evaluation is expensive, we focus on comparing our best performing \model\ variant (\model-XL-all with beam search) to nucleus sampling, one of the most popular decoding algorithms in use today. We conduct blind A/B testing comparing the two methods, hiring English teachers and writers on Upwork (\S \ref{sec:eval-setup-metrics}). \tableref{tab:human-eval-main} shows that humans significantly prefer outputs from \model~over nucleus sampling (74.5\% preference by majority vote, $p < 0.001$). \model\ preference is higher with more inter-annotator agreement (\tableref{tab:human-agreement}) for outputs from the smaller GPT2-medium. Finally, humans show slightly higher \model\ preference for Wikipedia generations compared to PG19.

\begin{table}[t]
\small
\begin{center}
\begin{tabular}{ lrrrr } 
 \toprule
& PG19 & Wikipedia & Overall \\
 \midrule
GPT2-md & 80.0 {\scriptsize (72.0)} & 82.0 {\scriptsize (78.3)}  & 81.0 {\scriptsize (75.1)} \\
T5-XXL-C4 & 68.0 {\scriptsize (63.3)} & 68.0 {\scriptsize (65.3)} & 68.0 {\scriptsize (64.3)} \\
Overall & 74.0 {\scriptsize (67.8)} & 75.0 {\scriptsize (71.9)} & 74.5 {\scriptsize (69.8)} \\
 \bottomrule
\end{tabular}
\end{center}
\vspace{-0.1in}
\caption{Percentage of instances for which English writers prefer \model~outputs over nucleus samples in a blind A/B test. Scores shown are majority vote, with mean accuracy in subscript.  Humans significantly prefer \model\ ($p$ < $10^{-3}$); agreement stats in \tableref{tab:human-agreement}.}
\vspace{-0.1in}
\label{tab:human-eval-main}
\end{table}

\begin{table}[t]
\small
\begin{center}
\begin{tabular}{ lrrrr } 
 \toprule
 & PG19 & Wikipedia & Overall \\
 \midrule
GPT2-md & 0.31, 48\% & 0.49, 60\% & 0.40, 54\% \\
T5-XXL-C4 & 0.27, 46\% & 0.30, 48\% & 0.29, 47\% \\
Overall & 0.29, 47\% & 0.40, 54\% & 0.35, 51\% \\
 \bottomrule
\end{tabular}
\end{center}
\vspace{-0.1in}
\caption{Inter-annotator agreement for the human evaluation in \tableref{tab:human-eval-main} using Fleiss $\kappa$~\shortcite{fleiss1971measuring}, and \% of pairs with unanimous agreement among 3 annotators. Overall we see moderate agreement, higher for Wiki, GPT2.}
\vspace{-0.1in}
\label{tab:human-agreement}
\end{table}
\section{Analysis}
\subsection{Types of generation improvements}
\label{sec:scarecrow}
To get more insight into the human preference judgments made in~\sectionref{sec:human-eval-main}, we asked our annotators to provide a 1-3 sentence free-form explanation for each of their choices.\footnote{All 600 human explanations are provided in submission.} We manually categorized each of 600 explanations into nine broad categories loosely based on the \textsc{Scarecrow} schema designed by~\citet{dou-etal-2022-gpt}. In \tableref{tab:scarecrow-summary} we see that 81\% of the explanations preferring \model\ mentioned some aspect of the relationship between the prefix and the generated text, including relevance, continuity, and stylistic similarity. 8.0\% of the explanations said that \model~outputs displayed fewer commonsense errors, while 4.7\% said that they were less repetitive. We show some generations and human explanations in \tableref{tab:model-outputs-with-scarecrow} and several more full-length generations in \appendixref{sec:more-generations}.

\begin{table}[t]
\small
\begin{center}
\begin{tabular}{ p{5.5cm}r } 
 \toprule
\multicolumn{2}{l}{\emph{Reasons relating the prefix with the generation} (81\%)} \\
\midrule
More topically relevant to the prefix & 37.7\% \\
Better continuity / flow / chronology & 31.6\%  \\
Does not contradict prefix & 6.8\% \\
Stylistically closer to prefix & 4.7\% \\
 \midrule
 \multicolumn{2}{l}{\emph{Reasons related only to the generated text} (19\%)} \\
 \midrule
Better commonsense understanding & 8.0\% \\
Less repetitive & 4.7\% \\
More grammatical & 3.1\% \\
Less contradictions & 1.7\%  \\
More coherent / other & 1.7\%  \\
 \bottomrule
\end{tabular}
\end{center}
\vspace{-0.1in}
\caption{Distribution of reasons given by our human evaluators (English writers/teachers) for preferring \model~outputs over nucleus samples. Relevance / continuity to prefix was a common explanation.}
\vspace{-0.1in}
\label{tab:scarecrow-summary}
\end{table}


\begin{table*}[t!]
\footnotesize
\begin{center}
\begin{tabular}{ p{5.4cm}p{5.2cm}p{4.3cm} } 
\toprule
\bf Prefix & \bf Generations & \bf Annotator Preference \\
\midrule
 \textbf{PG19}, \href{https://www.gutenberg.org/files/2547/2547-0.txt}{\emph{Half a Life-time Ago}}, by \emph{Elizabeth Gaskell}: ... If \textcolor{mycolor1}{thou} doesn’t choose to marry me on those terms--\textcolor{mycolor1}{why!} I can snap my fingers at \textcolor{mycolor1}{thee}, never fear. I’m not so far gone in love as that. But I will not have thee, if \textcolor{mycolor1}{thou say’st} in such a hectoring way that Willie must go out of the house--and the house his own too--before \textcolor{mycolor1}{thou’lt} set foot in it. ...  “\textcolor{mycolor1}{Thou hast} may-be spoken a word too much,” said Michael, \textcolor{mycolor2}{pale with rage}. & Text 1 (\textbf{Nucleus}): “How on earth could it be? He must be part of the marriage, \textcolor{mycolor2}{my love}. But he can’t go away--he can’t go away with me. And if he does go away, he’ll have to pay ... \newline Text 2 (\textbf{\model}): “Take heed how \textcolor{mycolor1}{thou handlest thy} words in the future.” “I haven’t spoken a word too much,” she murmured, coolly. “I meant--I meant the last word!” “Swear it, \textcolor{mycolor1}{thou hastna} sworn a word yet...  & \textbf{Text 2}. Text 1 has a completely different \textcolor{mycolor1}{style}, way more modern. First sentence in Text 2 fits the tone of enraged Michael. \newline \textbf{Text 2}. The writing style is more similar, Text 1 sounds too modern. Plus, the atmosphere of the fragment is more consistent. The characters \textcolor{mycolor2}{seem to be having an argument, so "My love"} in doesn't make sense. \\
\midrule
\textbf{Wiki}, \href{https://en.wikipedia.org/wiki/American_Indian_Movement}{\emph{American Indian Movement}}: ... named \textcolor{mycolor2}{three lower-level AIM members involved in her death}: Arlo Looking Cloud, John Graham, and Theda Nelson Clarke. This was the first time that an AIM leader active at the time of the Aquash murder had publicly accused AIM of having been involved. Looking Cloud and Graham were convicted of murder in 2004 and 2010, by ... By then Clark was being cared for in a nursing home and \textcolor{mycolor1}{was \textcolor{mycolor2}{not} indicted}. & Text 1 (\textbf{Nucleus}): Clarke was the \textcolor{mycolor2}{second AIM leader to be charged}. On July 19, 2009, her defense attorney, Michael Kranz had filed a motion ... His request for a new trial failed in December 2009. In 2009, the ...  \newline Text 2 (\textbf{\model}): \textcolor{mycolor1}{Clarke has also denied any involvement in Aquash's murder.} In the early months of 2001 the FBI began an effort to break through AIM's defenses, to try to identify and bring charges against all three AIM members...  & \textbf{Text 2} - The last sentence of the prefix paragraph ("By then Clark was being cared for in a nursing home and was not indicted") \textcolor{mycolor1}{flows well with Text 2, implying that Clarke was absolved of guilt.} \newline \textbf{Text 2}. Text 2 further goes into Clark's involvement in the case and Aquash's murder while \textcolor{mycolor2}{Text 1 contradicts} part of the prefix. \\
\bottomrule
\end{tabular}
\end{center}
\vspace{-0.1in}
\caption{Representative model outputs using \model~vs nucleus sampling~\citep{holtzman2020curious}, along with human explanations (from English teachers/writers) for preferring \model. For every row the \textcolor{mycolor2}{color coding} grounds the annotator explanation in the prefix and generation. See \appendixref{sec:more-generations} for more \emph{full-length} generations.}
\label{tab:model-outputs-with-scarecrow}
\vspace{-0.15in}
\end{table*}

\subsection{How fast is decoding with \model?}
\label{sec:timing-analysis-main}

\begin{figure}[t!]
    \centering
    \includegraphics[width=0.47\textwidth]{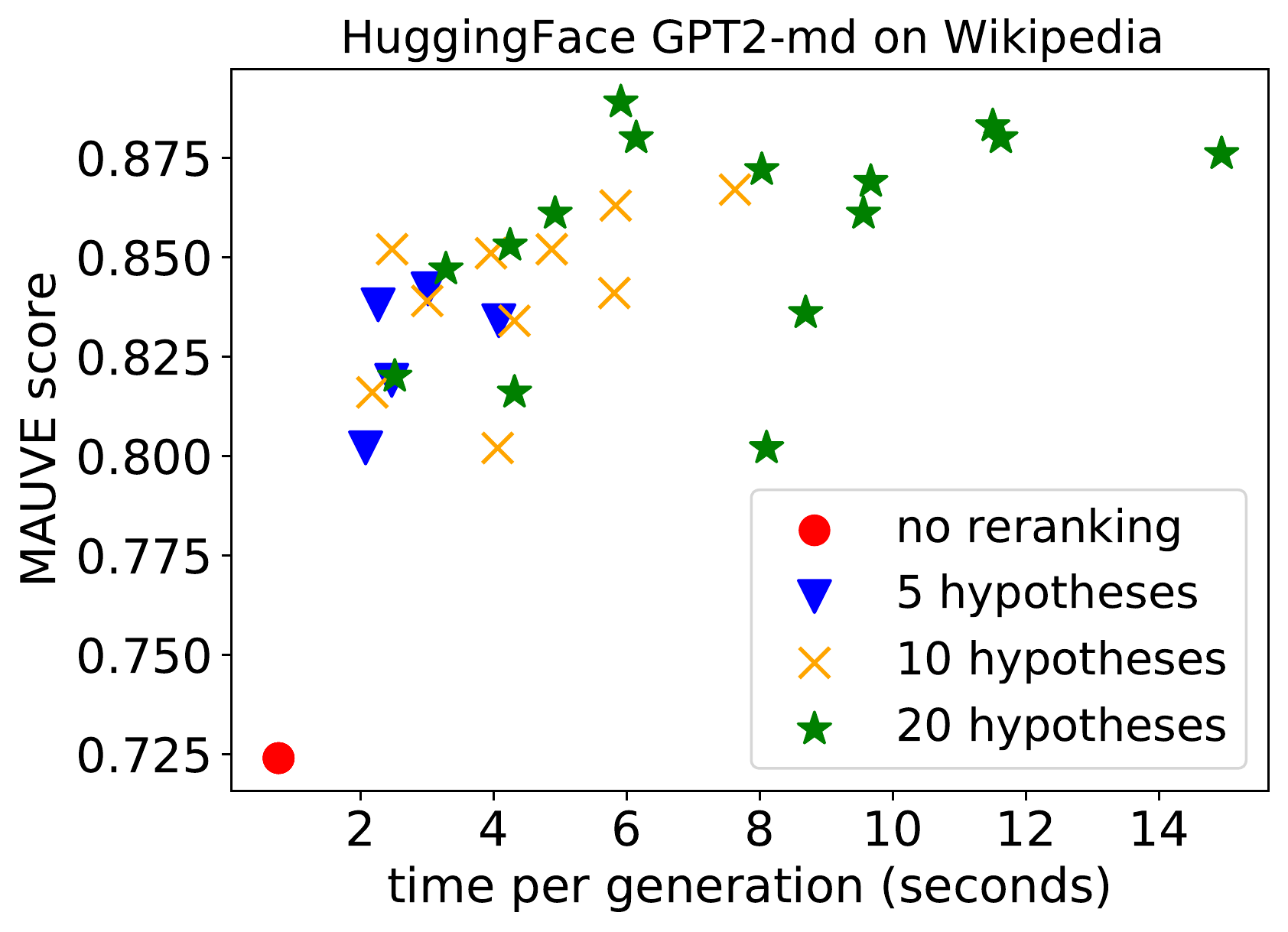}
    \vspace{-0.1in}
    \caption{Performance/time trade-off across hyperparameters (grid search details in \S \ref{appendix:grid-search}). \model\ re-ranking significantly improves MAUVE, but need an order of magnitude more time due to overgeneration.}
    \vspace{-0.1in}
    \label{fig:scatter-time-mauve}
\end{figure}

Our algorithm requires over-generation followed by \model\ re-ranking. How much extra decoding time does this add? In \figureref{fig:scatter-time-mauve}, we show the trade-off between MAUVE score and decoding time across different hyperparameters.\footnote{Timing depends on library / hardware. We analyze HuggingFace on RTX3090, T5X on TPU-v3 in appendix \ref{sec:implementations}.} While decoding a single nucleus sample takes just 0.8 seconds, generating 20 samples followed by re-ranking with \model~requires 2.5 seconds. The best-performing hyperparameters use multiple re-ranking steps, taking 5.9 seconds.\footnote{See \appendixref{appendix:speed-grid-search} for more speed tradeoff plots.} 
In \appendixref{appendix:speed-grid-search}, we see that over-generation is the bottleneck, since re-ranking takes only a fraction of the time (1-10\%) compared to generation.  Developing methods that avoid over-generation (e.g., via distillation) is an exciting future work direction.

\subsection{Generation with different length prefixes}
\label{sec:prefix-suffix-length-vary}

\begin{figure}[t!]
    \centering
    \includegraphics[width=0.47\textwidth]{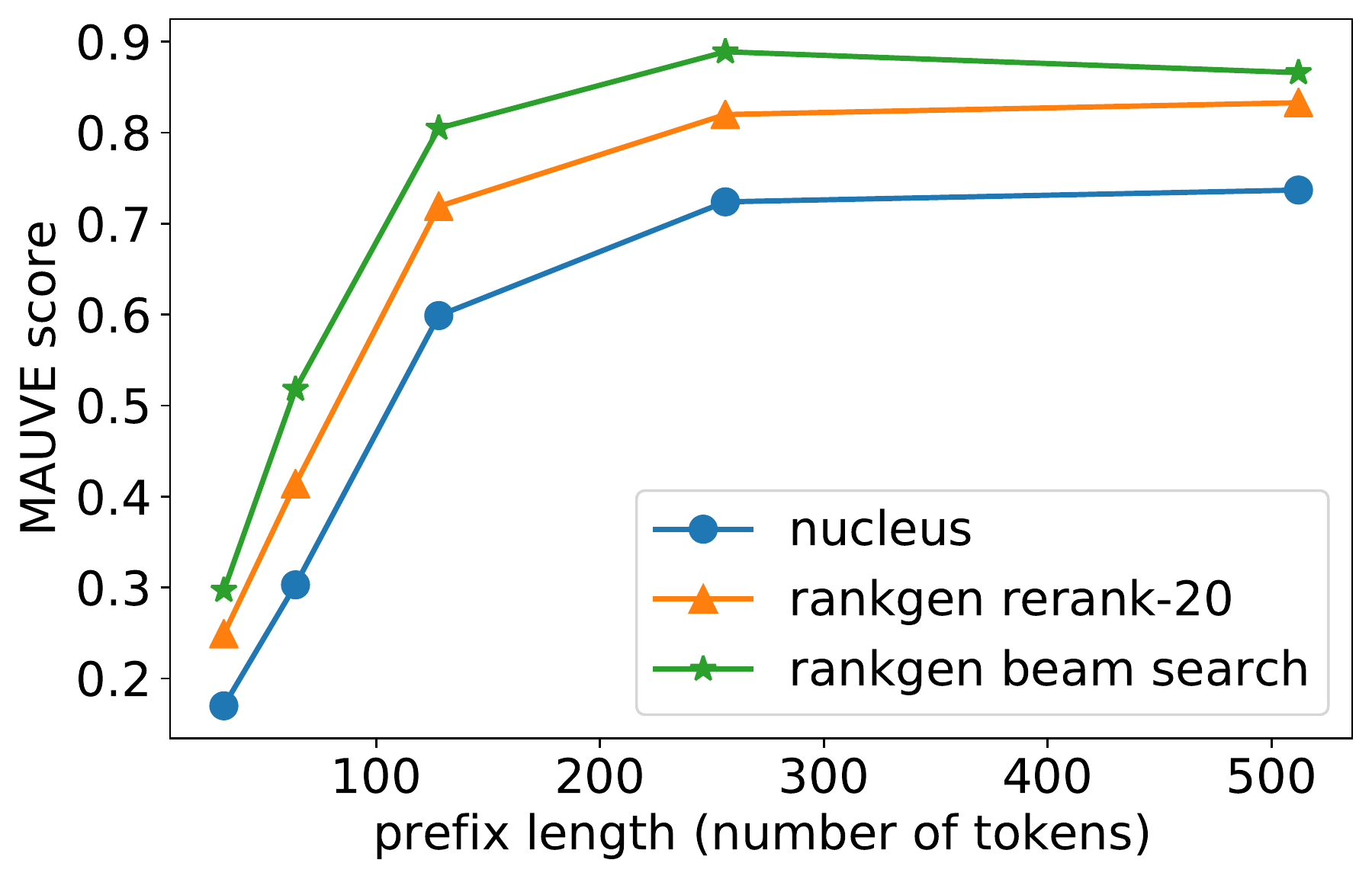}
    \vspace{-0.1in}
    \caption{MAUVE score variation with change in prefix length for GPT2-medium on Wikipedia. Across prefix lengths re-ranking with \model-XL-all boosts performance, and using it in beam search does best.}
    \vspace{-0.1in}
    \label{fig:prefix-length-mauve}
\end{figure}

Our \model\ model is trained with a fixed prefix length of 256 tokens, and all of the evaluations in \sectionref{sec:experiments} also assume a prefix length of 256 tokens. However, many text generation applications take shorter prefixes as input, like short writing prompts in story generation~\citep{fan-etal-2018-hierarchical}. How well does \model~generalize to shorter and longer prefixes? \figureref{fig:prefix-length-mauve} compares nucleus sampling to \model~ across varying prefix lengths. We observe that \model\ consistently outperforms nucleus sampling in terms of MAUVE, and beam search with \model\ always provides further gains, suggesting robustness to the prefix length.

\subsection{\model~as a retriever}
\label{sec:retrieval-results}

While we designed \model\ for text generation, we find that it is also an effective zero-shot retriever.
\model~follows a dual encoder architecture similar to those of several recent dense retrievers like DPR~\citep{karpukhin-etal-2020-dense} and REALM~\citep{pmlr-v119-guu20a}.
We test \model\ on RELiC~\citep{relic22}, a complex literary retrieval task. Given a literary analysis excerpt, systems must retrieve a quote from a book which is most relevant to the excerpt. RELiC requires a deep understanding of literary phenomena (like irony, metaphors, co-reference, style), and current retrievers struggle on it. We test models in a \textbf{zero-shot} setting, without finetuning on RELiC training data. 
In \tableref{tab:relic-results} we find \textbf{\model\  significantly outperforms other retrievers}, achieving a new state of the art on RELiC.\footnote{\url{https://relic.cs.umass.edu/leaderboard.html}} PG-XL-\inbookobjective~performs best (6.0 vs 2.9 recall@1 against the next-best ColBERT), approaching a fully supervised upperbound (9.4). While our XL model has many more parameters than baselines, even PG-base-both~outperforms all baselines~(3.8 vs 2.9), which has a similar number of parameters as our baselines. Dropping \inbookobjective~leads to poor performance (0.7), further confirming its efficacy. Besides RELiC, we investigate retrieval over PG19 books in appendix \S\ref{appendix:gold-beats-neg-extra}, and suffix identification in \S\ref{sec:suffix-id-results}, achieving state-of-the-art on ChapterBreak~\citep{sun2022chapter}.

\begin{table}[t!]
\small
\begin{center}
\begin{tabular}{ lrrrrrr } 
 \toprule
 \bf Model &  \multicolumn{5}{c}{\bf Recall@\emph{k} ($\uparrow$)} \\
 \cmidrule{2-6}\vspace{-0.3cm}\\
  & 1 & 3 & 5 & 10 & 50 \\
 \midrule
 BM25~\shortcite{robertson1995okapi} & 1.3 & 2.9 & 4.1 & 6.7 & 14.5 \\
 SIM~\shortcite{wieting-etal-2019-beyond} & 1.3 & 2.8 & 3.8 & 5.6 & 13.4 \\
 DPR~\shortcite{karpukhin-etal-2020-dense} & 1.3 & 3.0 & 4.3 & 6.6 & 15.4 \\
 c-REALM~\shortcite{krishna-etal-2021-hurdles} & 1.6 & 3.5 & 4.8 & 7.1 & 15.9 \\
 ColBERT~\shortcite{khattab2020colbert} & 2.9 & 6.0 & 7.8 & 11.0 & 21.4 \\
 \midrule
 \model~(ours) \\
 PG-XL-\textsc{gen} & 0.7 & 1.9 & 2.7 & 4.1 & 9.1 \\
PG-XL-\inbookobjective & \textbf{6.0} & \textbf{12.2} & \textbf{15.4} & \textbf{20.7} & \textbf{37.3} \\
 PG-base-both & 3.8 & 8.2 & 10.8 & 15.4 & 31.6  \\
 PG-XL-both & 4.5 & 8.4 & 11.0 & 15.1 & 27.9 \\
 \modelallboth & 4.9 & 9.2 & 11.9 & 16.5 & 31.5 \\
 \midrule
 full supervision ($\uparrow$) &  9.4 & 18.3 & 24.0 & 32.4 & 51.3  \\
 \bottomrule
\end{tabular}
\end{center}
\vspace{-0.1in}
\caption{Performance on RELiC~\shortcite{relic22} compared to other retrievers. We achieve state-of-the-art on the \emph{zero-shot} setting, nearing the supervised upperbound ($\uparrow$).}
\vspace{-0.2in}
\label{tab:relic-results}
\end{table}

\section{Related Work}
\label{sec:related-work}
\label{sec:more-related-work}

Our work on \model\ draws inspiration from previous research on self-supervised learning, energy-based models, and modeling non-local dependencies. For instance, our \inbookobjective~negative sampling is related to popular \textbf{self-supervised representation learning} methods that leverage discourse information across multiple sentences, which is useful for learning sentence embeddings~\citep{kiros2015skip, hill-etal-2016-learning, jernite2017discourse}. Our formulation is most similar to QuickThought~\citep{logeswaran2018efficient}, which uses in-batch negative sampling on a contiguous set of sentences. More recently, the next sentence prediction task has been used for pretraining large LMs~\citep{devlin-etal-2019-bert, lan2020albert, aroca-ouellette-rudzicz-2020-losses}. Unlike these works, we focus specifically on text generation rather than self-supervised pretraining for natural language understanding tasks. 

\model\ is also closely related to efforts in \textbf{energy-based methods}~\citep{lecun2006tutorial} for generative modeling~\citep{grover2019bias,parshakova-etal-2019-global}, speech recognition~\citep{wang2018learning}, open-ended text generation~\citep{bakhtin2019real, deng2020residual}, machine translation~\citep{shen-etal-2004-discriminative,lee-etal-2021-discriminative,bhattacharyya-etal-2021-energy}, constrained generation~\citep{qin2022cold,mixmatch2022}, and models for specific attributes like style~\citep{dathathri2020plug,yang-klein-2021-fudge}, length~\citep{li2017learning}, or repetition \& relevance~\citep{holtzman-etal-2018-learning}. Unlike prior work, we use human-written text from the same document as negative samples (\inbookobjective) in addition to machine-generated text. \model~is also trained at a much larger scale than prior energy-based models for text (1.2B parameters, contrastive learning with 3K negatives on 4 domains).

Finally, \model\ is related to efforts in \textbf{modeling non-local dependencies} in generation, which include methods that predict multiple tokens~\citep{oord2018representation,qi-etal-2020-prophetnet}, rely on retrieval~\citep{khandelwal2020generalization}, use bidirectional LMs~\citep{serdyuk2018twin}, employ contrastive learning~\citep{su2022contrastive,an2022cont}, use BERT for sentence-level language modeling~\citep{ippolito-etal-2020-toward}, and designing sequence-level losses~\cite{wiseman-rush-2016-sequence,edunov-etal-2018-classical,welleck2019neural,liu2022brio} for reducing exposure bias~\citep{bengio2015scheduled,ranzato2016sequence}. While the \model~approach is significantly different from these prior works, it can be intuitively viewed as a ``$k$-word sequence-level'' language modeling approach, which is discriminative rather than generative.

\vspace{0.05in}
\section{Conclusion and Future Work}
\label{sec:future-work}

 We present \model, a large encoder which scores \suffix{continuations} given a \prefix{prefix} and can be plugged into any text generation system. \model~significantly outperforms popular decoding methods on both automatic and human evaluations. We note several exciting future directions for \model, including:
 
 \begin{itemize}
    \setlength\itemsep{0.1em}
     \item training (or adapting) a multilingual variant of \model, as our current models are trained on English text only
     \item training larger \model~models (T5-XXL size or bigger), with longer prefix / suffix lengths, to see if generation quality continues to improve with scale
     \item exploring the utility of \model~in other generation tasks like dialog generation, summarization, or long-form question answering
     \item \model~re-ranking of significantly larger hypothesis sets generated using search algorithms like that in~\citet{xu2021massive}
     \item more directly incorporating \model~into generative modeling to eliminate the need for over-generation, either via gradient-based sampling~\citep{qin2022cold}, distilling \model~knowledge into LMs via unlikelihood training~\citep{welleck2019neural} or reward modeling with RL~\citep{ouyang2022training}
     \item using \model~as a retriever in knowledge retrieval augmented generation~\citep{nakano2021webgpt,komeili2021internet}
     \item further exploring the capability of \model~as a retriever, either zero-shot or by fine-tuning on retrieval benchmarks like BEIR~\citep{thakur2021beir}
     \item utilizing of \model~as a text generation evaluation metric like CARP~\citep{matiana2021cut} or CLIPScore~\citep{hessel-etal-2021-clipscore}
     \item using RankGen on other domains with sequential data, like code completion, protein synthesis, or generating mathematical proofs.
 \end{itemize}

\section*{Limitations}

An important limitation of \model~compared to other decoding methods is the need for over-generation, which we discuss in \sectionref{sec:timing-analysis-main}. While \model~itself is efficient, generating multiple samples increases decoding time by an order of magnitude. \model~is a re-ranking method, so it relies on other decoding methods to produce the \suffix{candidate output set}. Biases in the output candidate set from existing decoding algorithms may be present in \model~outputs. Besides this, \model~may be vulnerable to adversarial examples~\citep{szegedy2013intriguing} --- gibberish text which gets high \model~score, obtained by white-box attacks~\citep{ebrahimi-etal-2018-hotflip,wallace-etal-2019-universal}.

This study is limited to open-ended text generation, which has a large space of possible outputs. \model~or our findings may not be directly applicable to other generation tasks which have a more constrained output space like summarization, long-form QA or machine translation.
\section*{Acknowledgements}

We are very grateful to the freelancers on Upwork and volunteers who helped us evaluate generated text. We thank Xavier Garcia and the T5X team for helping us with technical issues related to the T5X library. We are grateful to William Cohen, Elizabeth Clark, Marzena Karpinska, Tu Vu, Simeng Sun, Ari Holtzman, Slav Petrov, Ciprian Chelba, Nader Akoury, Neha Kennard, Dung Thai, the UMass NLP group and the Google AI language research group in Pittsburgh for several useful discussions during the course of the project. This work was mostly done while Kalpesh Krishna (KK) was a student researcher at Google Research hosted by John Wieting. KK was partly supported by a Google PhD Fellowship awarded in 2021.
\section*{Ethical Considerations}

Current text generation technology produces fluent outputs but suffer from several issues like factual inaccuracies, lack of faithfulness to the input prefix, commonsense issues etc., which makes their real-world deployment difficult. \model~is an effort at rectifying some of these issues, with a focus on faithfulness to input prompts. However, \model~outputs continue to be factually inaccurate at times, as noted by some of our human annotators. This should be strongly considered before any direct deployment of this system. To tackle this issue, using \model~for retrieval augmented generation~\citep{nakano2021webgpt} is a promising direction for future work. We have also open-sourced all 600 human annotations, which have detailed explanations highlighting the strengths / weaknesses of \model~compared to nucleus sampling.

Our final XL-sized models were trained using a Google Cloud TPUv3 Pod slice with 128 chips for a total of 2 days per model. Several similarly-sized models were trained during the development of this project, roughly one XL-size model every week from October 2021 to February 2022. Due to expensive training costs, we have open-sourced our model checkpoints for the community to use and build upon. Note that ``TPUs are highly efficient chips which have been specifically designed for machine learning applications'' as mentioned in the Google 2020 environment report. These accelerators run on Google Cloud, which is ``carbon neutral today, but aiming higher: our goal is to run on carbon-free energy, 24/7, at all of our data centers by 2030.'' (\url{https://cloud.google.com/sustainability}). More details on model size and training are provided in \appendixref{sec:rankgen-training-details}.

\bibliography{anthology,custom,bib/journal-full,bib/arr2022,bib/prefixsuffix}
\bibliographystyle{acl_natbib}
\newpage
\appendix

\section*{Appendices accompanying ``\titlestr''}

\section{More \model~details}

\subsection{\model~training details}
\label{sec:rankgen-training-details}

We fine-tune the encoder of the T5 v1.1 models from ~\citet{raffel2020exploring} using large minibatches (see \tableref{tab:rankgen-model-size-table} for sizes) on a Cloud TPU v3 Pod slice with 128 chips. Our models are implemented in JAX~\citep{jax2018github} using the T5X library~\citep{roberts2022t5x}. Each model was fine-tuned for 100k steps, using a constant learning rate of 0.002 using the Adafactor optimizer~\citep{shazeer2018adafactor}.

\begin{table}[h]
\begin{center}
\begin{tabular}{ lrr } 
 \toprule
  Model & Batch Size & Parameters \\
 \midrule
\model-base & 4096 & 110.2M\\
\model-large & 4096 & 342.3M\\
\model-XL & 1536 & 1.2B\\
 \bottomrule
\end{tabular}
\end{center}
\caption{Minibatch size and number of trainable parameters across different \model~variants. See \appendixref{appendix:ablations} for ablation studies justifying scale.}
\label{tab:rankgen-model-size-table}
\end{table}

\subsection{Implementation and timing details}
\label{sec:implementations}

In \figureref{fig:python-code-beam-search} we provided a simplified Python implementation (without minibatching) of our \model~beam search algorithm. We implement this algorithm in two libraries --- the first uses PyTorch with the popular HuggingFace Transformers library~\citep{wolf-etal-2020-transformers}, which we test on a RTX 3090 GPU with 25GB memory. The second uses JAX~\citep{jax2018github} with the T5X library~\citep{roberts2022t5x}, and is tested on a single Cloud TPU v3 board with 32GB memory.\footnote{\url{https://cloud.google.com/tpu/docs/system-architecture-tpu-vm\#single\_tpu\_board}} While measuring decoding time for various hyperparameters (\appendixref{appendix:speed-grid-search}), we focus on \emph{throughput}~\citep{dehghani2021efficiency}, measuring wall-clock time after minibatching to the extent the hardware permits. We ensure consistent experimental settings across hyperparameters, using the same machine and making sure no other computationally expensive process is running on it.

\subsection{\model~hyperparameter grid search}
\label{appendix:grid-search}

Our hyperparameter grid search is conducted on Wikipedia data with the smallest model considered (GPT2-medium), using MAUVE as our hill-climbing criteria. Our \model~algorithm has three main hyperparameters --- rerank length $L$, beam size $B$ and number of samples per beam $N$. The rerank length denotes the number of new tokens which are generated before a re-ranking step takes place. Number of samples denotes the number of generated sequences for each beam. The number of samples retained across different re-ranking cycles is the beam size (see \figureref{fig:python-code-beam-search} for exact implementation). Our \model~grid search is conducted over the following configurations ---

\noindent \textbf{rerank length} $L$: 5, 10, 20, 50, max\_length tokens \\
\noindent \textbf{number of samples} (beam size $B$ * number of samples in every beam $N$):

\noindent 1 sample --- (1 * 1);\\
\noindent 5 samples --- (1 * 5);\\
\noindent 10 samples --- (1 * 10); (2 * 5); \\
\noindent 20 samples --- (1 * 20); (2 * 10); (4 * 5); \\
\noindent 40 samples --- (1 * 40); (2 * 20); \\

Additionally, we measure the extent to which full-length reranking works ($L$ = max length, $B$ = 1) by simply increasing the number of samples $N$ over-generated and then for re-ranking.

\subsubsection{MAUVE score tradeoffs}
\label{appendix:performance-grid-search}

\begin{figure*}[t!]
    \centering
    \includegraphics[width=0.51\textwidth,valign=t]{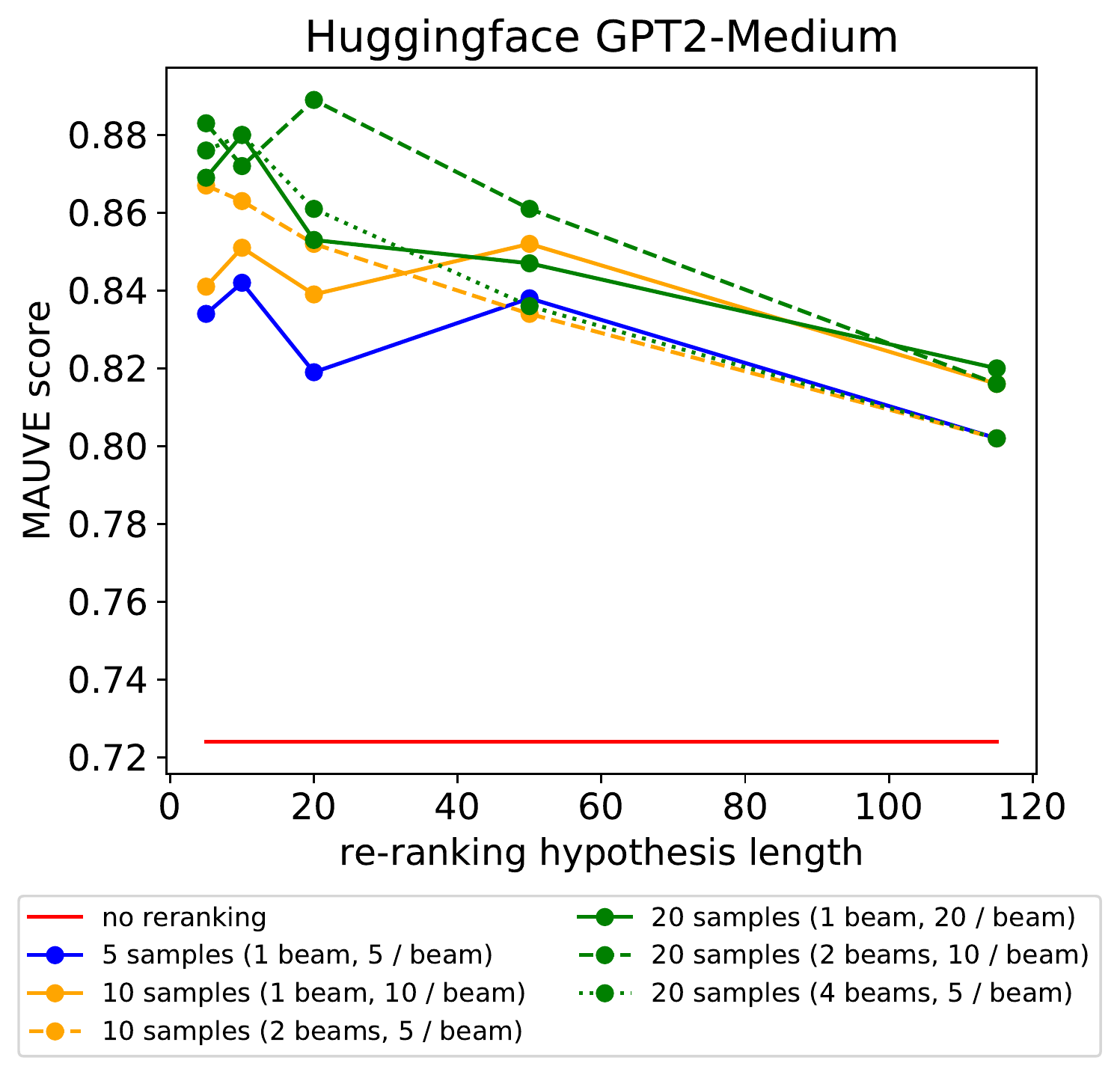}
    \includegraphics[width=0.48\textwidth,valign=t]{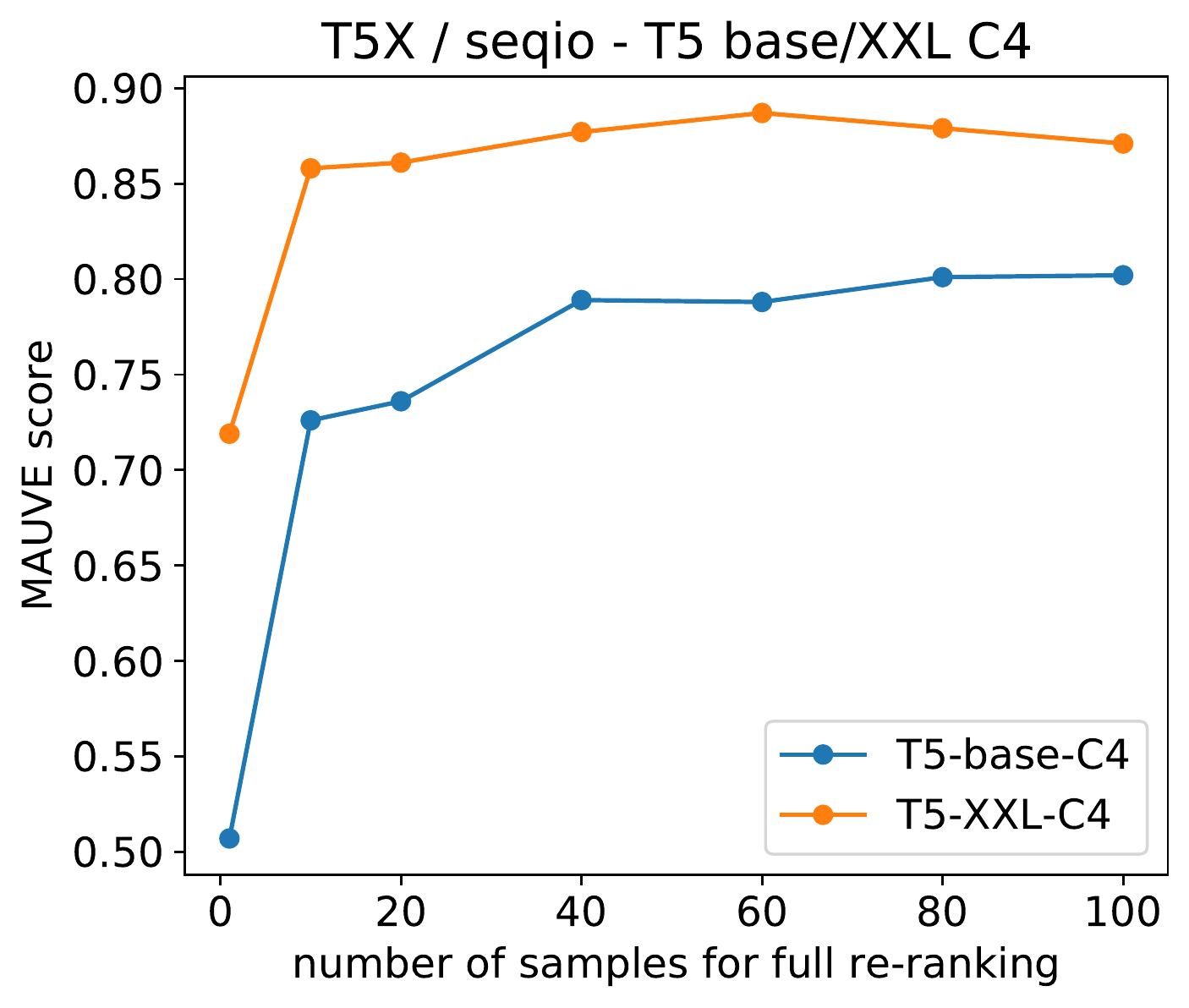}
    \vspace{-0.1in}
    \caption{Variation in MAUVE score across different \model~hyperparameters on Wikipedia data (\appendixref{appendix:performance-grid-search}). \textbf{Left}: Experiments on GPT2-medium show that \model~improvements are robust to hyperparameter choice, re-ranking shorter hypotheses improves performances over full re-ranking, re-ranking more samples improves performance. \textbf{Right}: Full re-ranking performance generally improves with more samples, but this improvement saturates after a point, especially for larger models (T5-XXL).}
    \label{fig:mauve-hparams}
\end{figure*}

In \figureref{fig:mauve-hparams} we study the MAUVE performance tradeoffs for different hyperparameter configurations for the GPT2-medium model evaluated on Wikipedia data. Overall, we observe ---

\begin{itemize}
    \item Across all hyperparameter configurations, \model~significantly improves MAUVE score over a no re-ranking baseline.
    \item MAUVE scores improve for shorter rerank lengths, justifying the benefit of beam search over re-ranking of complete generations.
    \item For cases of full re-ranking (re-rank length = max length), increasing number of samples improves the MAUVE score (since \model~has more generations to choose from), but improvements saturates after 60 samples (for both model sizes), with the largest gain from 1 to 10 samples.
    \item We find that rerank length = 20 with 20 samples (beam size 2, samples per beam 10) performs best across all configurations. 
\end{itemize}

\subsubsection{Speed tradeoffs}
\label{appendix:speed-grid-search}

\begin{figure*}[t!]
    \centering
    \includegraphics[width=0.49\textwidth]{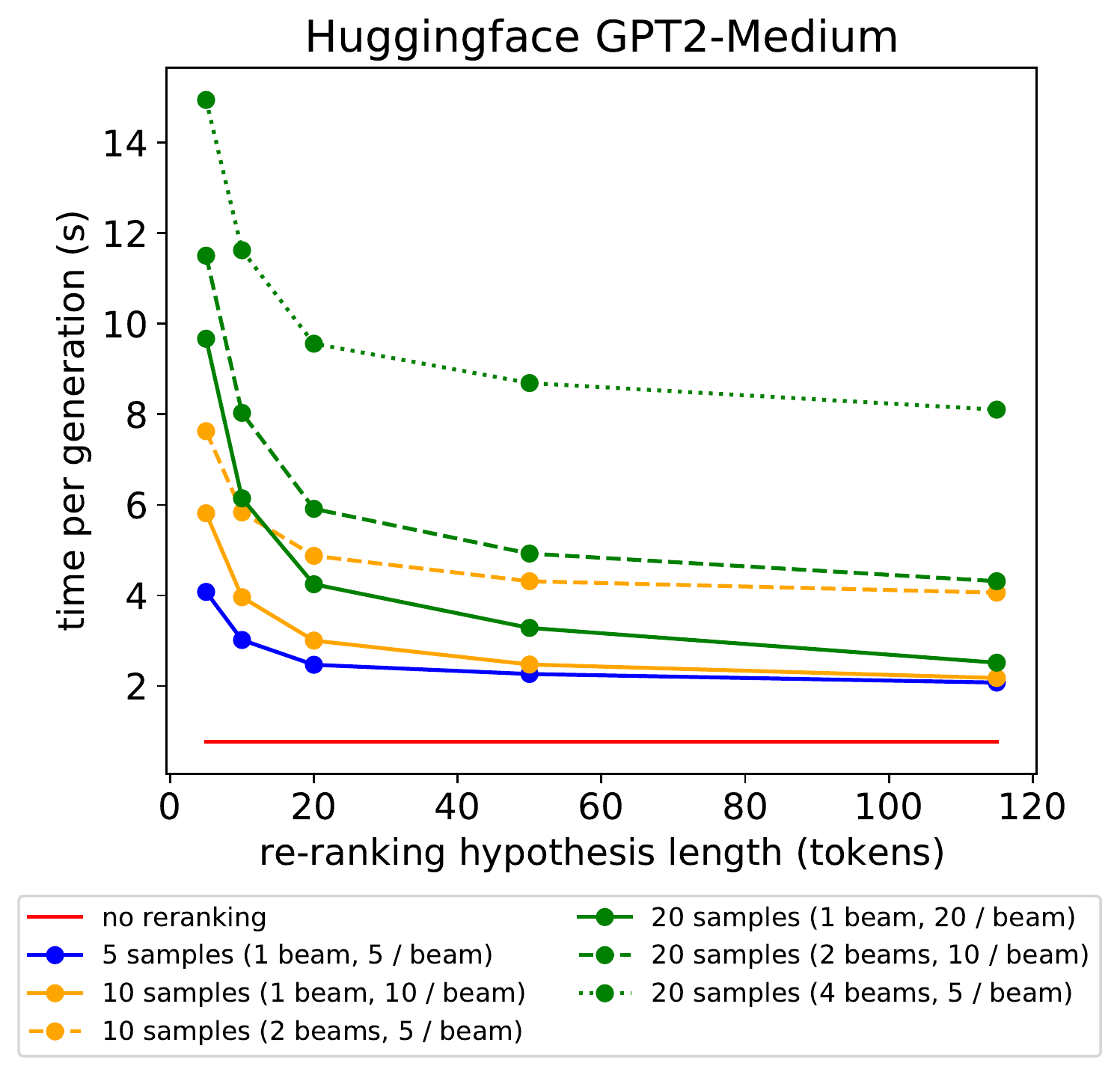}
    \includegraphics[width=0.49\textwidth]{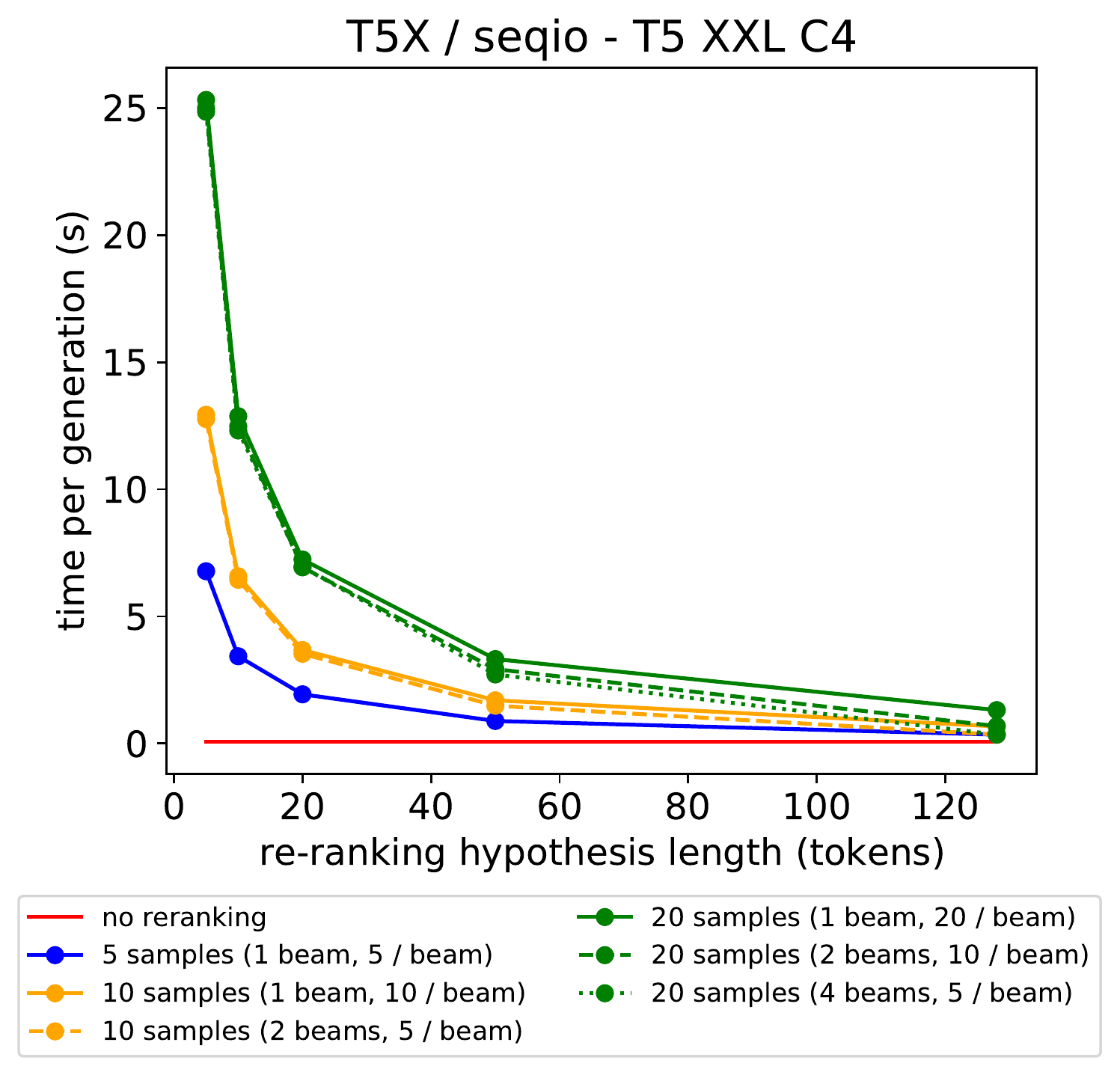}
    \vspace{-0.1in}
    \caption{Time taken (in seconds) for a single generation across different hyperparameter settings in both our implementations (HuggingFace / T5X). We see roughly linear increase in decoding time with number of samples, and linear increase with number of re-ranking steps (1 / rerank\_length).}
    \label{fig:timing-hparams}
\end{figure*}

In \figureref{fig:timing-hparams} we study the average time taken (in seconds) for a single generation on Wikipedia. Overall, in both our implementations we observe that ---

\begin{itemize}
    \item Decoding a single sample is an order of magnitude faster than decoding multiple samples (``over-generation''), which is needed before any re-ranking with \model~is possible.
    \item Reducing the rerank length increases decoding time, since more generate / re-rank cycles are needed. These cycles cannot be parallelized since the generate and re-rank steps are dependent on each other.
    \item Overall, we see observe that decoding time is roughly $\mathcal{O}(BN/L)$, where $B$ is beam size, $N$ is the number of samples per beam and $L$ is rerank length. This is especially true for the T5X implementation.
\end{itemize}

We dig a little deeper into these numbers: is the extra compute time due to over-generation (generation of 10 or 20 samples instead of one) or \model\ re-ranking? In \tableref{tab:timing-analysis-individual}, we measure the time taken to generate and score an individual instance. We see that re-ranking with \model~takes only a fraction of the time (1-10\%) compared to generation, which means that over-generation is the bottleneck. Also see \sectionref{sec:timing-analysis-main} in the main body of the paper for a performance / time tradeoff scatter plot.

\begin{table}[h]
\small
\begin{center}
\begin{tabular}{ lrrrr } 
 \toprule
  & \multicolumn{2}{c}{HuggingFace (GPT2)} & \multicolumn{2}{c}{T5X / seqio (T5)} \\
   & medium & XL & base & XXL\\
 \midrule
 secs / gen & 7.7e-1 & 2.9e0 & 8.1e-3 & 7.4e-2\\
 \midrule
\multicolumn{5}{l}{\textbf{\model~calls in same time as one generation}}\vspace{0.15cm} \\ 
base & 108.5 & 408.5 & 8.4 & 77.0 \\
large & 42.8 & 161.1 & 4.3 & 38.9 \\
XL & 16.4 & 61.7 & 1.7 & 15.7 \\
 \bottomrule
\end{tabular}
\end{center}
\vspace{-0.1in}
\caption{Number of \model~calls in the same time as one LM generation. Across libraries and LM sizes, \model~needs only a fraction of time vs generation.}
\label{tab:timing-analysis-individual}
\end{table}

\section{Human Evaluation Details}
\label{appendix:human-eval-details}

We hired freelancers from Upwork\footnote{\url{https://www.upwork.com}} as well as two volunteers to perform our human evaluation. In total, our human evaluation had eight annotators. Following recent recommendations from~\citet{karpinska-etal-2021-perils}, we ensured that each annotator (except one) was either an English teacher or an English writer. To avoid bias, we ensured that none of the annotators were computer science researchers, making them unaware of text generation research / \model.\\

\begin{figure*}[t!]
    \centering
    \includegraphics[width=0.99\textwidth]{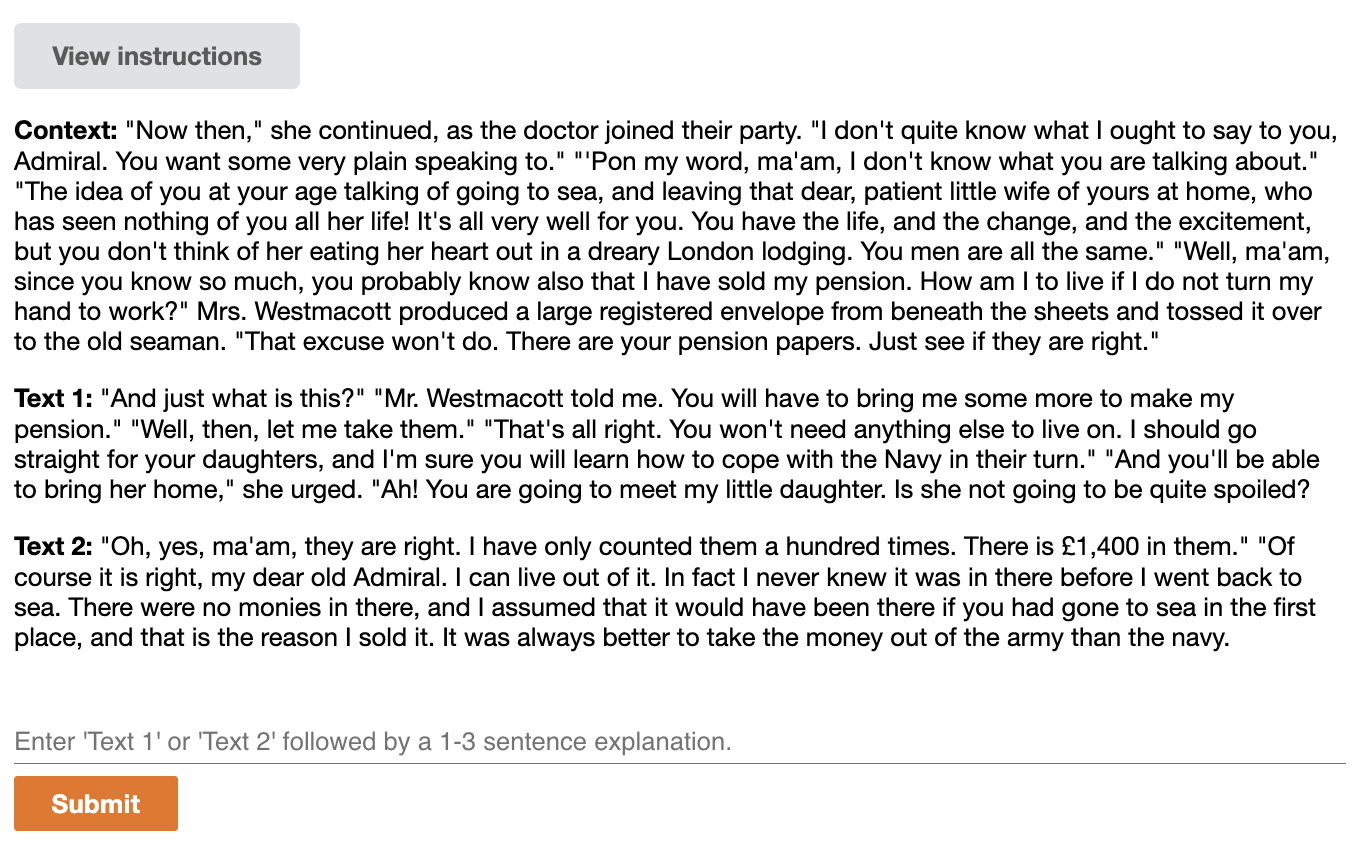}
    \caption{The interface shown to Upwork freelancers for human evaluation. We used Amazon Mechanical Turk Sandbox to collect our annotations (note that we use the MTurk Sandbox interface only; we do not hire any workers from MTurk due to poor annotation quality~\citep{karpinska-etal-2021-perils}.}
    \label{fig:mturk-interface}
\end{figure*}

\noindent \textbf{Setup}: Annotators were shown a 200-250 word prefix, and were asked to choose one of two 80-100 word continuations. Annotators were not told which model generated each continuation, and we shuffled the continuations in a random order to avoid position biases (``\emph{blind A/B testing}''). The job posting and instructions shown to the annotators are provided in \tableref{tab:upwork-instructions}. We used Amazon Mechanical Turk Sanbox\footnote{\url{https://requestersandbox.mturk.com/}} to collect our annotations, using the interface shown in \figureref{fig:mturk-interface}. Note that we used the MTurk Sandbox \emph{interface} only --- no MTurk \emph{workers} are recruited in our human study due to poor annotation quality for open-ended text generation~\citep{karpinska-etal-2021-perils,clark-etal-2021-thats}.\\

\noindent \textbf{Screening}: To ensure high annotation quality, we first asked annotators to complete a small screening test of 20 pairs with \inbookobjective~distractors, keeping 80\% accuracy as our passing criteria (estimated human performance on this set is 90-95\%). We paid annotators 10\$ for the screening test. Around half the interviewed Upworkers passed the test.\\

\noindent \textbf{Main Task (comparing generations)}: In our main task comparing generations from \model~with nucleus sampling, we asked annotators to choose the better continuation as well as provide a 1-3 sentence free-form explanation for their choice. We paid annotators 1\$ for each pair, and provided a 10\$ bonus at the end of a 100 pairs. Each annotator was provided with 100 instances (50 each from Wikipedia and PG19) either generated by the T5-XXL-C4 model~\citep{lester-etal-2021-power} or GPT2-medium~\citep{radford2019language}, with beam search outputs from \model-XL-all. Three annotators rate each model, giving us a total of 600 human annotations with explanations. \\

\noindent \textbf{Main Task (\inbookobjective~human estimate)}: Our second main task involved choosing the gold human-written continuation vs random \inbookobjective~negatives. We paid annotators 0.5\$ for this task, and did not ask them to explain their choices. This main task was similar in nature to our screening task.

\section{Suffix Identification}

\subsection{Gold vs \inbookobjective~- Hard examples}
\label{sec:gold-beats-neg-hard}

In \sectionref{sec:suffix-identification-ppl} and \appendixref{appendix:gold-beats-neg-extra} we make use of ``hard negatives''. To select these harder negative from the document, we use a trained \model\ model (XL sized, trained on all four domains). Specifically, we use \model~to score the compatibility of every 128-word token sequence in the document to the prefix, and take the highest scoring 10 sequences that are not the gold continuation (``Hard'' negative). All negatives sequences start and end at sentence boundaries so that LMs cannot rely on local syntactic patterns. For our two-way classification experiments in \sectionref{sec:suffix-identification-ppl}, we consider a random sequence among these 10 hard negatives. Since \model-all-XL-both was used to find these hard negatives, results on this \model\ variant are not very meaningful (since they are adversarial to this variant by construction).

\subsection{Gold vs \inbookobjective~- more negatives}
\label{appendix:gold-beats-neg-extra}

In \sectionref{sec:suffix-identification-ppl}, we used a single \inbookobjective~to test models. How do models fare when they need to choose the gold continuation over multiple \inbookobjective~negatives? In \tableref{tab:gold-beats-neg-main-10-way} we perform experiments on a 11-way classification task (10 \inbookobjective~negatives). Overall, we find that most LMs do barely above chance, whereas \model~significantly outperforms large LMs (even GPT3).

\begin{table}[h]
\small
\begin{center}
\begin{tabular}{ lrrrr } 
 \toprule
 \inbookobjective~neg type $\rightarrow$  & \multicolumn{2}{c}{Random} & \multicolumn{2}{c}{Hard} \\
  \cmidrule(lr){2-3} \cmidrule(lr){4-5}\vspace{-0.3cm}\\
  & PG & Wiki & PG & Wiki\\
 \midrule
 Random & 9.1 & 9.1 & 9.1 & 9.1\\
 Unigram Overlap & 42.3 & 18.5 & 8.6 & 5.0 \\
 GPT2-medium & 25.5 & 12.0 & 7.8 & 4.8 \\
 GPT2-XL~\shortcite{radford2019language} & 29.1 & 12.6 & 8.3 & 5.0 \\
 T5-base (f.t. PG19)  & 28.8 & 14.3 & 7.8 & 5.1  \\
 T5-XXL (f.t. PG19) & 38.8 & 17.5 & 9.8  & 6.0 \\
 T5-XXL-C4~\shortcite{lester-etal-2021-power} & 34.3 & 14.6 & 9.2 & 5.5 \\
 GPT3 170B*~\shortcite{brown2020language} & 32.0 & 14.0 & 14.0  & 8.0 \\
 \midrule
 \model~(ours) \\
 PG19-XL-\inbookobjective & \textbf{94.4} & 69.8 & 49.1 & 36.5 \\
 PG19-XL-\textsc{Generate} & 45.0 & 28.5 & 11.7 & 11.8 \\
 PG19-XL-both & \textbf{94.4} & 69.0 & \textbf{49.5} & 35.7\\
 \modelallboth & 92.6 & \textbf{84.6} & 39.5$^\dagger$  & \textbf{52.1}$^\dagger$ \\
\bottomrule
\end{tabular}
\end{center}
\vspace{-0.1in}
\caption{A version of \tableref{tab:gold-beats-neg-main} with 10 distractors (11-way classification). Like \tableref{tab:gold-beats-neg-main}, large LMs perform poorly and close to chance on hard sets. *GPT3 scores computed using 100 datapoints. $^\dagger$The hard sets were adversarially constructed using this \model~variant.}
\vspace{-0.1in}
\label{tab:gold-beats-neg-main-10-way}
\end{table}

\noindent \textbf{Gold vs all \inbookobjective~negatives (``retrieval'')}: What if instead of 10 negatives, we used all possible \inbookobjective~negatives in the book? This task could be framed as a \emph{retrieval} problem akin to RELiC (\sectionref{sec:retrieval-results}): given a prefix, find the correct continuation from all possible continuations in the same book. Since PG19 books can be quite long, retrievers needs to search among 2538 candidates on average in the PG19 validation set. We present results on this retrieval task in \tableref{tab:pg19-retrieval}. Overall, we find that \model~is quite successful at this task, getting a recall@1 of 48.2\% with a model trained on just PG19 data and \inbookobjective~negatives. Training on just PG19, increase model size, increasing minibatch size and using just \inbookobjective~negatives helps improve retrieval performance. In initial experiments, we extensively used performance on this task to hill-climb and justify our design choices. Note that we do not test LMs on this retrieval task, since it is computationally expensive to do a forward pass for each of the 2538 candidates for each of the 100K datapoints.

\begin{table}[h!]
\small
\begin{center}
\begin{tabular}{ lrrrrr } 
 \toprule
  &  & \multicolumn{4}{c}{Retrieval over PG19 books}  \\
 \cmidrule{3-6}
 Model & Batch & \\
 Size & Size & R@1 & R@3 & R@5 & R@10 \\
 \midrule 
 \multicolumn{6}{l}{\textit{(\model~models trained on PG19)}} \vspace{-0.1in} \\\\ 
 base & 4096 & 34.9 & 52.6 & 60.6 & 70.5 \\
 large & 4096 & 45.2 & 62.8 & 69.9 & 78.1 \\
  XL & 1536 & 48.1 & 65.4 & 72.1 & 79.7  \\
   XL-inbook & 1536 &  48.2 & 65.5 & 72.1 & 79.7 \\
   XL-gen & 1536 &  4.4 & 10.4 & 14.4 & 20.5 \\
  \midrule
  \multicolumn{6}{l}{\textit{(\model~models trained on all 4 domains)}} \vspace{-0.1in} \\\\ 
 base & 4096 & 28.4 & 44.4 & 52.1 & 62.4 \\
 large & 4096 & 39.6 & 56.8 & 64.0 & 72.9 \\
  XL & 256 & 24.3 & 38.7 & 45.7 & 55.4 \\
 XL & 512 & 31.7 & 47.5 & 54.6 & 64.1 \\
 XL & 768 & 34.6 & 51.0 & 58.5 & 67.5 \\
 XL & 1536 & 41.5 & 58.8 & 65.7 & 74.3  \\
 \bottomrule
\end{tabular}
\end{center}
\caption{\model~retrieval performance on PG19 validation books. On average, retrieval takes place over 2538 candidates. \model~gets high performance on this task, and scaling model size, scaling minibatch size, training on just PG19 and using just \inbookobjective~negatives improves recall@1 (R@1).}
\label{tab:pg19-retrieval}
\end{table}

\subsection{Gold vs \genobjective~- breakdown by generative model}
\label{appendix:gold-beats-gen-extra}

See \tableref{tab:gold-beats-generation-extra} for a breakdown by the model used to create the \genobjective~negatives.

\begin{table*}[t!]
\small
\begin{center}
\begin{tabular}{ lrr|rr|rr|rr|r } 
 \toprule
 Discriminator & \multicolumn{2}{c|}{GPT2-md} & \multicolumn{2}{c|}{GPT2-XL} & \multicolumn{2}{c|}{T5-XXL-PG19} & \multicolumn{2}{c|}{T5-XXL-C4} & Average \\
 & PG19 & wiki & PG19 & wiki  & PG19 & wiki & PG19 & wiki \\
 \midrule
 Random & 50.0 & 50.0 & 50.0 & 50.0 & 50.0 & 50.0 & 50.0 & 50.0 & 50.0 \\
 Unigram Overlap & 38.4 & 43.6 & 36.7 & 39.8 & 48.5 & 56.8 & 37.2 & 37.4 & 42.3\\
 \midrule
 GPT2-medium~\shortcite{radford2019language} & 2.1 & 4.9 & 3.0 & 6.6 & 36.1 & 59.1 & 17.2 & 22.7 & 19.0 \\
 GPT2-XL~\shortcite{radford2019language} & 12.7 & 23.3 & 1.7 & 4.6 & 45.1 & 68.7 & 26.5 & 29.3 & 26.5\\ 
 T5-XXL (f.t. PG19) & 46.2 & 54.6 & 23.5 & 29.7 & 28.5 & 26.3 & 31.5 & 24.1 & 33.1\\
 T5-XXL-C4~\shortcite{lester-etal-2021-power} & 24.7 & 52.2 & 10.9 & 26.1 & 31.9 & 65.2 & 8.5 & 13.0 & 29.1 \\
 \midrule
 \model~(ours)   \\
 PG-XL-\genobjective &  \textbf{96.9} & \textbf{91.4} & \textbf{95.7} & \textbf{88.8} & \textbf{91.8} & 92.3 & \textbf{94.3} & \textbf{84.4} & \textbf{91.9} \\
 PG-XL-\inbookobjective & 78.4 & 66.3 & 69.7 & 60.3 & 65.9 & 60.1 & 65.2 & 52.2 & 64.8 \\
 PG-XL-both & 97.4 & 81.3 & 93.7 & 74.0 & 87.4 & 79.4 & 89.7 & 65.0 & 83.5 \\
 \modelallboth & 94.3 & 84.5 & 88.8 & 78.0 & 80.3 & \textbf{95.3} & 81.3 & 67.3 & 83.7 \\
\bottomrule
\end{tabular}
\end{center}
\vspace{-0.1in}
\caption{A version of \tableref{tab:gold-beats-generation-main} breaking down performance by domain (Project Gutenberg PG19, Wikipedia) and model used to generate \genobjective~negatives using nucleus sampling~\citep{holtzman2020curious} with $p=0.9$. Language model perplexity prefers \genobjective~sequences over human text (as previously noted by~\citealp{gehrmann-etal-2019-gltr}), especially when the \genobjective~negative is generated by the same language model.}
\vspace{-0.1in}
\label{tab:gold-beats-generation-extra}
\end{table*}

\subsection{Details of Suffix Identification Datasets}
\label{appendix:details-suffix-id-datasets}

\noindent \textbf{ChapterBreak}~\citep{sun2022chapter} is a 6-way classification task in which models are provided as input a long segment from a narrative that ends in a chapter boundary. Models must then identify the correct ground-truth chapter beginning from a set of negatives sampled from the same narrative --- a task requiring global narrative understanding. ChapterBreak has two settings: (1) PG19 --- the validation set of the Project Gutenberg language modeling benchmark~\cite{rae2019compressive}; (2) AO3 --- a ChapterBreak split adapted from fan-fiction posted to Archive of Our Own (AO3).\footnote{\url{https://archive.org/download/AO3_story_dump_continuing}} Although~\citet{sun2022chapter} provide prefixes up to 8192 tokens, we study ChapterBreak in the setting using just 256 tokens of prefix to ensure compatibility with the input lengths of \model. The ChapterBreak dataset is not divided into validation / test splits, so we simply use the single available split.\\

\noindent \textbf{HellaSwag}~\citep{zellers-etal-2019-hellaswag} is a 4-way classification task focusing on commonsense natural language inference. For
each question, a prefix from a video caption is provided as input and a model must choose the correct continuation for this prefix. Only one out of the four choices is correct – the actual next caption of the video. HellaSwag is scraped from the video captions in ActivityNet~\citep{krishna2017dense} and how-to paragraph instructions on WikiHow. We study the setting where each of the 4 endings are complete sentences, which is constructed by prepending \texttt{ctx\_b} to the given endings). We use the validation set of the HellaSwag corpus since the test set answers are hidden.\\

\noindent \textbf{StoryCloze}~\citep{mostafazadeh-etal-2016-corpus, sharma-etal-2018-tackling} is a 2-way classification task designed to test commonsense reasoning. Systems are provided with the first four sentences of a five-sentence commonsense story, and must choose the correct ending to the story. We used the test set for the Spring 2016 split and the validation set for the Winter 2018 split (due to the hidden test set).

\subsection{\model~for suffix identification}
\label{sec:suffix-id-results}

\model~is trained on a \emph{suffix identification} objective: given a prefix, choose the gold continuation over \inbookobjective~and \genobjective~negatives. How well does \model~learn this task? How does \model~fare on existing suffix identification benchmarks?

\begin{table}[t!]
\small
\begin{center}
\begin{tabular}{ lrrrrr } 
 \toprule
 & \multicolumn{2}{c}{ChapterBreak} & \multicolumn{2}{c}{StoryCloze} & HSw  \\
  & PG19 & AO3 & 2016 & 2018 &  \\
 \midrule
 prefix length & 240.3 & 241.6 & 35.4 & 35.3 & 39.5 \\
 suffix length & 152.9 & 156.1 & 7.4 & 7.4 & 26.0 \\
 \midrule
 \midrule
 Random & 16.7 & 16.7 & 50.0 & 50.0 & 25.0 \\
 Token overlap & 37.3 & 28.7 & 39.9 & 40.9 & 27.4 \\
 GPT2-md & 20.3 & 21.5 & 66.7 & 66.9 & 36.8 \\
GPT2-XL & 21.6 & 23.2 & 71.5 & 72.6 & 48.2 \\
 T5-base-PG & 23.2 & 23.4 & 59.0  & 61.9 &  33.1 \\
 T5-XXL-PG & 28.6 & 25.3 & 69.3 & 73.5 &  62.3 \\
 T5-XXL-C4 & 24.1 & 24.3 & 76.0 & \textbf{77.8} & 63.6 \\
 GPT3 (170B) & 26.0 & 23.8 & 83.2 & - & 78.9 \\
 PaLM (540B) & - & - & \textbf{84.6} & - & \textbf{83.4} \\
 \midrule
 \multicolumn{4}{l}{\model~(1.2B, ours)} \\
 PG-XL-\textsc{gen} & 33.6 & 21.8 & 57.9 & 57.9 & 35.0 \\
 PG-XL-\textsc{inbk} & \textbf{64.3} & \textbf{39.5} & 73.4  & 72.6 & 39.3 \\
 PG-XL-both & 63.5 & 36.9 & 71.1 & 72.6 & 40.7 \\
 \modelallboth & 59.3 & 32.8 & 75.4 & 75.8 & 46.3 \\
 \bottomrule
\end{tabular}
\end{center}
\vspace{-0.1in}
\caption{Zero-shot suffix identification results on existing datasets. \model~significantly outperforms all LMs on ChapterBreak which has long prefix/suffix lengths. \model~performs similar to similar-sized GPT2-XL on StoryCloze and HellaSwag, with shorter inputs and more local dependencies.}
\vspace{-0.1in}
\label{tab:suffix-id-results}
\end{table}

\paragraph{Performance on \inbookobjective~/ \genobjective:} In \sectionref{sec:suffix-identification-ppl} we motivated the \model~design by showing the inability of LM perplexity to prefer the gold continuations over negatives. How does \model~fare on these negatives? In \tableref{tab:gold-beats-neg-main} and \tableref{tab:gold-beats-generation-main} we evaluate the performance at distinguishing gold continuations from negatives, and compare \model~to large LMs. Since \model~is directly optimized on this objective, it significantly outperforms large LMs (99.1\% vs 78.2\% with GPT-3 for \inbookobjective). \model~variants trained on just \inbookobjective~or just \genobjective~perform best at their respective tasks, but we observe some generalization (\inbookobjective~model gets 69.8\% on \genobjective~PG19 negatives, \genobjective~model gets 80.2\% on \inbookobjective~negatives, both higher than all LMs). Strong performance on \genobjective~could have several applications like fake news detection~\citep{zellers2019defending,gehrmann-etal-2019-gltr}, and is an interesting future work direction.

\paragraph{Performance on existing suffix identification benchmarks:} We test \model~on three existing suffix identification datasets --- ChapterBreak~\citep{sun2022chapter}, ROCStories cloze test~\citep{mostafazadeh-etal-2016-corpus} and HellaSwag~\citep{zellers-etal-2019-hellaswag}; dataset details are provided in \appendixref{appendix:details-suffix-id-datasets}. To measure their intrinsic capability, models are evaluated \textbf{zero-shot}, without finetuning on training sets.\footnote{\citet{zellers-etal-2019-hellaswag} also describe \textit{zero-shot} HellaSwag experiments, testing models on unseen WikiHow / ActivityNet categories; however they still finetune models on HellaSwag data for seen categories, while we do no such finetuning.}

 In \tableref{tab:suffix-id-results} we find that \model~significantly outperforms all LMs on ChapterBreak (64.3 vs 28.6). \model~performs comparably to similar-sized GPT2-XL (1.5B parameters) on other tasks, beating it on StoryCloze (75.8 vs 72.6), but slightly worse on HellaSwag (46.3 vs 48.2). Much larger LMs like GPT3 170B~\citep{brown2020language} and PaLM 540B~\citep{Chowdhery2022PaLMSL} perform best on StoryCloze and HellaSwag. Scaling also benefits \model~(30.4 vs 40.7 on HellaSwag for base vs XL), and we believe further scaling \model~is a promising direction for future work. We also find \inbookobjective~negatives are more beneficial than \genobjective~negatives~(64.3 vs 33.6 on ChapterBreak PG19). We hypothesize that the different trends on different datasets can be attributed to input length. As seen in \tableref{tab:suffix-id-results}, ChapterBreak has much longer inputs (240 prefix, 153 suffix tokens) than other datasets (35 prefix, 7 suffix tokens for ROCStories). The focus on local context in LMs~\citep{khandelwal-etal-2018-sharp,sharan2018prediction,sun-etal-2021-long} helps with short-range tasks but also likely contributes to their underperformance on complex long-range tasks like ChapterBreak.

\begin{table}[t]
\small
\begin{center}
\begin{tabular}{ lrrrrr } 
 \toprule
 Scorer & CB-PG & SC-2016 & HS & PG19 & Wiki \\
 \midrule
 Random & 16.7 & 50.0 & 25.0 & 9.1 & 9.1 \\
 CLL & 16.2 & 63.0 & 32.2 & 15.9 & 8.5 \\
 avg CLL & 20.3 & 66.7 & 36.8 & 25.5 & 12.0 \\
 avg ULL & 20.8 & 66.0 & \textbf{37.0} & 25.2 & 11.8\\
 PMI & \textbf{38.2} & \textbf{68.3} & 32.9 & \textbf{62.3} & \textbf{26.3} \\
 \midrule
 \multicolumn{6}{l}{\model~(1.2B, ours)} \\
 PG-\textsc{inbk} & \textbf{64.3} & 73.4 & 39.3 & \textbf{94.4} & 69.8 \\
 all-\textsc{both} & 59.3 & \textbf{75.4} & \textbf{46.3} & 92.6 & \textbf{84.6} \\
 \bottomrule
\end{tabular}
\end{center}
\caption{GPT2-medium suffix identification performance with different scoring functions (\sectionref{sec:choice-metric-gold-beats-neg}). Datasets used are ChapterBreak-PG19 (CB-PG), StoryCloze-2016 (SC-2016), HellaSwag (HS) and PG19 / Wikipedia \inbookobjective~negatives with 10 random distractors, as computed in \tableref{tab:gold-beats-neg-main-10-way}. }
\vspace{-0.1in}
\label{tab:suffix-id-metrics}
\end{table}

\subsection{Choice of Scoring Function}
\label{sec:choice-metric-gold-beats-neg}

It is argued in~\citet{holtzman-etal-2021-surface} that average log likelihood is a sub-optimal scoring function when LMs are used to score sequences. In this section, we compare several scoring functions on GPT2-medium. Let $p$ be a prefix and $c$ be a continuation. We consider: (1) conditional log likelihood (CLL), or $\log P(c|p)$; (2) average conditional log likelihood (avg CLL), or $\frac{1}{|c|} \log P(c|p)$; (3) average \emph{un}conditional log likelihood (avg ULL), or $\frac{1}{|c| + |p|} \log P(p \oplus c)$; and (4) pointwise mutual information (PMI), or $\log \frac{P(c|p)}{P(c)}$. We compare these scoring functions on several datasets in \tableref{tab:suffix-id-metrics}. Overall, we find that PMI is a strong scoring function, outperforming all other functions on four out of five datasets. Length normalized scoring functions (avg CLL/ULL) are better than CLL across all datasets, consistent with findings in prior work~\citep{wu2016google,koehn-knowles-2017-six,brown2020language}. All scoring functions lag behind \model\ in all five datasets.

Throughout this paper we use ``avg CLL'' to report suffix identification scores. Length normalized conditional log likelihood is the most closely aligned to how text is generated (sampling from the next-token distribution), and is the objective language models are directly optimized on. However, given the strong performance of PMI compared to ``avg CLL'' on four out of five datasets, an interesting future direction is studying the benefit of PMI or domain-conditioned PMI~\citep{holtzman-etal-2021-surface} in generating text.

\section{More Evaluation Details \& Results}
\label{appendix:evaluation}

\subsection{MAUVE setup}
\label{appendix:mauve-details}

We extensively use the MAUVE metric from~\citet{pillutla2021mauve} for automatic evaluation of our model. MAUVE is shown to have high correlation with human judgements of the quality of generated text. We closely follow the best practices listed in the official MAUVE repository,\footnote{\url{https://github.com/krishnap25/mauve\#best-practices-for-mauve}} which we found critical in preliminary experiments. Specifically,

\begin{enumerate}
    \item We ensure that each run has the exact same hyperparameters --- using the default hyperparameters in the official MAUVE library.
    \item We use 7713 generations per run, which is the size of our Wikipedia validation set. This follows the suggestion in the official codebase README of having at least 5000 generations for comparing models. While our PG19 validation set is much bigger, we truncate it to 7713 generations since MAUVE scores tend to reduce with more generations.
    \item Since MAUVE scores are higher for shorter generations, we ensure that all tested methods have roughly equal generation lengths, between 70-80 words / 120-130 tokens. We also truncate human text / generations to ensure that each instance ends at a sentence boundary. In initial experiments we observed that truncating consistently for human text and machine text leads to lower MAUVE variation. 
    \item Due to variation in MAUVE score from run to run, we average the MAUVE score for nucleus / top-k / typical sampling over five runs. For the T5-XXL-C4 model on Wikipedia with nucleus sampling, the MAUVE scores were [0.803, 0.778, 0.759, 0.785, 0.768], giving a standard deviation of 0.015.
\end{enumerate}

\subsection{MAUVE Divergence Curves}
\label{sec:mauve-divergence-curves}

The MAUVE metric is the area under a divergence curve, a curve which attempts to analyze the type of errors the model is making. Given $P$ is the distribution of human text and $Q$ is the distribution of machine-generated text, ~\citet{pillutla2021mauve} describe two types of errors made by models  ---

\vspace{0.1in}

\noindent \textbf{Type I}: $\text{KL}(Q|P)$ --- False positives, or cases where models generate text which is unlikely to be written by humans, like semantic repetitions common in neural text generators~\citep{holtzman2020curious,zhang-etal-2021-trading}.

\vspace{0.1in}

\noindent \textbf{Type II}: $\text{KL}(P|Q)$ --- False negatives, or cases where models cannot generate text which is likely to be written by humans, sometimes seen with truncation strategies~\citep{see-etal-2019-massively}.

\vspace{0.1in}

\begin{figure*}[t!]
    \centering
    \includegraphics[width=0.4\textwidth]{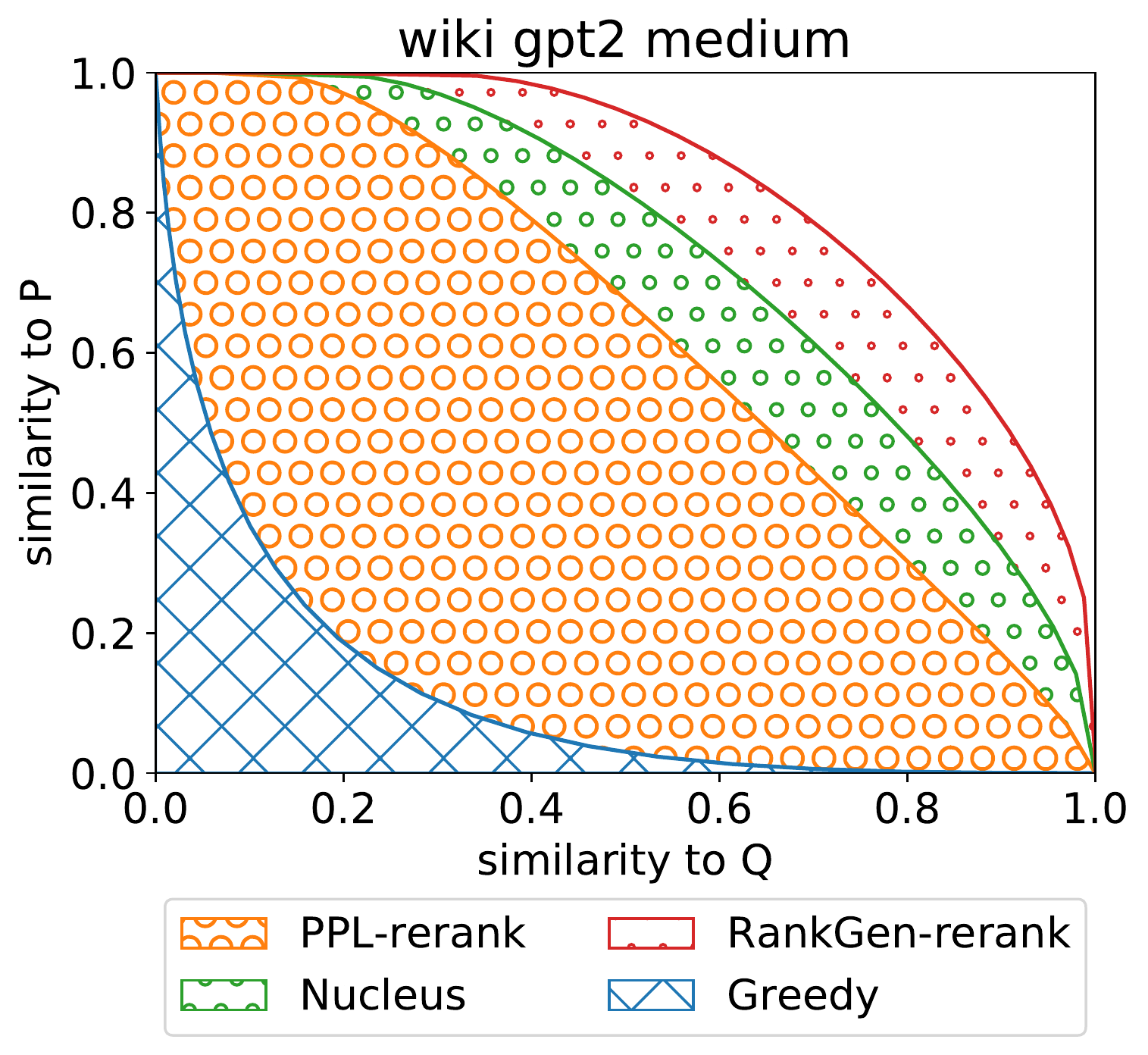}
    \includegraphics[width=0.4\textwidth]{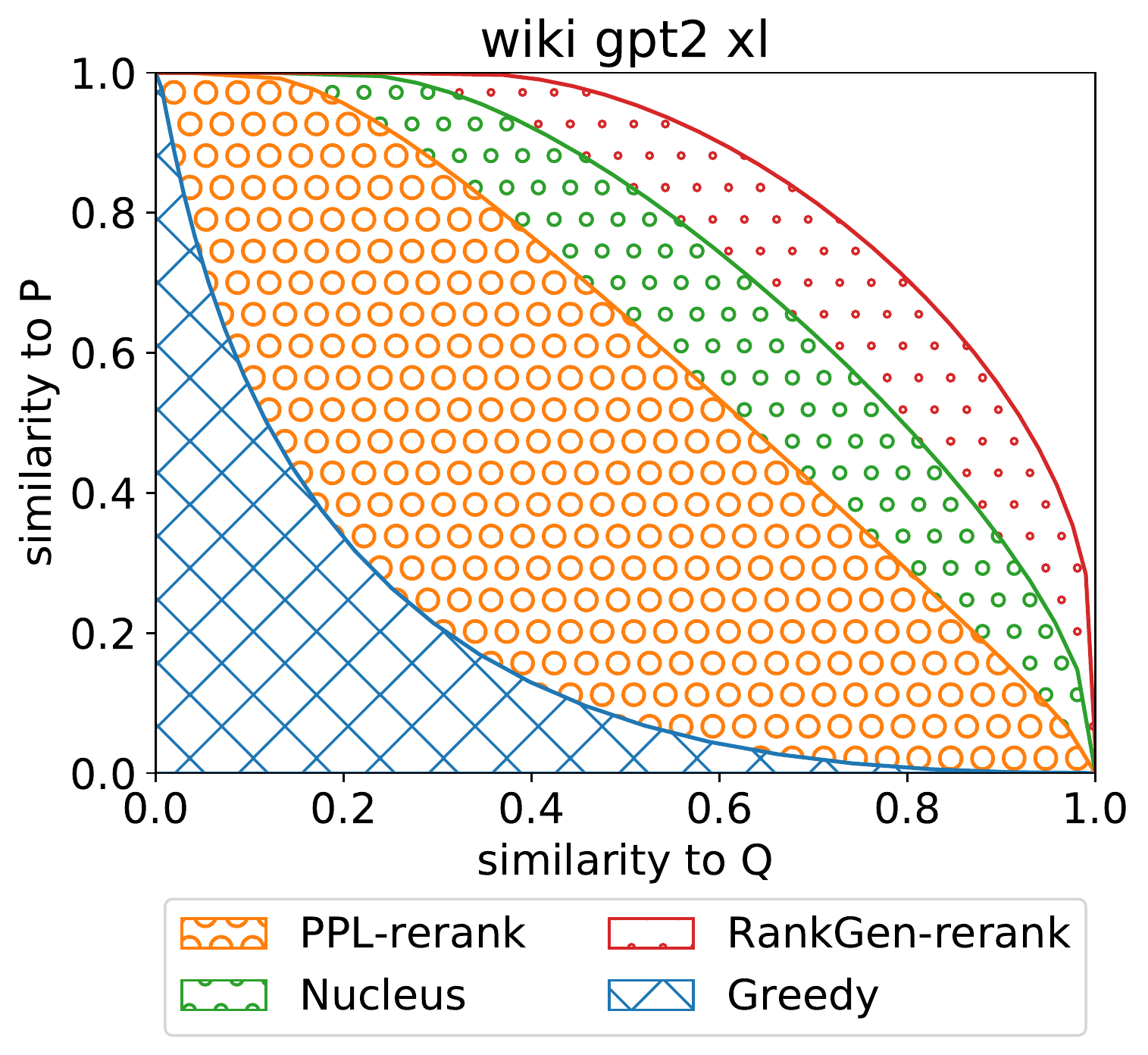}
    \includegraphics[width=0.4\textwidth]{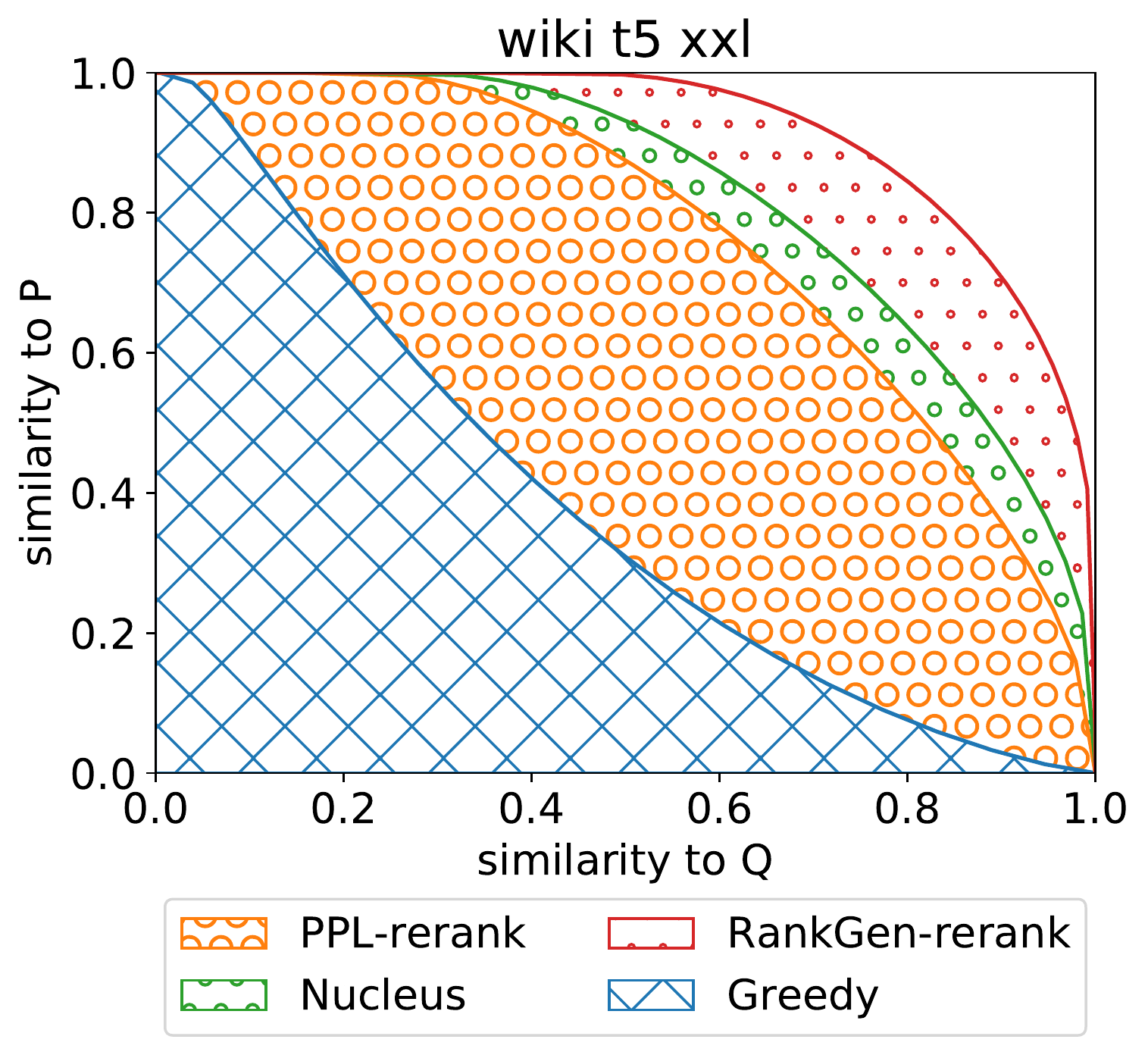}
    \includegraphics[width=0.4\textwidth]{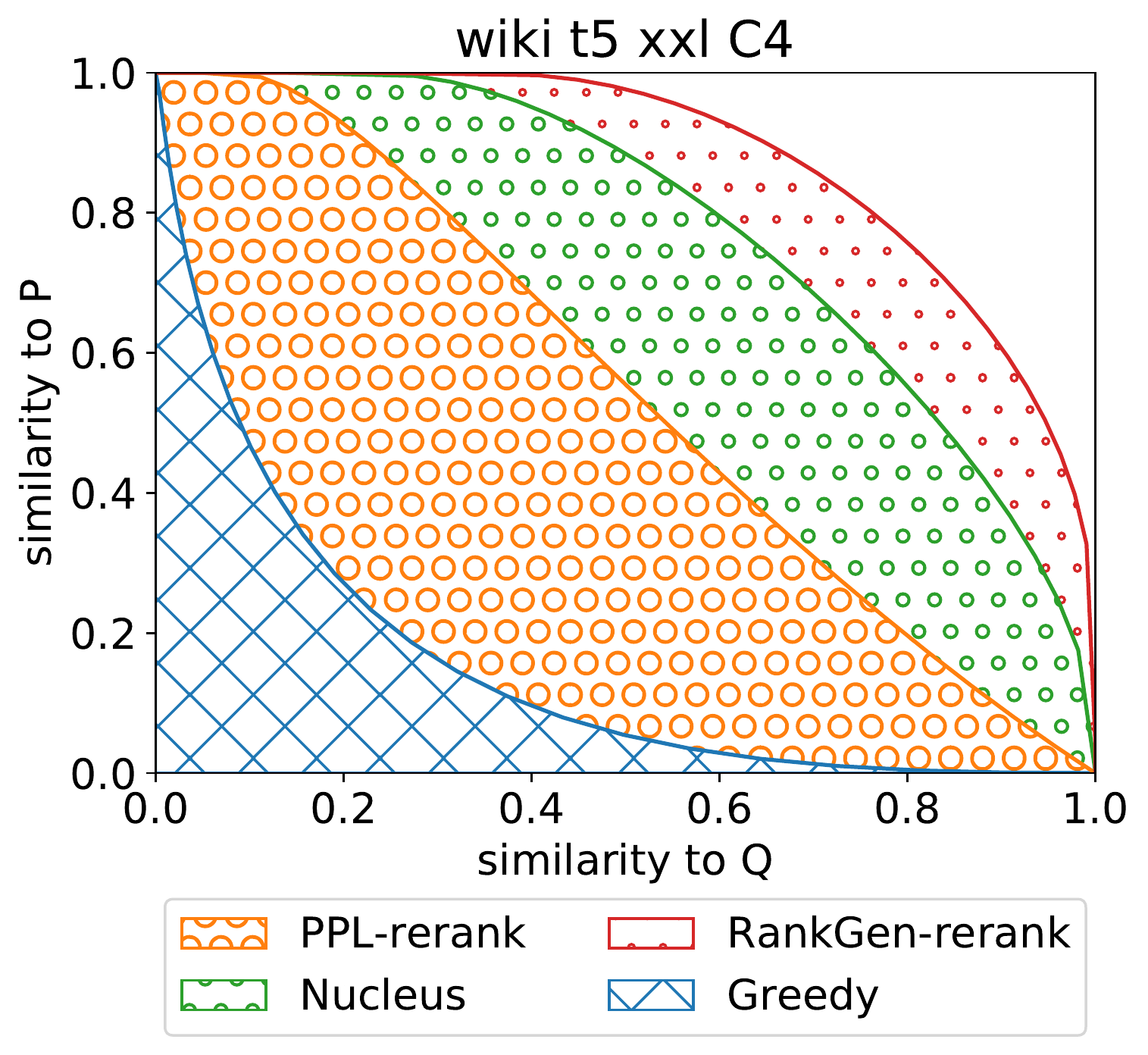}
    \caption{Divergence curves~\citep{pillutla2021mauve} after full sample re-ranking on Wikipedia inputs using \model-XL trained on all four domains. The area under this curve is the MAUVE score. Overall, we see that \model~ makes fewer Type I (bigger intercept with $y=1$ line) and Type II style errors (bigger intercept with $x=1$ line). PPL re-ranking increases the amount of repetition in generated text (\tableref{tab:rep-scores}), leading to more Type I errors (smaller intercept with $y=1$ line).}
    \label{fig:wiki-divergence-curves}
\end{figure*}

\begin{figure*}[t!]
    \centering
    \includegraphics[width=0.4\textwidth]{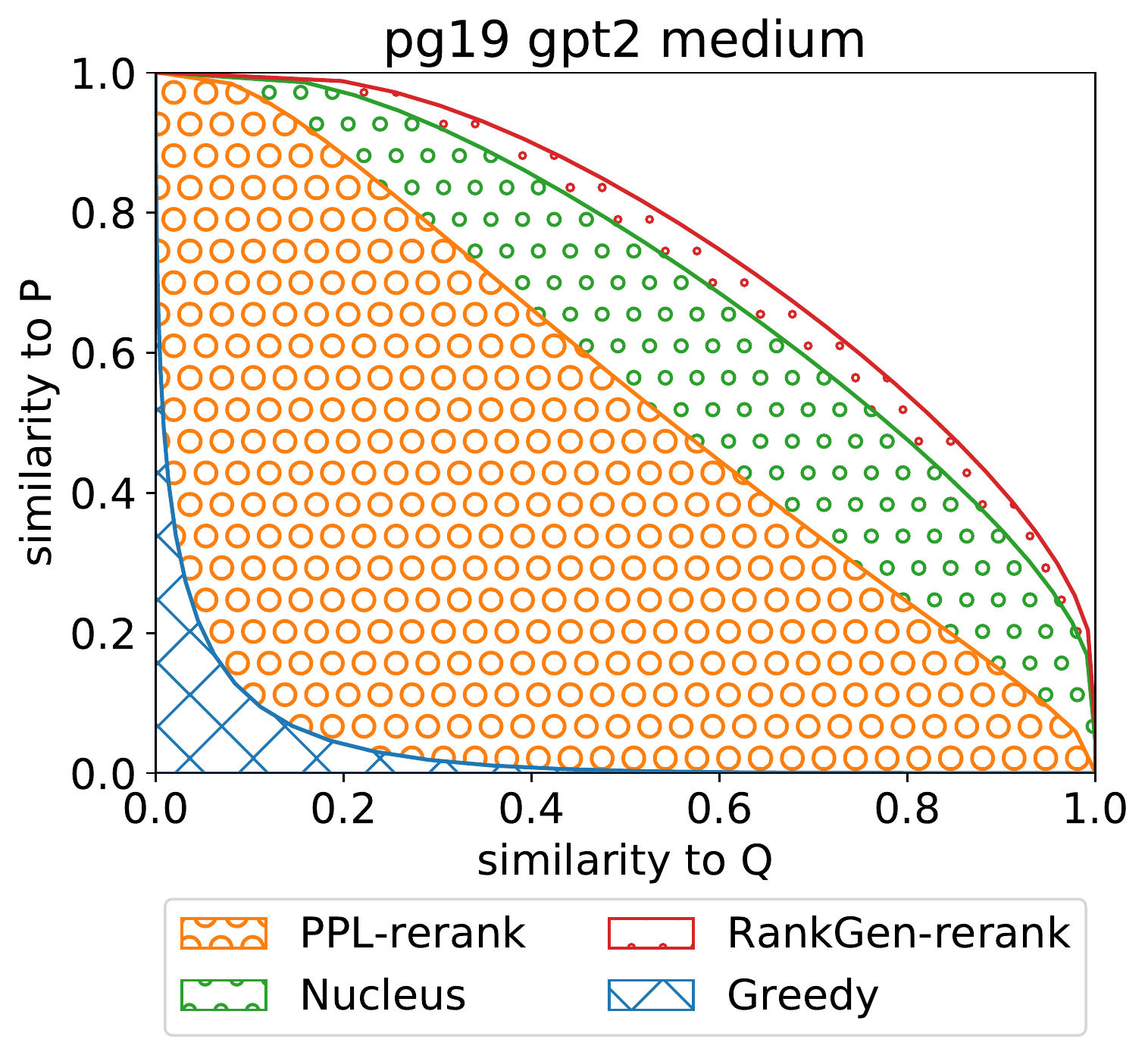}
    \includegraphics[width=0.4\textwidth]{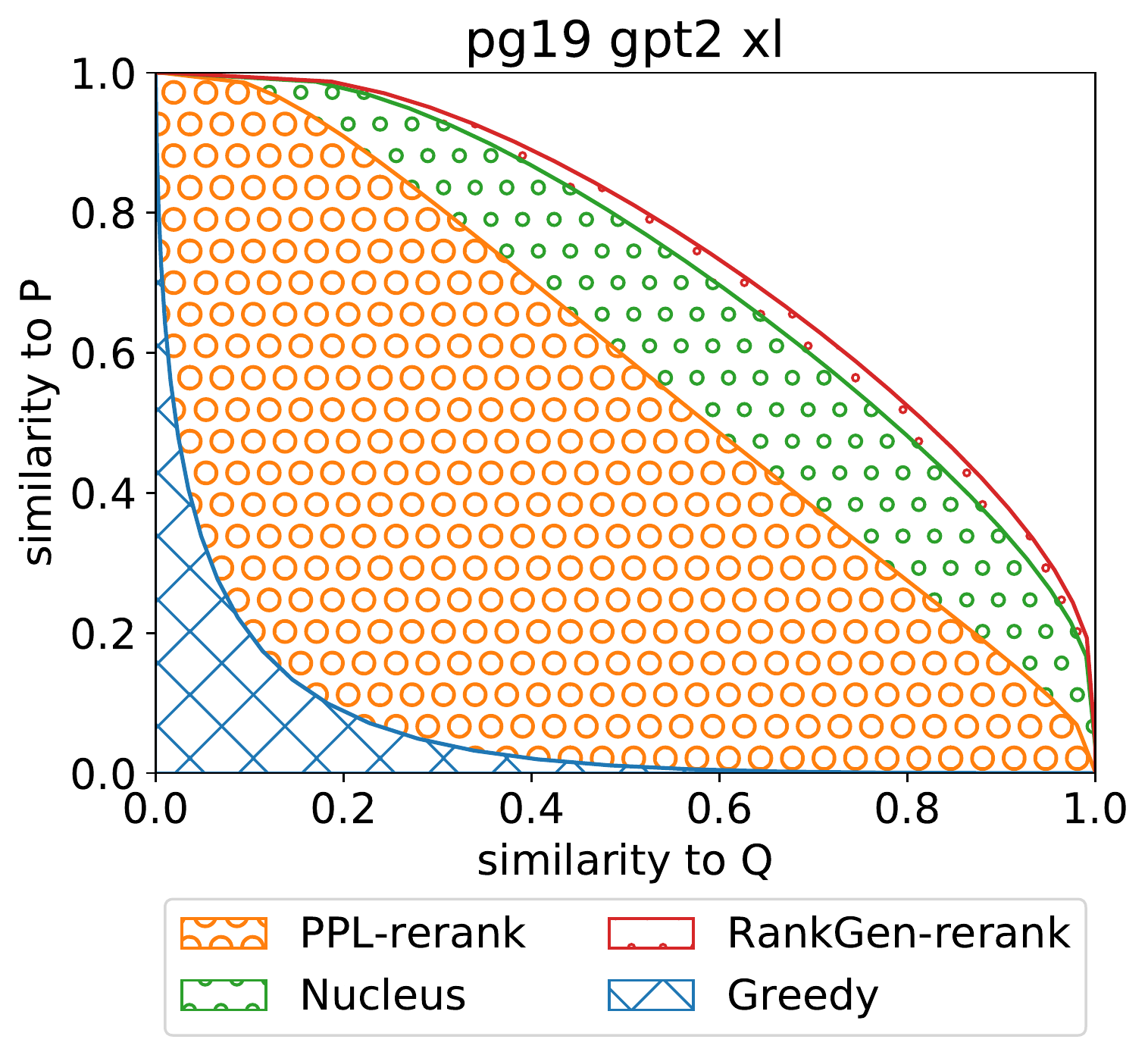}
    \includegraphics[width=0.4\textwidth]{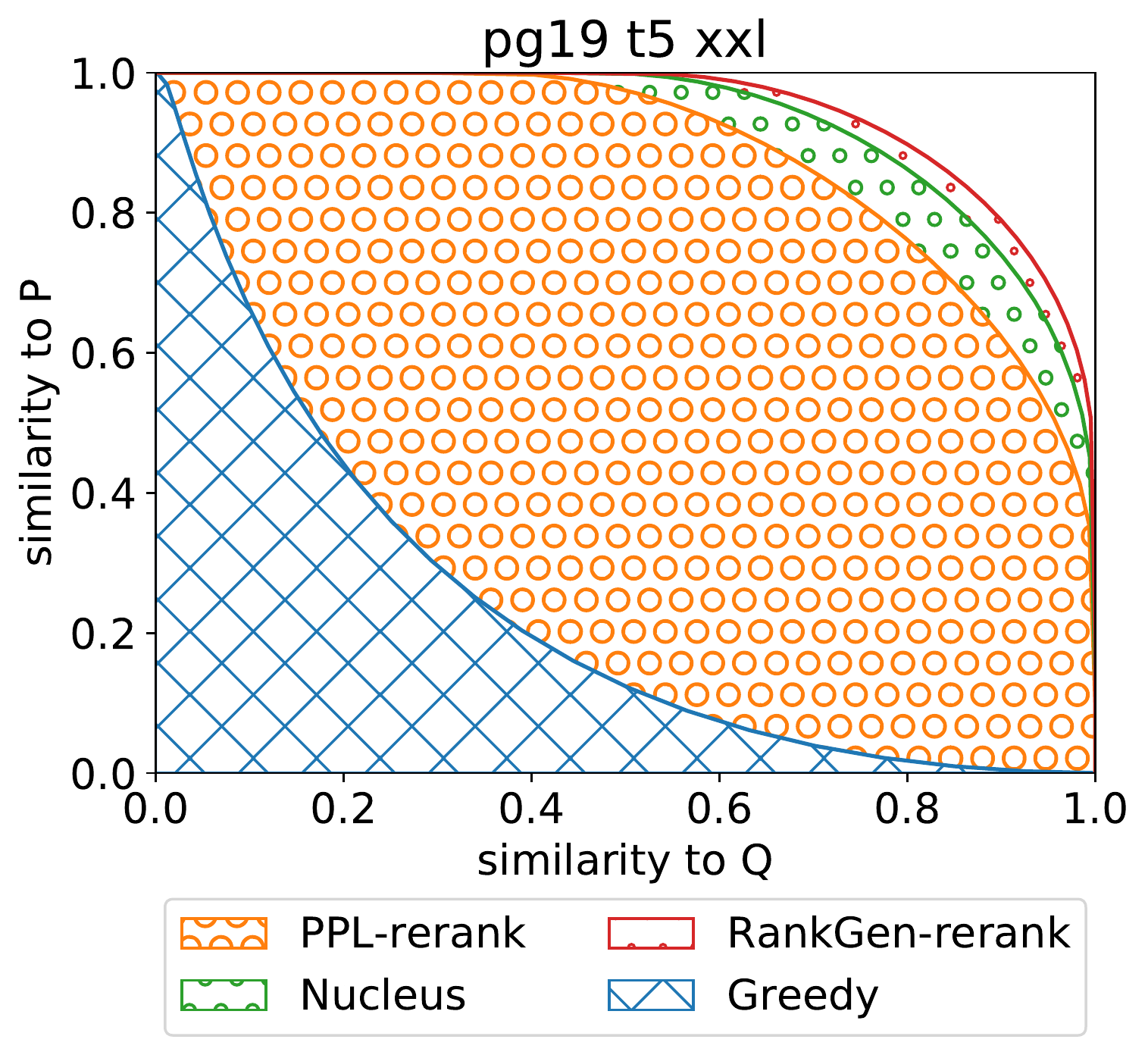}
    \includegraphics[width=0.4\textwidth]{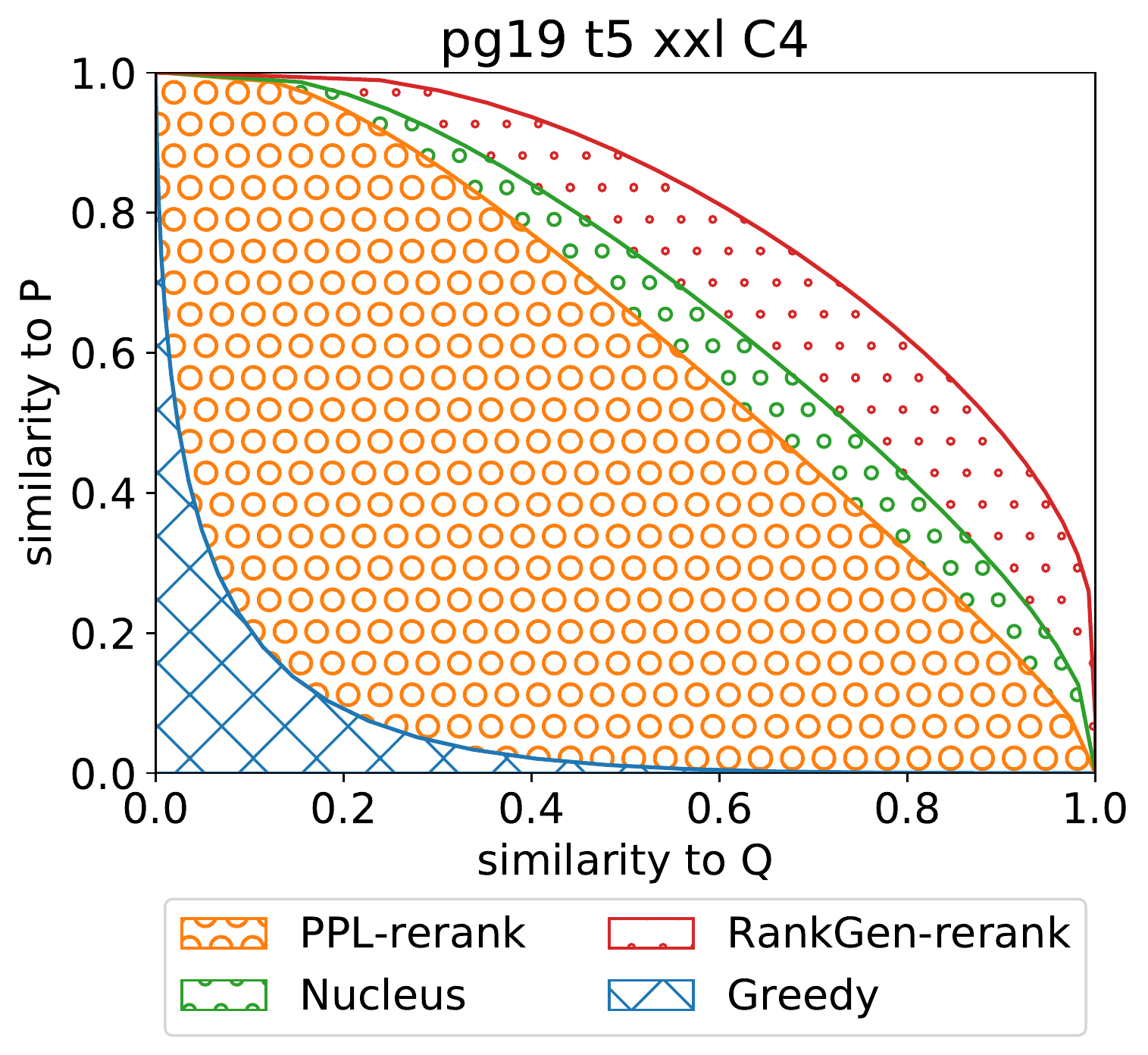}
    \caption{Divergence curves~\citep{pillutla2021mauve} after full sample re-ranking on PG19 inputs using \model-XL trained on PG19. The area under this curve is the MAUVE score. Overall, we see that \model~ makes fewer Type I (bigger intercept with $y=1$ line) and Type II style errors (bigger intercept with $x=1$). PPL re-ranking increases the amount of repetition in generated text (\tableref{tab:rep-scores}), leading to more Type I errors (smaller intercept with $y=1$).}
    \label{fig:pg19-divergence-curves}
\end{figure*}

In \figureref{fig:wiki-divergence-curves} and \figureref{fig:pg19-divergence-curves} we plot the divergence curves comparing greedy decoding, nucleus sampling, and full sample re-ranking with perplexity and \model. We observe that re-ranking with \model~increases the area under the curve, whereas re-ranking with model perplexity reduces the area. Re-ranking with \model~reduces both Type I (bigger intercept on $y=1$) and Type II errors (bigger intercept on $x=1$). Re-ranking with perplexity leads to higher Type I errors, or more repetition (as also observed in \appendixref{sec:token-overlap}).

\subsection{Token Overlap metrics}
\label{sec:token-overlap}

In addition to the MAUVE scores calculated in \sectionref{sec:experiments}, we measure token overlap statistics comparing different decoding methods. First, we measure the \textbf{rep} metric from~\citet{welleck2019neural}, which is an approximate measurement of the amount of repetition in generated text. We measure the percentage of generated tokens which are exactly copied from the immediate local prefix of 20 tokens. In \tableref{tab:rep-scores} we find that re-ranking with \model~slightly reduces \textbf{rep} compared to nucleus sampling (18.9 vs 19.5). We get even lower repetition on the \model~trained on just generative negatives (17.8), while \model~trained on just inbook negatives gets 20.0 --- thus generative negatives are better at reducing repetition. Re-ranking with perplexity increases \textbf{rep} to 23.9, whereas greedy decoding has the highest repetition of 59.5. This is consistent with recent findings of repetition in greedy decoded outputs~\citep{holtzman2020curious,zhang-etal-2021-trading}. Human text is the least repetitive, with a \textbf{rep} score of 15.4.

Next, we measure the fraction of unigrams in the generation which are also present in the prefix. Higher scores could either imply more faithfulness to the prefix (less hallucination), or lower amounts of abstraction. We present two versions of this metric --- (1) considering all tokens (\tableref{tab:token-overlap-prefix-suffix}); (2) considering only only lemmatized nouns and numbers (\tableref{tab:token-overlap-prefix-suffix-entities}). Overall, we find that re-ranking samples with \model~slightly increases this overlap score (19.5 vs 21.7), but re-ranking by token overlap (38.4) or perplexity (25.0) leads to a much higher score. Given the lower MAUVE scores for these two approaches (\tableref{tab:mauve-scores-full-rerank}), we suspect that token overlap / perplexity re-ranking leads to lower amounts of abstraction / repetitiveness. Human written text has the lowest overlap, perhaps indicating more abstractive text.
\section{Ablation Studies}
\label{appendix:ablations}

We conduct several ablation studies studying the importance of three aspects --- (1) model size; (2) minibatch size, or number of negative samples during contrastive learning; (3) the type of negative samples (inbook, generative or both). Overall, we see clear benefits of increasing model size and increasing minibatch size for suffix identification (\tableref{tab:suffix-id-results-ablations}, \tableref{tab:gold-beats-neg-model-ablations}) and human-text identification (\tableref{tab:gold-beats-generation-ablations}). We see a similar, but less prominent trend on MAUVE scores after re-ranking generations (\tableref{tab:mauve-scores-full-rerank-ablations}). For some settings we find that the \model-large variant produces slightly better generations than \model-XL. We hypothesize this is due to the much larger minibatch used to train \model-large models (4096) compared to \model-XL (1536) due to memory constraints.

\section{More Model Generations}
\label{sec:more-generations}

More model generations with human explanations are provided in \tableref{tab:model-outputs-with-scarecrow-2} to  \tableref{tab:more-model-generations-5}. See our Github repository\footnotemark[1] for all 600 annotations for the 200 generation pairs.

\begin{figure*}
\centering
\begin{lstlisting}[language=Python]
def rankgen_search(prefix, scorer, generator,
                   rerank_length, beam_size, samples_per_beam):
  all_beams = [""]
  for _ in range(0, MAX_LENGTH, rerank_length):
    # concatenate input prefix with current beams
    all_inputs = [prefix + " " + beam for beam in all_beams]
    # for each beam, generate next rerank_length tokens.
    # samples_per_beam hypotheses are generated per beam,
    # making a total of (num_beams * samples_per_beam) hypotheses
    hypotheses = generator(all_inputs,
                           num_new_tokens=rerank_length,
                           num_samples=samples_per_beam)
    # measure RankGen score between prefix and each hypothesis
    scores = scorer(prefix, hypotheses)
    # take top-K scores where K=beam size
    top_indices = np.argsort(-1 * scores)[:beam_size]
    all_beams = [outputs[x] for x in top_indices]
  return all_beams
\end{lstlisting}

\caption{A simplified Python implementation showing our \model~beam search algorithm (without minibatching). For every \texttt{rerank\_length} tokens, a generator suggests hypotheses and the \model~scorer ranks them. The top \texttt{beam\_size} hypotheses are retained for the next stage of generation and re-ranking.}
\label{fig:python-code-beam-search}
\end{figure*}

\begin{table*}[t!]
\small
\begin{center}
\begin{tabular}{ lrr|rr|rr|rr|r } 
 \toprule
 & \multicolumn{8}{c}{\textbf{Generator Language Model}} \\
 \cmidrule{2-9} \vspace{-0.3cm} \\
 & \multicolumn{2}{c|}{GPT2-md} & \multicolumn{2}{c|}{GPT2-XL} & \multicolumn{2}{c|}{T5-XXL-PG19} & \multicolumn{2}{c|}{T5-XXL-C4} & Average \\
\textbf{Decoding method} & PG19 & wiki & PG19 & wiki & PG19 & wiki & PG19 & wiki \\
 \midrule
 Human Text & 15.8 & 15.0 & 15.8 & 15.0 & 15.8 & 15.0 & 15.8 & 15.0 & 15.4 \\
 Greedy decoding & 71.4 & 56.6 & 66.8 & 51.6 & 55.6 & 52.7 & 67.6 & 53.7 & 59.5 \\
 Nucleus, $p=0.9$~\shortcite{holtzman2020curious} & 21.8 & 18.8 & 22.4 & 19.5 & 17.7 & 17.4 & 20.3 & 18.4 & 19.5 \\
 
 Top-k, $k=40$~\shortcite{fan-etal-2018-hierarchical} & 19.4 & 17.0 & 19.9 & 19.7 & 17.9 & 17.9 & 20.4 & 18.6 & 18.9 \\
 Typical, $p=0.9$~\shortcite{meister2022typical} & 21.6 & 18.6 & 22.2 & 19.5 & 17.6 & 17.4 & 20.3 & 18.5 & 19.5  \\
 
 \midrule
 \multicolumn{3}{l}{\ul{\textbf{Re-ranking 20 nucleus samples}}}\vspace{0.15cm} \\
 Unigram overlap & 22.2 & 19.9 & 22.9 & 20.6 & 19.0 & 18.7 & 21.5 & 19.8 & 20.6 \\
 LM perplexity & 26.9 & 23.2 & 27.9 & 24.3 & 20.4 & 21.5 & 24.6 & 22.5 & 23.9 \\
 
  \model~PG-XL-gen & 20.0 & 17.2 & 20.5 & 17.9 & 16.3 & 15.8 & 18.3 & 16.6 & 17.8 \\
  \model~PG-XL-inbook & 22.1 & 19.5 & 22.7 & 20.0 & 18.2 & 17.8 & 20.7 & 18.6 & 20.0 \\
  \model~PG-XL-both & 20.9 & 18.4 & 21.6 & 19.2 & 17.4 & 16.9 & 19.7 & 18.2 & 19.0 \\
  \model~all-XL-both & 20.5 & 18.6 & 21.1 & 19.4 & 17.3 & 16.6 & 19.5 & 18.2 & 18.9 \\
\bottomrule
\end{tabular}
\end{center}
\vspace{-0.1in}
\caption{Fraction of generated tokens which are copied from the previous 20 tokens, \emph{roughly measuring the amount of repetition} in text (the \textbf{rep} metric from~\citealp{welleck2019neural}). Overall we find that ranking samples with \model~reduces repetition, whereas ranking with perplexity increases repetition. Greedy decoded outputs are the most repetitive, whereas human-written text is the least repetitive.}
\label{tab:rep-scores}
\end{table*}

\begin{table*}[t!]
\small
\begin{center}
\begin{tabular}{ lrr|rr|rr|rr|r } 
 \toprule
 & \multicolumn{8}{c}{\textbf{Generator Language Model}} \\
 \cmidrule{2-9} \vspace{-0.3cm} \\
 & \multicolumn{2}{c|}{GPT2-md} & \multicolumn{2}{c|}{GPT2-XL} & \multicolumn{2}{c|}{T5-XXL-PG19} & \multicolumn{2}{c|}{T5-XXL-C4} & Average \\
\textbf{Decoding method} & PG19 & wiki & PG19 & wiki & PG19 & wiki & PG19 & wiki \\
 \midrule
 Human Text & 14.0 & 20.7 & 14.0 & 20.7 & 14.0 & 20.7 & 14.0 & 20.7 & 17.4 \\
 Greedy decoding & 16.1 & 25.5 & 15.9 & 25.0 & 15.8 & 21.0 & 20.0 & 27.3 & 20.8 \\
 Nucleus, $p=0.9$~\shortcite{holtzman2020curious} & 16.7 & 22.8 & 17.3 & 23.7 & 14.0 & 19.0 & 17.8 & 24.8 & 19.5 \\
 
 Top-k, $k=40$~\shortcite{fan-etal-2018-hierarchical} & 15.6 & 21.0 & 15.8 & 15.9 & 15.1 & 20.2 & 19.3 & 25.7 & 18.6 \\
 Typical, $p=0.9$~\shortcite{meister2022typical} & 16.6 & 22.5 & 17.2 & 23.8 & 14.1 & 18.8 & 18.0 & 25.0 & 19.5 \\
 \midrule
 \multicolumn{3}{l}{\ul{\textbf{Re-ranking 20 nucleus samples}}}\vspace{0.15cm} \\
 Unigram overlap &33.6 & 43.5 & 34.4 & 45.7 & 28.9 & 34.1 & 39.9 & 47.0 & 38.4 \\
 
 LM perplexity & 19.9 & 29.4 & 20.2 & 30.2 & 16.9 & 22.7 & 27.3 & 33.1 & 25.0 \\
  \model~PG-XL-gen & 18.8 & 25.5 & 19.3 & 26.5 & 14.6 & 20.0 & 20.9 & 26.6 & 21.5 \\

  \model~PG-XL-inbook & 18.8 & 25.1 & 19.4 & 26.4 & 15.9 & 21.0 & 19.7 & 26.5 & 21.6 \\
  \model~PG-XL-both & 19.4 & 25.2 & 19.7 & 26.5 & 15.7 & 21.3 & 21.2 & 26.7 & 22.0\\
  
  \model~all-XL-both & 19.1 & 24.8 & 19.5 & 26.1 & 15.7 & 21.3 & 20.4 & 26.3 & 21.7 \\
\bottomrule
\end{tabular}
\end{center}
\vspace{-0.1in}
\caption{Percentage of unigrams in generation also present in the prefix. Overall, we see that re-ranking nucleus samples with \model~increases this overlap, but not as much as re-ranking with LM perplexity. Human text has the lowest overlap, which we hypothesize is due to higher amounts of abstraction.}
\label{tab:token-overlap-prefix-suffix}
\end{table*}

\begin{table*}[t!]
\small
\begin{center}
\begin{tabular}{ lrr|rr|rr|rr|r } 
 \toprule
 & \multicolumn{8}{c}{\textbf{Generator Language Model}} \\
 \cmidrule{2-9} \vspace{-0.3cm} \\
 & \multicolumn{2}{c|}{GPT2-md} & \multicolumn{2}{c|}{GPT2-XL} & \multicolumn{2}{c|}{T5-XXL-PG19} & \multicolumn{2}{c|}{T5-XXL-C4} & Average \\
\textbf{Decoding method} & PG19 & wiki & PG19 & wiki & PG19 & wiki & PG19 & wiki \\
 \midrule
 Human Text & 19.6 & 27.3 & 19.6 & 27.3 & 19.6 & 27.3 & 19.6 & 27.3 & 23.4 \\
 Greedy decoding & 23.8 & 31.1 & 23.0 & 30.5 & 21.8 & 26.2 & 26.5 & 33.2 & 27.0 \\
 Nucleus, $p=0.9$~\shortcite{holtzman2020curious} & 23.8 & 29.7 & 24.2 & 30.3 & 19.3  & 24.4 & 24.6 & 31.6 & 26.0 \\
 
 Top-k, $k=40$~\shortcite{fan-etal-2018-hierarchical} & 22.0 & 27.6 & 22.2 & 28.7 & 21.0 & 26.4 & 27.1 & 33.2 & 26.0 \\
 Typical, $p=0.9$~\shortcite{meister2022typical} & 23.7 & 29.2 & 24.2 & 30.3 & 19.4 & 24.5 & 24.8 & 32.0 & 26.0 \\
 \midrule
 \multicolumn{3}{l}{\ul{\textbf{Re-ranking 20 nucleus samples}}}\vspace{0.15cm} \\
 Unigram overlap & 42.0 & 51.0 & 42.4 & 52.9 & 35.1 & 41.0 & 47.4 & 54.7 & 45.8 \\
 
 LM perplexity & 27.8 & 35.1 & 27.1 & 35.4 & 23.0 & 28.9 & 35.2 & 39.2 & 31.4  \\
  \model~PG-XL-gen &  26.3 & 32.6 & 26.5 & 33.4 & 20.4 & 26.5 & 28.6 & 34.2 & 28.6 \\

  \model~PG-XL-inbook & 26.5 & 32.7 & 26.9 & 34.1 & 21.8 & 27.7 & 27.4 & 34.2 & 28.9 \\
  \model~PG-XL-both & 27.0 & 32.8 & 27.5 & 33.9 & 21.8 & 28.0 & 29.2 & 34.5 & 29.3 \\
  
  \model~all-XL-both & 27.0 & 32.6 & 27.3 & 33.7 & 21.7 & 28.0 & 28.4 & 34.0 & 29.1\\
\bottomrule
\end{tabular}
\end{center}
\vspace{-0.1in}
\caption{A version of \tableref{tab:token-overlap-prefix-suffix} considering only lemmatized nouns, proper nouns and numbers, with similar trends.}
\label{tab:token-overlap-prefix-suffix-entities}
\end{table*}

\begin{table*}[h!]
\small
\begin{center}
\begin{tabular}{ lrrrrrrrrrrr } 
 \toprule
 Model & Batch & \multicolumn{2}{c}{ChapterBreak} & \multicolumn{2}{c}{StoryCloze} & Hella & \multicolumn{5}{c}{RELiC (Recall@k)}  \\
  Size & \multicolumn{1}{l}{Size} & PG19 & AO3 & 2016 & 2018 & Swag & 1 & 3 & 5 & 10 & 50 \\
 \midrule 
 \multicolumn{6}{l}{\textit{(\model~models trained on PG19)}} \vspace{-0.1in} \\\\ 
 base & 4096 &  57.7 & 36.0 &  67.6 & 68.7 & 30.7 & 3.8 & 8.2 & 10.8 & 15.4 & 31.6\\
 large & 4096 & 60.6 & 31.9  & 69.3 & 69.8 & 34.2 & \textbf{5.7} & \textbf{11.0} & \textbf{14.5} & \textbf{20.0} & \textbf{36.6}\\
  XL & 1536 & \textbf{63.5} & \textbf{36.9} & \textbf{71.1} & \textbf{72.6} & \textbf{40.7} & 4.5 & 8.4 & 11.0 & 15.1 & 27.9\\
  \midrule
  \multicolumn{6}{l}{\textit{(\model~models trained on all 4 domains)}} \vspace{-0.1in} \\\\ 
 base & 4096 & 48.1 & 33.0 &  69.0 & 69.1 & 34.0 & 3.1 & 6.2 & 8.3 & 11.8 & 25.6\\
 large & 4096 & 51.4 & 31.1 & 70.3 & 71.7 & 40.6 & 3.7 & 7.3 & 9.5 & 13.1 & 25.8\\
  XL & 256 & 38.2 & 28.3 & 70.6 & 68.5 & 35.9 & 2.8 & 5.6 & 7.4 & 10.8 & 22.9 \\
 XL & 512 & 47.3 & 31.3 & 72.3 & 69.8 & 39.3 &  3.3 & 7.1 & 9.7 & 13.6 & 26.5 \\
 XL & 768 & 45.2 & 30.1 & 72.5 & 71.2 & 41.4 & 3.8 & 7.2 & 9.6 & 13.7 & 27.5\\
 XL & 1536 & \textbf{59.3} & \textbf{32.8} & \textbf{75.4} & \textbf{75.8} & \textbf{46.3} & \textbf{4.9} & \textbf{9.2} & \textbf{11.9} & \textbf{16.5} & \textbf{31.5} \\
 \bottomrule
\end{tabular}
\end{center}
\caption{Variation in performance on existing suffix identification and literary retrieval datasets with model size and minibatch size (number of negative samples). Overall, we see that scaling both model size and minibatch size improves suffix identification performance. See \tableref{tab:suffix-id-results} for comparisons with non-\model~baselines.}
\label{tab:suffix-id-results-ablations}
\end{table*}

\begin{table*}[t!]
\small
\begin{center}
\begin{tabular}{ lrrrrr|rrrr } 
 \toprule
 Model & Batch & \multicolumn{2}{c}{pg19-random} & \multicolumn{2}{c|}{pg19-hard} & \multicolumn{2}{c}{wiki-random} & \multicolumn{2}{c}{wiki-hard} \\
Size & \multicolumn{1}{l}{Size} & 2-way & 11-way & 2-way & 11-way & 2-way & 11-way & 2-way & 11-way\\
 \midrule 
 \multicolumn{6}{l}{\textit{(\model~models trained on PG19)}} \vspace{-0.1in} \\\\ 
 base & 4096 & 98.6 & 91.7 & 69.4 & 36.8  & 88.4 & 57.0 & 65.6 & 25.7 \\
 large & 4096 & 99.0 & 94.2 & 76.0 & 46.4 & 91.3 & 66.3 & 69.7 & 32.7 \\
 XL & 1536 & \textbf{99.1} & \textbf{94.4} & \textbf{78.0} & \textbf{49.5} & \textbf{92.3} & \textbf{69.0} & \textbf{71.4} & \textbf{35.7}\\
 \midrule
   \multicolumn{6}{l}{\textit{(\model~models trained on all 4 domains)}} \vspace{-0.1in} \\\\ 
 base & 4096 & 97.9 & 88.4 & 63.5 & 29.8 & 95.6 & 77.8 & 74.7 & 42.3  \\
 large & 4096 & 98.6 & 92.1 & \textbf{68.6} & 39.3 & 97.0 & 83.7 & \bf 79.1 & 50.7 \\
 XL & 256  & 96.8 & 83.7 & 60.3 & 26.0 & 95.0 & 75.9 & 73.5 & 39.8 \\
 XL & 512 & 97.7 & 87.8 & 63.1 & 31.6 & 96.1 & 80.0 & 76.0 & 45.0 \\
 XL & 768 & 98.1 & 89.7  & 64.7 & 34.2 & 96.6 & 82.1 & 77.6 & 48.2 \\
 XL & 1536 & \textbf{98.7} & \textbf{92.6} & 61.3* & \textbf{39.5}* & \textbf{97.3} & \bf 84.6 & 77.2* & \bf 52.1* \\
\bottomrule
\end{tabular}
\end{center}
\vspace{-0.1in}
\caption{Variation in performance on our PG19 / Wikipedia suffix identification datasets with model size and minibatch size (number of negative samples). Overall, we see that scaling both model size and minibatch size improves suffix identification performance. See \tableref{tab:gold-beats-neg-main} for comparisons with non-\model~baselines. * Note that these numbers are lower since hard sets were adversarially constructed using this \model~variant. }
\vspace{-0.1in}
\label{tab:gold-beats-neg-model-ablations}
\end{table*}

\begin{table*}[t!]
\small
\begin{center}
\begin{tabular}{ lrrrrrr } 
 \toprule
  & & \multicolumn{4}{c}{\textbf{Generator Language Model} (re-ranking 20 nucleus samples)} \\
 \cmidrule{3-6} \vspace{-0.3cm} \\
 & batch size & GPT2-md & GPT2-XL & T5-XXL-PG19 & T5-XXL-C4 & Average \\
 \midrule
 \multicolumn{7}{l}{\textit{(\model~models trained on PG19 and evaluated on PG19 prefixes)}} \vspace{-0.1in} \\\\ 
 base & 4096 & \textbf{78.4} & \textbf{77.5} & \textbf{94.6} & 72.2 & 80.7 \\
 large & 4096 & 77.1 & \textbf{77.6} & 93.4 & 73.4 & 80.4 \\
 XL & 1536 & 76.3 & 75.2 & \textbf{94.3} & \textbf{80.7} & \textbf{81.6}  \\
 \midrule
  \multicolumn{7}{l}{\textit{(\model~models trained on all 4 domains and evaluated on Wikipedia prefixes)}} \vspace{-0.1in} \\\\ 
 base & 4096 & 83.8 & 83.0 & 90.1 & 87.4 & 86.1\\
 large & 4096 & \bf 86.3 & \textbf{85.8} & \textbf{92.0} & \textbf{88.5} & \textbf{88.1} \\
  XL & 256 & 81.5 & 84.2 & 89.7 & 87.9 & 85.8 \\
  XL & 512 & 82.5 & 84.5 & 90.2 & 87.3 & 86.1 \\
  XL &768 & 81.0 & \bf 85.1 & 89.7 & 87.8 & 85.9\\
  XL & 1536 & 83.9  & \bf 85.7 & \bf 91.8 & \bf 88.1 & \bf 87.3 \\
\bottomrule
\end{tabular}
\end{center}
\vspace{-0.1in}
\caption{Variation in MAUVE score of top-ranked generation (among 20 nucleus samples with $p=0.9$) using \model~variants having a different model / minibatch size. On average, increasing model size and minibatch size boosts performance, but the trend is less prominent than in other tasks. However, all \model~variants outperform baselines like nucleus sampling (see \tableref{tab:mauve-scores-full-rerank} for details).}
\vspace{-0.1in}
\label{tab:mauve-scores-full-rerank-ablations}
\end{table*}

\begin{table*}[t!]
\small
\begin{center}
\begin{tabular}{ lrrrrrr } 
 \toprule
 Model & batch size & GPT2-md & GPT2-XL & T5-XXL-PG19 & T5-XXL-C4 & Average \\
 \midrule
  \multicolumn{7}{l}{\textit{(\model~models trained on PG19 and evaluated on PG19 prefixes)}} \vspace{-0.1in} \\\\ 
 PG19-base & 4096 & 84.4 & 78.3 & 68.3 & 70.9 & 75.5 \\
 PG19-large & 4096 & 93.7 & 87.9 & 79.1 & 81.3 & 85.5 \\
 PG19-XL & 1536 & \textbf{97.4} & \textbf{93.7} & \textbf{87.4} & \textbf{89.7} & \textbf{92.1} \\
 \midrule
  \multicolumn{6}{l}{\textit{(\model~models trained on all 4 domains and evaluated on Wikipedia prefixes)}} \vspace{-0.1in} \\\\ 
 all-base & 4096 & 71.9 & 68.2 & 88.2 & 60.0 & 72.1\\
 all-large & 4096 & 80.4 & 74.7 & 93.0 & 64.7 & 78.2 \\
all-XL& 256&  73.4 & 68.8 & 88.8 & 60.7 & 72.9\\
all-XL & 512 & 78.5 & 73.6 & 93.1 & 64.3 & 77.4 \\
all-XL & 768 & 81.9 & 76.1 & \textbf{95.4} & 65.8 & 79.8 \\
 all-XL & 1536 & \textbf{84.5} & \textbf{78.0} & 95.3 & \textbf{67.3} & \textbf{83.7} \\
\bottomrule
\end{tabular}
\end{center}
\vspace{-0.1in}
\caption{Variation in human-written text identification (vs machine generated with $p=0.9$) performance with model size and minibatch size (number of negative samples). Overall, we see that scaling both model size and minibatch size improves human text identification performance. See \tableref{tab:gold-beats-generation-main} for comparisons with causal LMs.}
\vspace{-0.1in}
\label{tab:gold-beats-generation-ablations}
\end{table*}

\begin{table*}[t!]
\footnotesize
\begin{center}
\begin{tabular}{ p{14.8cm} } 
\toprule
We are currently looking for people with some experience in English content writing / teaching / editing to read a prompt text (~200-250 words) and choose which of two article fragments (70-100 words each) is a valid continuation of the prompt text. This study is a part of a bigger academic research project on text evaluation. If you decide to help us in this project, you will be asked to:
- set up an account on Amazon Mechanical Turk Sandbox (this is what we use as the interface, payment will be through Upwork only)
- read and evaluate two sets of 200 fragments, choosing which fragment is a better continuation of the prompt. You will NOT need to go through complicated and lengthy guidelines.  You do NOT need to provide any written feedback on each story fragment, and you do NOT need to mark mistakes or edit the article fragments. Simply choose the fragment which continues the context better. The budget we have for this project is \$100, which is calculated assuming a \$25/h rate (calculated based on the average time per story fragment from the data we have already collected). \newline 

\textbf{\emph{Additional instructions for adding explanations:}}

In this task you need to choose which better completion is better, along with 2-3 sentences explaining why you felt so. Some examples of this kind of annotation --- (1) Text 1; Text 1 is more relevant to the context because .... (2) Text 2; Both texts are relevant to the context, but Text 1 has lesser repetitions and is more coherent because .... (3) Text 2; Text 2 does not contradict itself like Text 1. In general it would be great if you quote certain parts of the context / continuation to support your argument.. for instance --- The context talks about the adventures of Frodo, and how he they started after "he inherited the ring from Bilbo". Text 1 goes on to talk about how Bilbo "suddenly left on his birthday" which "gave the ring to Frodo", whereas Text 2 contradicts the context by saying "Bilbo went out for the adventure with the ring."\\ 
\bottomrule
\end{tabular}
\end{center}
\vspace{-0.1in}
\caption{The job posting and instructions shown to Upworkers before they performed the annotation task.}
\vspace{-0.1in}
\label{tab:upwork-instructions}
\end{table*}

\begin{table*}[t!]
\small
\begin{center}
\begin{tabular}{ p{5.4cm}p{5.2cm}p{4.3cm} } 
\toprule
\bf Prefix & \bf Generations & \bf Annotator Preference \\
\midrule
 \textbf{PG19}, \href{https://www.gutenberg.org/files/2547/2547-0.txt}{\emph{Half a Life-time Ago}}, by \emph{Elizabeth Gaskell}: ... If \textcolor{mycolor1}{thou} doesn’t choose to marry me on those terms--\textcolor{mycolor1}{why!} I can snap my fingers at \textcolor{mycolor1}{thee}, never fear. I’m not so far gone in love as that. But I will not have thee, if \textcolor{mycolor1}{thou say’st} in such a hectoring way that Willie must go out of the house--and the house his own too--before \textcolor{mycolor1}{thou’lt} set foot in it. Willie bides here, and I bide with him.” “\textcolor{mycolor1}{Thou hast} may-be spoken a word too much,” said Michael, \textcolor{mycolor2}{pale with rage}. & Text 1 (\textbf{Nucleus}): “How on earth could it be? He must be part of the marriage, \textcolor{mycolor2}{my love}. But he can’t go away--he can’t go away with me. And if he does go away, he’ll have to pay ... \newline Text 2 (\textbf{\model}): “Take heed how \textcolor{mycolor1}{thou handlest thy} words in the future.” “I haven’t spoken a word too much,” she murmured, coolly. “I meant--I meant the last word!” “Swear it, \textcolor{mycolor1}{thou hastna} sworn a word yet. I’ll go to talk with ...  & \textbf{Text 2}. Text 1 has a completely different \textcolor{mycolor1}{style}, way more modern. First sentence in Text 2 fits the tone of enraged character, Michael. \newline \textbf{Text 2}. The writing style is more similar, Text 1 sounds too modern. Plus, the atmosphere of the fragment is more consistent. The characters \textcolor{mycolor2}{seem to be having an argument, so "My love"} in doesn't quite make sense. \\
\midrule
\textbf{PG19}, \href{https://www.gutenberg.org/files/48089/48089-h/48089-h.htm}{\emph{The Fall of a Nation}} by \emph{Thomas Dixon}: The grim \textcolor{mycolor2}{gray wave of destruction} from the sand dunes had rolled into \textcolor{mycolor2}{battleline} and spread out over the green clothed hills and valleys of the Island--swiftly, remorselessly, with an uncanny precision ... a puff of black smoke streamed downward and the distant officer, ... \textcolor{mycolor2}{gunners of his battery. Our rifles cracked in vain.} The birdmen laughed and paid no attention. & Text 1 (\textbf{\model}): They raced across the plains and away. Our artillery fell silent and rested. It would have to be our last salvo. "Are they coming down here?” shouted an American, as he watched ... \newline Text 2 (\textbf{Nucleus}): \textcolor{mycolor2}{With a bark of laughter}, a group of strong men fell among the men and \textcolor{mycolor2}{laughed with them}. And with the \textcolor{mycolor2}{general smile} on his face he began to wave his finger in the air at them and ... & \textbf{Text 1}. \textcolor{mycolor2}{The jolly atmosphere of Text 2 really doesn't fit} with the prefix. The prefix read together with Text 2 has kind of a \textcolor{mycolor2}{Monty Python vibe} to it. \newline \textbf{Text 1.} Chose the first one, because of the \textcolor{mycolor2}{stark change in the tone in the second text}, which has fair amount of laughter and sleeping during a battle. \\
\midrule 
\textbf{Wiki,} \href{https://en.wikipedia.org/wiki/Tim\_Richmond}{\emph{Tim Richmond}}: ... Richmond raced in a \textcolor{mycolor1}{1978} Mini Indy car event at Phoenix International Raceway, winning the Formula Super Vee support event in a Lola T620. The win attracted sponsors and attention from ... He also competed in USAC's Silver Crown series. Richmond's father bought an Eagle Indy Car chassis and an Offenhauser engine for the 1979 race at Michigan International Speedway. \textcolor{mycolor1}{Richmond qualified 21st fastest with a lap, significantly slower than Bobby Unser's pole position speed.} & Text 1 (\textbf{\model}): However, \textcolor{mycolor1}{his effort earned him an invitation to join the 1979 Indy 500} at Phoenix International Raceway. After finishing sixth, Richmond was called upon to replace the injured Jimmy Corder. A rookie, Richmond began his race in a three-car pack ... \newline Text 2 (\textbf{Nucleus}): \textcolor{mycolor1}{In 1982 the pair switched to the SuperCar} chassis that year. As a result of the change, Richmond's driving style evolved somewhat. At age 42 he returned to IndyCar. At the \textcolor{mycolor1}{1977} ... & \textbf{Text 1} - \textcolor{mycolor1}{Chronology - the events of text 1 follow prefix text and are before the events of text 2}. It makes sense the text would be written in the correct order. \newline \textbf{Text 1} - \textcolor{mycolor1}{Text 1 continues the idea of the final sentence of the prefix} paragraph. Despite his unimpressive position in the qualifying race, his effort earned him an invitation into the Indy 500.\\
\bottomrule
\end{tabular}
\end{center}
\vspace{-0.1in}
\caption{A continuation of \tableref{tab:model-outputs-with-scarecrow} with representative model generations using \model~vs nucleus sampling~\citep{holtzman2020curious}, along with human explanations (from English teachers/writers) for preferring \model~outputs. For every row the \textcolor{mycolor2}{color coding} is used to ground the annotator explanation in the prefix and generation. See \tableref{tab:more-model-generations}-\ref{tab:more-model-generations-5} for full-length generations.}
\label{tab:model-outputs-with-scarecrow-2}
\vspace{-0.1in}
\end{table*}

\begin{table*}[t!]
\footnotesize
\begin{center}
\begin{tabular}{ p{14.8cm} } 
\toprule

\textbf{Prefix} (\textbf{PG19}) \href{https://www.gutenberg.org/files/42142/42142-0.txt}{\emph{A Young Inventor’s Pluck}} by \emph{Arthur M. Winfield}: "Oh, I only wanted to know," replied the young man, apparently unconcerned. "Pooler said I was never to speak of it," returned Meg. "He thinks I don't know where it is, but I do." "Show it to us, will you?" asked Mont, eagerly. Meg shrank back. "It's haunted!" she exclaimed. "Besides, Pooler would kill me if he found it out." "We'll never tell him, will we, Jack?" "Certainly not," replied the young machinist, to whom this new move was only another mystery. "Besides, we intend to be your friends, you know," continued Mont. "Well, then, come on," replied Meg, finally. "I don't care if he likes it or not. 'Taint a very nice place though." "Never mind. We can't find any worse or more dangerous places than we have already gone through." Meg picked her way along one bank of the island for nearly a quarter of a mile. Jack and Mont followed closely.\\\\

\textbf{Text 1 (\model, T5-XXL-C4)}: The path led through an area of woods, shrubbery and scrub oak. Neither meadow nor open ground was seen. Jack called out as they scampered around an occasional rough rock, often from the height of a dozen feet. "I think I can see it now," said Meg, suddenly, as they rounded a grassy slope and followed another bank. A few yards farther on she stopped, and glared at the clear sky and the undisturbed valley. "That's the place," she said. The young man could tell she had a point.\\
\textbf{Text 2 (Nucleus, T5-XXL-C4)}: Jack would not let her out of his sight. He made a sharp look at Meg as she walked forward. Then, his expression became grave. "Meg," he whispered, "are we going to the haunted house or not?" "This is the haunted house, Jack," she replied. "Is it haunted?" "Of course it's haunted," returned Mont, "but if you were so scared you would be shaking all over." "Of course it's haunted," replied the young man.\\\\
\textbf{Human Annotation 1}: Text 1. Second text appears to omit a part when they get to their destination having them appear at their destination instantly. Although a haunted hous sounds more in-line with the sample text than a haunted valley, the first text fits more naturally with the sample text.\\
\textbf{Human Annotation 2}: Text 1 - Meg is taking them somewhere - so leading them would make sense. 'That's the place' indicates she had found it. Jack does not sound worried in prefix text, so it would not really fit that he doesn't let Meg out of his sight as indicated in  Text 2. \\
\textbf{Human Annotation 3}: Text 2. The collocations and style in Text 1 sound too colloquial and off when compared to the prefix. "(...) he had a point. "\\\\
\midrule 
\midrule \\
\textbf{Prefix} (\textbf{PG19}) \href{https://www.gutenberg.org/files/42142/42142-0.txt}{\emph{A Young Inventor’s Pluck}} by \emph{Arthur M. Winfield}: But at that moment, he did not stop to question how the young man had gotten into a position similar to his own. His one thought was whether or not his friend was alive. He placed the body as comfortably as possible on the fork of the tree, and then watched eagerly for some sign of life. There was a cut upon Mont's brow, and presently a few drops of blood oozed from the wound, and trickled down his cheek. Jack took this as a good sign, and he was not mistaken, for a few seconds later the young man gave a deep sigh, and slowly opened his eyes. "Where am I?" he asked, faintly. "Here with me, Mont," replied the young machinist, bending over him. "Don't you know me?" "Jack!" was the low response. "Oh, how my head hurts! Where are we?" "On the river." "We are?" Mont attempted to rise, but Jack pushed him back. "Oh, I remember now!" he continued, shuddering. "Remember what?" \\\\
\textbf{Text 1 (Nucleus, GPT2-md)}: "Mont, when I was twenty-two, I met some girls who took care of my grandfather, and when I wanted to give a hand in the shop, I got him to promise to do it for me for free, if he liked me, or give me money. And now, when I tell you I feel a sense of belonging, Jack, I have been a slave for four hundred years. I have been born of slavery, and yet I feel like a slave." Jack listened to this.\\
\textbf{Text 2 (\model, GPT2-md)}: "That time I tried to drag you off my back and then went back on to kill myself." "That is wrong," replied the young machinist, trembling. "I was in your arms. I never intended to kill myself. It happened." "It was a mistake, then," said Jack with a shake of his head. "That was just the first of many," said Mont. The young man looked at him sadly. "Don't lie, Jack," he said gently. "You are not dead. You are alive. What's going on? \\\\
\textbf{Human Annotation 1}: Text 2 - Text 1 seems too mundane to be the memory on waking after being knocked out. Text 2 would fit a little better perhaps. \\
\textbf{Human Annotation 2}: Text 2 - Text 2 was chosen via deductive reasoning. Text 1 mentions ideas that have nothing to do with the prefix paragraph. \\
\textbf{Human Annotation 3}: Text 2. Text 2 answers the question that was asked at the end of the paragraph and continues with the same characters. \\
\bottomrule
\end{tabular}
\end{center}
\vspace{-0.1in}
\caption{More model generations (Part 1) from each decoding method with human preferences and explanations. See our attached data submission for all 600 annotations for the 200 generation pairs.}
\vspace{-0.1in}
\label{tab:more-model-generations}
\end{table*}

\begin{table*}[t!]
\footnotesize
\begin{center}
\begin{tabular}{ p{14.8cm} } 
\toprule
\textbf{Prefix} (\textbf{PG19}) \href{https://www.gutenberg.org/files/876/876-h/876-h.htm}{\emph{Life in the Iron-Mills}} by \emph{Rebecca Harding Davis}: While she was skinning the potatoes, and munching them, a noise behind her made her stop. "Janey!" she called, lifting the candle and peering into the darkness. "Janey, are you there?" A heap of ragged coats was heaved up, and the face of a young girl emerged, staring sleepily at the woman. "Deborah," she said, at last, "I'm here the night." "Yes, child. Hur's welcome," she said, quietly eating on. The girl's face was haggard and sickly; her eyes were heavy with sleep and hunger: real Milesian eyes they were, dark, delicate blue, glooming out from black shadows with a pitiful fright. "I was alone," she said, timidly. "Where's the father?" asked Deborah, holding out a potato, which the girl greedily seized. "He's beyant,--wid Haley,--in the stone house." (Did you ever hear the word \_jail\_ from an Irish mouth?) "I came here. \\\\

\textbf{Text 1 (Nucleus, T5-XXL-C4)}: I needed to find my feet." "What have you done here?" Deborah was amazed. "Whoever came in here yesterday found his feet," she replied, "and went away,--not hearing my voice." "What a wonder! What chance do you have of hearing your own voice?" She went over to the fire, lowered it, and began to put the potatoes on it, by degrees, till she had covered all the floor. "Why did you come here?" Deborah inquired, as she looked at the girl's face. "Just for the night. Where's the father? \\
\textbf{Text 2 (\model, T5-XXL-C4)}: I came to ask for food. Hur git me?" "You don't see it so bad," said Deborah, leaning her long arms about the girl's small shoulders. "You're poor." "That's true," the girl said, weeping softly. "Hus poor is no better than a beast's. I'd be well if they'd only set food on the table," she added, trembling again. "Evil grows in the dark," said Deborah sadly. \\\\
\textbf{Human Annotation 1}: Text 2. Both bit chaotic and funny, but the second one seems at least more outlandish. Being poor or saying Hur and Hus isn't as bad as putting potatos in fire that's covering the floor. \\
\textbf{Human Annotation 2}: Text 2 - Prefix text says she is hungry, so it would make sense that she would say she came to ask for food. \\
\textbf{Human Annotation 3}: Text 2. Deborah already told the girl in the prefix part where the father is, but she asks again in Text 1. So Text 2 makes more sense.\\\\
\midrule 
\midrule \\
\textbf{Prefix} (\textbf{Wikipedia}) \href{https://en.wikipedia.org/wiki/Homarus\_gammarus}{\emph{Homarus Gammarus}}: It may grow to a length of and a mass of , and bears a conspicuous pair of claws. In life the lobsters are blue, only becoming "lobster red" on cooking. Mating occurs in the summer, producing eggs which are carried by the females for up to a year before hatching into planktonic larvae. Homarus gammarus is a highly esteemed food, and is widely caught using lobster pots, mostly around the British Isles. Homarus gammarus is a large crustacean, with a body length up to and weighing up to , although the lobsters caught in lobster pots are usually long and weigh . Like other crustaceans, lobsters have a hard exoskeleton which they must shed in order to grow, in a process called ecdysis (molting). This may occur several times a year for young lobsters, but decreases to once every 1–2 years for larger animals. The first pair of pereiopods is armed with a large, asymmetrical pair of feet. \\\\

\textbf{Text 1 (Nucleus, T5-XXL-C4)}: They are armed with two large claws, whose upper surfaces are split into segmented claws, to allow them to grasp with maximum force. The claws are bent backwards when not in use, to keep them from rolling off the lobster's body, and are hooked into the shell. One claw can also bear a thick, sharp biting claw (known as the "hammer claw"). Both claws contain a fan-shaped set of “feathered” hairs, arranged around a central spine. \\
\textbf{Text 2 (\model, T5-XXL-C4)}: The foot is covered in spines, and each skeletal segment has a claw (which is a pair of claws in an asymmetrical arrangement). The claws bear nails (trapezoids) and powerful radulae, and are able to claw prey through a small hole in the ground. The second pair of pereiopods has a small foot, with three claws arranged symmetrically. The foot of the second pair is covered in concave spines which are longer and thinner than those of the first pair.\\\\
\textbf{Human Annotation 1}: Text 2. Because "The first pair of pereiopods is armed with a large, asymmetrical pair of feet. They (the feet???) are armed with two large claws..) just sounds weird. \\
\textbf{Human Annotation 2}: Text 2 - prefix mentions the feet and text 2 describes the foot. Text 1 just sounds odd as a follow on from prefix. \\
\textbf{Human Annotation 3}: Text 2. Both text seem to be very factually incorrect, but the second one at least appears more like a continuation of the sample text. The first text goes back to talk about the claws, while the second talks about feet and second pair of pereiopods. \\
\bottomrule
\end{tabular}
\end{center}
\vspace{-0.1in}
\caption{More model generations (Part 2) from each decoding method with human preferences and explanations. See our attached data submission for all 600 annotations for the 200 generation pairs.}
\vspace{-0.1in}
\label{tab:more-model-generations-2}
\end{table*}

\begin{table*}[t!]
\footnotesize
\begin{center}
\begin{tabular}{ p{14.8cm} } 
\toprule
\textbf{Prefix} (\textbf{PG19}) \href{https://www.gutenberg.org/files/55658/55658-h/55658-h.htm}{\emph{The Horse in History}} by \emph{Basil Tozer}: Two years after James I. had ascended the throne there set in one of the coldest winters this country has ever known, with the result that a long stretch of the River Ouse became frozen over and so afforded the king an opportunity, of which he was quick to avail himself, of organising a race-meeting on the ice. Drake tells us that the course extended “from the tower at the end of Marygate, under the great arch of the bridge, to the crane at Skeldergate Postern.” But even so early as this in the reign of King James the opponents of horse racing began to raise indignant protests against “the folly and wickedness of betting on running horses,” protests to which but scant attention was paid. Not until some years later did the extremely zealous clergyman named Hinde set seriously to work to denounce the practice of gambling in any and every form, and he appears then to have spoken and written so forcibly that many persons of intelligence and education—I quote from a trustworthy source—gathered round and strove to encourage him to the best of their ability.\\\\

\textbf{Text 1 (\model, GPT2-md)}: He did not, however, heed their advice and instead turned his attention to betting. "The king, who at that time was of a temperance and reform bent, had taken into his royal commission the recommendation that betting and horse-trading, while in principle harmless, be avoided. Such was the feeling among the clergy at that time as well as among the people, that in their opinions bettors and dealers, when dealing openly with men, did more harm than good.\\
\textbf{Text 2 (Nucleus, GPT2-md)}: ” Here again I have a record that can be said to be worthy of mention of a high level. There is one point which the facts of the situation require me to address, namely that from that day till our own day it is an object of our interest to try to remove all evidence of a race-course. That purpose is, I believe, the greatest thing that we can hope to attain in the course of this study.\\\\
\textbf{Human Annotation 1}: Text 1. The original paragraph mentions a "zealous clergyman" and "denounce the practice of gambling", and text 1 continues the discourse about "the clergy at the time" and "their opinions betters and dealers". Text 2 does not follow the same discourse, nor does it provide more information in prefix. \\
\textbf{Human Annotation 2}: Text 2 - prefix text is setting up for a quote with 'I quote from a trustworthy source, then Text 2 is the start of a quote as it begins with quotation marks. \\
\textbf{Human Annotation 3}: Text 1; Text 1 continues the narrative established around horse-race betting established in the prefix. The prefix mentions, "the opponents of horse racing began to raise indignant protests against [betting]", then, Text 1 continues this idea when it states, "He did not, however, heed their advice and instead turned his attention to betting".\\\\
\midrule 
\midrule \\
\textbf{Prefix} (\textbf{Wikipedia}) \href{https://en.wikipedia.org/wiki/Thelnetham\_Windmill}{\emph{Thelnetham Windmill}}: The mill is a small four storey tower mill with Patent sails and winded by a fantail. It drives two pairs of millstones, with a third pair driven by an auxiliary engine. The earliest record of a windmill in Thelnetham was a post mill which was shown of Joseph Hodgkinson's map of Suffolk of 1778. In 1818, the post mill was moved to Sandy Lane, Diss, Norfolk The post mill was replaced by a tower mill, which was built by millwright George Bloomfield for William Button in 1819. The mill was set to work on Christmas Day 1819. In 1832, the mill was modernised by the fitting of a cast-iron windshaft, Patent sails and a fantail. The new windshaft was fitted on 16 July 1832. It was made by J Aickman, the King's Lynn millwright, and weighs 1 3/4 tons (1,780 kg). A new stock was fitted in September 1836. William Button died on 11 February 1837. The mill passed jointly to his widow Rebecca and their son Richard. \\\\

\textbf{Text 1 (Nucleus, GPT2-md)}: In 1841, it was renamed The Postmill Mill in memory of its deceased owner. The mill has been restored to its original function. It is now used to run turbines for a turbine company. The mill was built for William Button. He used to work on it during the day, but he would also travel by carriage. In the years around 1816-17, he visited several other mills in the town, including his father's old mill. He bought it in 1810. This mill was not as modern as the mill on the left. \\
\textbf{Text 2 (\model, GPT2-md)}: Richard built the mill and sold it to John Beattie on 21 May 1845. On 8 June 1854, Thomas Moore purchased the mill from the Beattie family. In 1861 the mill was owned by the Earl of Dartmouth who rebuilt it as a tenement mill for a number of years, then sold it to a local paper company for £60 a week to pay for repairs. In 1913, the mill was purchased by David Cottrell. There is a stone chimney and iron grate set into the front wall of the mill. \\\\
\textbf{Human Annotation 1}: Text 2. Text 2 follows the original style of the paragraph by continuing the history of the mill, while Text 1 is redundant and repeats some things from the original text. Due to following the same pattern as the original paragraph, text 2 is the best choice. \\
\textbf{Human Annotation 2}: Text 2 - I really don't know, neither seem to fit. Text 1 says  ' The mill was built for William Button' but we've just been told William Button is dead so that seems to be a little odd to mention, so Text 2 is a better choice. \\
\textbf{Human Annotation 3}: Text 2; Text 2 is the better contintuation of the prefix. Prefix ends with mentioned that, "The mill passed jointly to his widow Rebecca and their son Richard. Then, the Text 2 begins with, "Richard built the mill" - this connects the prefix to Text 2. \\
\bottomrule
\end{tabular}
\end{center}
\vspace{-0.1in}
\caption{More model generations (Part 3) from each decoding method with human preferences and explanations. See our attached data submission for all 600 annotations for the 200 generation pairs.}
\vspace{-0.1in}
\label{tab:more-model-generations-3}
\end{table*}

\begin{table*}[t!]
\footnotesize
\begin{center}
\begin{tabular}{ p{14.8cm} } 
\toprule
\textbf{Prefix} (\textbf{Wikipedia}) \href{https://en.wikipedia.org/wiki/Daniel_Radcliffe}{\emph{Daniel Radcliffe}}: Radcliffe's performance in the show earned him Drama Desk Award, Drama League Award and Outer Critics Circle Award nominations. The production itself later received nine Tony Award nominations. Radcliffe left the show on 1 January 2012. Radcliffe's first post-Harry Potter project was the 2012 horror film The Woman in Black, adapted from the 1983 novel by Susan Hill. The film was released on 3 February 2012 in the United States and Canada, and was released on 10 February in the UK. Radcliffe portrays a man sent to deal with the legal matters of a mysterious woman who has just died, and soon after he begins to experience strange events and hauntings from the ghost of a woman dressed in black. He has said he was "incredibly excited" to be part of the film and described the script as "beautifully written". In 2013, he portrayed American beat poet Allen Ginsberg in the thriller drama Kill Your Darlings, directed by John Krokidas. \\\\

\textbf{Text 1 (\model, T5-XXL-C4)}: The film premiered at Sundance Film Festival in January 2013, where it won two awards. The film follows Ginsberg during the 1959 American poets' conference at Bennington College. Radcliffe next appeared in the film adaptation of Mary Shelley's Frankenstein. Following the film's premiere at the Venice Film Festival on 21 August 2013, it was released on 5 October 2013. Radcliffe also starred as William H. Macy in the animated adaptation of Bram Stoker's Dracula in 2013. The film was directed by Christian Rivers and was released on 27 July 2014. \\
\textbf{Text 2 (Nucleus, T5-XXL-C4)}: He described the film as "so much fun, and a reminder that life after Hogwarts is exciting and fun". He also starred in the 2016 thriller The Woman in the Window, which was produced by his agent Lydia Dean Pilcher, in which he plays the titular role of a boy who is sent to the country to be married to his pregnant aunt. In March 2018, Radcliffe appeared in the romantic comedy/drama Imposter, co-starring Rooney Mara. \\\\
\textbf{Human Annotation 1}: Text 1. The second text says he plays a titular role, while the title is "The Woman in the Window", being a contradiction. First text has no major issues and follows very similar structure of listing acting jobs Radcliffe had after the unnamed show (be that true or not). \\
\textbf{Human Annotation 2}: Text 1. The quotation at the beginning of Text 2 seems pretty off, as it follows the exact same structure as the quote in the prefix section - the movie + a quote in the movie. Also, the two quotes together make Radcliffs vocabulary seem limited to the word "exciting" and its cognates. Text 2 sounds automatical and forced. Plus, the "so much fun" bit sounds pretty childish for a grown man, especially together with the Hogwarts remark. \\
\textbf{Human Annotation 3}: Text 1 - This refers to the film date from prefix, 2013, the same film presumably as it mentioned Ginsberg, and with it being a thriller, it may not perhaps be described as being 'so much fun' . \\\\
\midrule 
\midrule \\
\textbf{Prefix} (\textbf{PG19}) \href{https://www.gutenberg.org/files/13089/13089-h/13089-h.htm}{\emph{The Making of Arguments}} by \emph{J.H. Gardiner}: But besides the omissions there were only six among the forty which did not contain positively wrong statements; in twenty-four papers up to ten per cent of the statements were free inventions, and in ten answers--that is, in one fourth of the papers--more than ten per cent of the statements were absolutely false, in spite of the fact that they all came from scientifically trained observers. Only four persons, for instance, among forty noticed that the <DW64> had nothing on his head; the others gave him a derby, or a high hat, and so on. In addition to this, a red suit, a brown one, a striped one, a coffee- jacket, shirt sleeves, and similar costume were invented for him. He wore in reality white trousers and a black jacket with a large red neck-tie. The scientific commission which reported the details of the inquiry came to the general statement that the majority of the observers omitted or falsified about half of the processes which occurred completely in their field of vision. \\\\

\textbf{Text 1 (\model, T5-XXL-C4)}: That is, they gave the impression that they had not seen a thing; or they had only seen a part of it, and put it down wrongly; or they only saw what they wanted to see, and presented a falsified impression of a given process. Only one third of the papers gave correct information, but that figure is based on the assumption that an observer watches more than half the processes at a time. How can it be explained that the observers give a false impression of processes? \\
\textbf{Text 2 (Nucleus, T5-XXL-C4)}: It also came to the conclusion that the  <DW64>, since he was not standing still in a given position, did not exert his full influence on the proceedings of the people within his area of activity, and that there was no need for special observations on the phenomena. But the whole suggestion of the Commission was not well received. There was, the commission declared, a noticeable scepticism about their conduct. Thus a report of the commission was sent to the official newspaper, and three days later the secretary of the observatory came to it and spoke frankly about the commission's conduct. \\\\
\textbf{Human Annotation 1}: Text 1. It seems to be closer aligned with the prefix, covering pretty much the same points. It still sounds off, with the numbers not adding up (Only one third of the papers gave correct information, but that figure is based on the assumption that an observer watches more than half the processes at a time. <<< different numbers in the prefix). Also the rhetorical question at the end discrupts the style. However, Text 2 with this fragments: " did not exert his full influence on the proceedings of the people within his area of activity, and that there was no need for special observations on the phenomena" makes even less sense. \\
\textbf{Human Annotation 2}: Text 1. Second text has an awkward "word" in it followed by a convoluted explanation of something that appears only vaguely related to the sample text. Latter part of text 2 talks about the conduct of the commission while most of the text 1 sticks to talking about the results. \\
\textbf{Human Annotation 3}:  Text 1 - This continues the concept that observers are not remembering what they saw, or not truly seeing but guessing. Text 2 is about something else. \\
\bottomrule
\end{tabular}
\end{center}
\vspace{-0.1in}
\caption{More model generations (Part 4) from each decoding method with human preferences and explanations. See our attached data submission for all 600 annotations for the 200 generation pairs.}
\vspace{-0.1in}
\label{tab:more-model-generations-4}
\end{table*}

\begin{table*}[t!]
\footnotesize
\begin{center}
\begin{tabular}{ p{14.8cm} } 
\toprule
\textbf{Prefix} (\textbf{PG19}) \href{https://gutenberg.org/files/38214/38214.txt}{\emph{Letters of Lord Acton}} by \emph{Lord Acton}: In that character he showed, when occasion came, that his {lxvi} long silence in Parliament had not been due to incapacity for public speaking. At Windsor he was agreeable to the Queen from his German tastes and sympathies, not to mention the fact that he could speak German as fluently as English. Every moment of leisure during his "wait" there was spent in the Castle library. Yet the position was an unnatural one, and Lord Acton soon became anxious to escape from it. His thoughts turned to his favourite Bavaria, and he humbly suggested the Legation at Stuttgart as a possible sphere. But something infinitely better than any political or diplomatic post remained for this born student and truly learned man. In 1895, just a year after Mr. Gladstone's resignation, Sir John Seeley, Professor of Modern History at Cambridge, departed this life. The Chair was in the gift of the Crown, that is, of the Prime Minister, and Lord Rosebery appointed Lord Acton. The appointment was singularly felicitous, and the opportunity came in the nick of time.\\\\

\textbf{Text 1 (\model, GPT2-md)}: Lord Acton entered upon his first year in the Chair, and the year was filled with extraordinary events, the first of which was the assassination of King George V, which had so disastrous an effect upon the world that it has never since been fully restored. Then came the attempt on the life of Sir Edmund Burke at St. Petersburg, the downfall of William of Orange, and the beginning of the Spanish War.\\
\textbf{Text 2 (Nucleus, GPT2-md)}: The "youngest and most handsome man of his class", as they called him, had been nominated to succeed him. Mr. Gladstone, whose wit had been so well known for many years, was now at last able to be seen in public with dignity, and his manner was becoming more relaxed. The Prince felt confident enough to nominate Sir John, who was the most distinguished scholar in Bavaria, and he received his nomination in person. There were some curious features of the office.\\\\
\textbf{Human Annotation 1}: Text 1 - The prefix paragraph focuses on the life and career of Lord Acton, thus, Text 1 is the best selection to continue exposition of Lord Acton's career.\\
\textbf{Human Annotation 2}: Text 1 - Prefix states that Lord Acton was appointed as chair, Text 1 then mentions Lord Acton entering his first year as chair. This seems to follow more closely than Text 2.\\
\textbf{Human Annotation 3}: Text 1. At the end of the paragraph, Lord Acton is appointed Chair and text 1 delves into what occurred once that was fact. Text 2 does not closely follow the narrative of the paragraph.\\\\
\midrule 
\midrule \\
\textbf{Prefix} (\textbf{Wikipedia}, \href{https://en.wikipedia.org/wiki/Battle_of_Roanoke_Island}{\emph{Battle of Roanoke Island}}): They arrived too late to take part in the battle, but were involved in the attack on the fortress at Pao-ting Fu, where the Chinese government was believed to have found asylum after Peking was captured by western forces. The Victorians joined a force of 7,500 men on a ten-day march to the fort, once again only to find that it had already surrendered. The Victorians then garrisoned Tientsin and the New South Wales contingent undertook garrison duties in Peking. HMCS Protector was mostly used for survey, transport, and courier duties in the Gulf of Chihli, before departing in November. The naval brigades remained during the winter, unhappily performing policing and guard duties, as well as working as railwaymen and fire-fighters. They left China in March 1901, having played only a minor role in a few offensives and punitive expeditions and in the restoration of civil order. Six Australians died from sickness and injury, but none were killed as a result of enemy action. \\\\

\textbf{Text 1 (Nucleus, GPT2-md)}: In July 1901 they began their work in the Indian Ocean, arriving in Hong Kong in November 1901. In January 1902, the Australian Expeditionary Force joined the Chinese in a massive invasion of Peking. While Australian naval forces took an active part in this event, the Australian Navy was still engaged in military operations in the Gulf of Hainan. When the Chinese invaded Hong Kong, the navy withdrew from the mainland and its warships were deployed at the port of Hong Kong until December 1902. \\
\textbf{Text 2 (\model, GPT2-md)}: A second invasion of the Chinese mainland was attempted by British naval forces on 23 June 1901 when Victoria signed a treaty with China. Following the treaty signing, a force led by HMAS Sault Ste. Marie was sent to occupy Peking and Tientsin. HMAS San Francisco, HMAS Mackellar and HMAS Melbourne returned to Hong Kong after a two-month deployment in China in early 1902 and were replaced by a group of 14,000 men under HMCS Lendl, which was formed on 24 November 1902 as part of the second invasion. \\\\
\textbf{Human Annotation 1}: Text 2; Text 2 is the better continuation of the prefix. In Text 1, it isn't clear who "they" is in the phrase, "they began their work in the Indian Ocean" which makes Text 1 appear disjointed when reading directly after the prefix whereas Text 2's introduction flows more seamlessly even though it's introduction brings a slight change in idea. \\
\textbf{Human Annotation 2}: Text 1. Although both texts could follow the paragraph, Text 1 follows along with the timeline set in the paragraph. \\
\textbf{Human Annotation 3}: Text 2 - very difficult without more knowledge of these events. I'm picking text 2 just because the date mentioned, 23 June 1901, is closest to the date mentioned in prefix text - march 1901 \\
\bottomrule
\end{tabular}
\end{center}
\vspace{-0.1in}
\caption{More model generations (Part 5) from each decoding method with human preferences and explanations. See our attached data submission for all 600 annotations for the 200 generation pairs.}
\vspace{-0.1in}
\label{tab:more-model-generations-5}
\end{table*}
\end{document}